\documentclass[aos]{imsart}
\usepackage{booktabs}
\usepackage{bm}
\RequirePackage{amsthm,amsmath,amsfonts,amssymb}
\RequirePackage[numbers]{natbib}
\RequirePackage[colorlinks,citecolor=blue,urlcolor=blue]{hyperref}
\RequirePackage{graphicx}
\usepackage{xcolor}

\startlocaldefs
\numberwithin{equation}{section}
\theoremstyle{plain}
\newtheorem{Corollary}{Corollary}[section]
\newtheorem{Proposition}{Proposition}[section]
\newtheorem{Lemma}{Lemma}
\newtheorem{Theorem}{Theorem}[section]
\theoremstyle{remark}

\newtheorem{Assumption}{Assumption}
\usepackage[linesnumbered,ruled,vlined]{algorithm2e}
\usepackage{float}

\newcommand{\Var}{{\rm Var}}

\newcommand{\FF}{{\mathcal{F}}}
\newcommand{\HH}{{\mathcal{H}}}
\newcommand{\E}{\mathbb{E}}
\newcommand{\R}{\mathbb{R}}
\newcommand{\prob}{\mathbb{P}}
\newcommand{\C}{\mathcal{C}}
\newcommand{\PP}[1]{\mathbf{P}^{(#1)}}

\newcommand{\RR}{{\mathbf{R}}}
\newcommand{\QQ}{{\mathbf{Q}}}
\newcommand{\TT}{\mathbf{T}}
\newcommand{\A}{\mathcal{A}}
\newcommand{\B}{\mathcal{B}}
\newcommand{\Y}[1]{Y^{(#1)}}
\newcommand{\X}[1]{X^{(#1)}}
\newcommand{\G}[1]{G^{(#1)}}
\newcommand{\XQ}{X^\QQ}
\newcommand{\YQ}{Y^\QQ}
\newcommand{\eps}[1]{\varepsilon^{(#1)}}
\newcommand{\f}[1]{f^{(#1)}}
\newcommand{\hf}[1]{\widehat{f}^{(#1)}}
\newcommand{\fH}{f^*_\HH}
\newcommand{\hfH}{\widehat{f}_\HH}
\newcommand{\tfH}{\widetilde{f}_\HH}

\newcommand{\tGamma}{\widetilde{\Gamma}}
\newcommand{\hGamma}{\widehat{\Gamma}}
\newcommand{\w}[1]{\omega^{(#1)}}
\newcommand{\hw}[1]{\widehat{\omega}^{(#1)}}
\newcommand{\Del}[1]{\Delta^{(#1)}}
\newcommand{\eigenGamma}{\lambda_{\rm min}(\Gamma)}
\newcommand{\Err}{{\rm Err}}

\newcommand{\N}{\mathcal{N}}

\newcommand{\DRoLn}{\texttt{DRoL0}}

\newcommand{\ERM}{\texttt{ERM}}
\newcommand{\ind}{\mathbf{1}}

\DeclareMathOperator*{\argmax}{arg\,max}
\DeclareMathOperator*{\argmin}{arg\,min}
\newcommand{\RN}[1]{%
  \textup{\uppercase\expandafter{\romannumeral#1}}%
}
\usepackage[T1]{fontenc}

\endlocaldefs

\begin{document}

\begin{frontmatter}
\title{Distributionally Robust Learning for Multi-source Unsupervised Domain Adaptation}
\runtitle{Distributionally Robust Learning for MSDA}

\begin{aug}
\author[A]{\fnms{Zhenyu}~\snm{Wang}\ead[label=e1]{zw425@stat.rutgers.edu}},
\author[B]{\fnms{Peter}~\snm{B\"{u}hlmann}\ead[label=e2]{peter.buehlmann@stat.math.ethz.ch}}
\and
\author[A]{\fnms{Zijian}~\snm{Guo}\ead[label=e3]{zijguo@stat.rutgers.edu}}
\address[A]{Department of Statistics, Rutgers University\printead[presep={,\ }]{e1,e3}}

\address[B]{Seminar for Statistics, ETH Z\"urich\printead[presep={,\ }]{e2}}
\end{aug}

\begin{abstract}
Empirical risk minimization often performs poorly when the distribution of the target domain differs from those of the source domains. To address such potential distributional shifts, we develop an unsupervised domain adaptation approach
that leverages labeled data from multiple source domains and unlabeled data from the target domain. 
We introduce a distributionally robust model that optimizes an adversarial reward based on explained variance across a class of target distributions, ensuring generalization to the target domain. We show that the proposed robust model is a weighted average of conditional outcome models from the source domains. 
This formulation allows us to compute the robust model through the aggregation of source models, which can be estimated using various machine learning algorithms of the user’s choice such as random forests, boosting, and neural networks.
Additionally, we introduce a bias-correction step to obtain a more accurate aggregation weight, which is effective for  various machine learning algorithms.
Our framework can be interpreted as a distributionally robust federated learning approach that satisfies privacy constraints while providing insights into the importance of each source for prediction on the target domain. The performance of our method is evaluated on both simulated and real data.

\end{abstract}



\end{frontmatter}


\section{Introduction}
A fundamental assumption for the success of machine learning (ML) algorithms is that the training and test data sets share the same underlying distribution. However, when a domain shift occurs -- that is, when the distribution of the test data differs from that of the training data -- classical statistical learning algorithms often exhibit unreliable performance, even when they are carefully fine-tuned on the training data  \cite{quinonero2008dataset, koh2021wilds, malinin2021shifts}. 
Developing methods that remain robust and achieve excellent prediction performance under domain shift is both critical and challenging.

\begin{figure}[H]
    \centering
    \includegraphics[width=0.65\linewidth]{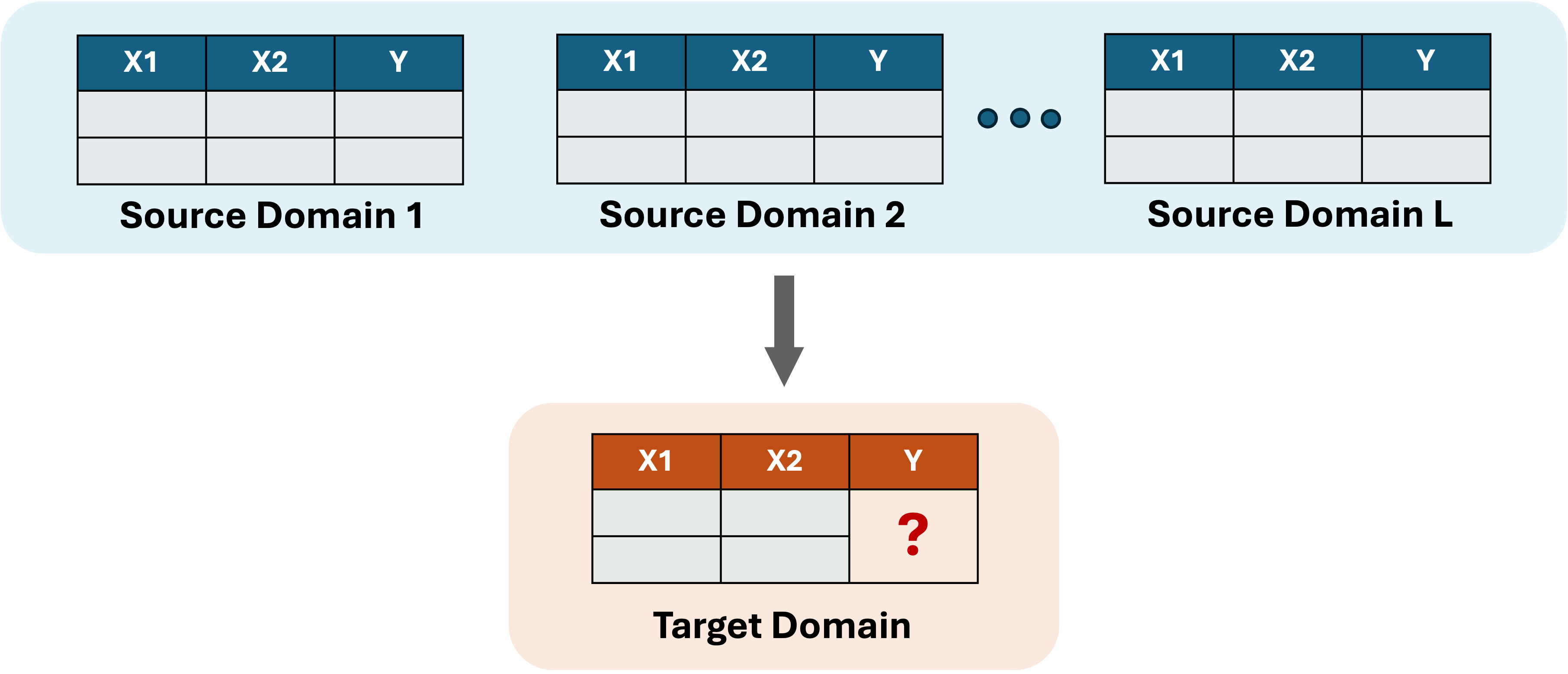}
    \vspace{-3.5mm}
    \caption{Illustration of the Multi-source Unsupervised Domain Adaptation (MSDA) framework, where source domains have labeled data and the target domain only has unlabeled data.}
    \label{fig: msda}
    \vspace{-1.5em}
\end{figure}

Domain adaptation techniques have proven to be effective for improving generalization
by transferring knowledge from one or multiple source domains to an unseen target domain \cite{mansour2008domain, duan2012domain, ganin2016domain, saito2018maximum}.
In contrast to traditional multi-source settings, where only data from multiple source domains are available, the multi-source unsupervised domain adaptation (MSDA) framework leverages not only multiple labeled source datasets but also unlabeled target data, as depicted in Figure \ref{fig: msda}. 
 By incorporating the unlabeled target observations, MSDA may lead to better predictive models tailored for the target domain. This approach is particularly important in fields such as healthcare \cite{schulam2015framework, perone2019unsupervised}
or finance \cite{cao2021data}, where the acquisition of target labels is often expensive or infeasible, and the target domain can differ substantially from any of the source domains. Meanwhile, compared to the semi-supervised setting, which considers more covariate data than outcome labels, this work focuses on the fully unsupervised setting for the target domain, where no target labels are available.

We now introduce a formal setup of our focused MSDA regime. Suppose that there are
independent datasets collected from $L$ source domains. For each source $1\leq l \leq L$, the data $\{\X{l}_i, \Y{l}_i\}_{1\leq i\leq n_l}$ are assumed to be i.i.d. samples generated according to:
\begin{equation}
    \X{l}_i {\sim} \PP{l}_X \quad \text{and} \quad
    \quad
    \Y{l}_i = \f{l}(\X{l}_i) + \eps{l}_i, 
    \label{eq: source-models}
\end{equation}
where $\X{l}_i\in \R^p$ and $\Y{l}_i\in \R$ represent the covariates and outcome  respectively, and $\f{l}(x) = \E[\Y{l}_i | \X{l}_i=x]$ stands for the conditional outcome model for the $l$-th source domain. Here, both the covariate distribution $\PP{l}_X$ and the conditional outcome model $\f{l}(\cdot)$ are allowed to vary by source $l$. 
Meanwhile, we denote the target distribution by $\QQ=(\QQ_{X},\QQ_{Y|X})$, which may differ from any of the source distributions in \eqref{eq: source-models}. Under the MSDA regime, although we have access to i.i.d. covariate observations $\{\XQ_j\}_{1\leq j \leq N}$ drawn from $\QQ_{X}$, no corresponding outcome labels are available.


\subsection{Our results and contribution}
Most existing unsupervised domain adaptation methods rely on the assumption that the conditional outcome distributions of the source and target domains are similar \cite{mansour2008domain, weiss2016survey, zhuang2020comprehensive}. 
However, verifying this similarity is inherently challenging, especially when outcome labels are unavailable in the target domain. 
To overcome this limitation, we borrow ideas from the distributionally robust optimization (DRO) framework \cite{ namkoong2017variance, sinha2017certifying,ben2013robust}, and introduce a robust prediction model that ensures robust predictive performance even when the target domain’s conditional outcome distribution deviates from those of the source domains.

We define an uncertainty set comprising target populations whose conditional outcome distributions are generated as mixtures from those of source populations. The proposed robust prediction model is then formulated to optimize the worst-case reward (about the explained variance) evaluated over this uncertainty set. In practical scenarios, domain experts may possess prior knowledge about the target population or seek to ensure robustness within a specific region around the observed mixture of source distributions. We incorporate such prior information into constructing the uncertainty set, which may improve the prediction accuracy of our proposed robust prediction model \cite{hu2018does, soma2022optimal}.

Our primary contribution, stated in Theorem \ref{thm: identification}, demonstrates that our distributionally robust model, 
defined via a minimax optimization, 
can be identified (in the population sense) as a weighted average of the individual source models $\{\f{l}\}_{1\leq l\leq L}$ in \eqref{eq: source-models}. The optimal aggregation weights are determined by solving a convex quadratic program, thereby simplifying the original minimax formulation and enabling easy implementation using standard quadratic programming solvers. 
Moreover, we introduce a novel bias correction step to obtain more accurate aggregation weights. We refer to our entire methodology as \textbf{D}istributionally \textbf{Ro}bust \textbf{L}earning (DRoL), which first fits individual source conditional mean models using any ML method of the user's choice, and then aggregates these models into a robust prediction model specifically tailored for the target domain. We leverage the weighted average formulation to establish the convergence rates of our proposed DRoL method in terms of both the objective function and the model parameters. We further demonstrate that the bias correction step improves these rates for certain regimes. 
As special but essential examples, we show that the estimation error for the model parameters achieves $n^{-1/2}$ in low-dimensional linear regression and $\sqrt{s\log p/n}$ in the high-dimensional setting, with $s$ denoting the maximum sparsity level in the source domains.

As another contribution, we explore and establish identification results for distributionally robust models that use alternative loss functions, including  standard squared error and regret loss \cite{agarwal2022minimax}. Our analysis reveals that the choice of loss function is pivotal for the constructed robust models in the MSDA setting. Specifically, while models employing squared error loss are similarly expressed as weighted averages analogous to DRoL, their aggregation weights are sensitive to heterogeneous noise levels across source domains. Regret-based models effectively mitigate the impact of heterogeneous noise, yet become NP-hard to solve when general prior information about the target domain is incorporated. In contrast, our proposed robust model not only alleviates the impact of heterogeneous noise levels but also effectively accommodates the general prior information; see the detailed discussions in Section \ref{subsec: other losses}.

In summary, the main contributions of this paper are as follows. 
\begin{itemize}
    \item For the MSDA  setting, we introduce a theoretically justified framework for constructing distributionally robust models, where 
    the identification result in Theorem \ref{thm: identification} reduces the minimax optimization to a standard quadratic program. 
    \vspace{1.5mm}
    \item We explore various losses when constructing distributionally robust models, and demonstrate the unique advantages offered by the proposed reward function.
     \vspace{1.5mm}
    \item We propose a bias-correction method to more accurately estimate the optimal aggregation weights, which are then used to construct the distributionally robust model. This method is compatible with various ML algorithms and achieves a faster convergence rate compared to direct plug-in approaches.
\end{itemize}

\subsection{Related works}
This section further discusses the connections and differences between our proposal and the existing literature. 
\vspace{.5em}

\noindent\textit{Unsupervised Domain Adaptation.} Inspired by the foundational work of \cite{ben2006analysis}, most unsupervised domain adaptation methods aim to align the distributions between source and target domains by minimizing a specific distance metric \cite{hu2015deep, long2018transferable, ren2018generalized}.
Another prominent approach employs adversarial training techniques to map source and target domains into a shared representation space, ensuring that a discriminator cannot distinguish between them \cite{ganin2016domain, tzeng2017adversarial, saito2018maximum}.
These existing methods implicitly assume that the conditional outcome distributions given the covariates, or their learned representations, are similar across both source and target domains, which is particularly challenging to verify without labeled target data. 
In contrast, without requiring such a similarity condition, our work leverages the idea of DRO to develop a robust prediction model that maintains good performance as long as the target distribution resides within a predefined uncertainty set.
\vspace{1.5mm}

\noindent\textit{Group DRO and agnostic federated learning.}
Existing approaches in Group DRO  \cite{sagawa2019distributionally, hu2018does, zhang2020coping} and agnostic federated learning \cite{mohri2019agnostic, deng2020distributionally} 
primarily aim to construct generalizable prediction models using data from source domains without using information from the target domain. In contrast, as shown in Figure \ref{fig: msda}, our work focuses on the MSDA setting, where, in addition to data from multiple sources, observations of target covariates are also available. We leverage this extra information to design the distributionally robust model tailored for this target domain.
{This difference in settings leads to distinct definitions of the uncertainty class and algorithmic designs, as detailed in Sections \ref{subsec: model} and \ref{subsec: alg}.}
Moreover, whereas Group DRO works typically
focus on classification applications, our approach is specially designed for the regression task.
\vspace{1.5mm}

\noindent\textit{{Minimax Regret Optimization.}} 
Minimax regret models have been explored in various contexts to address noise sensitivity in distributional shifts. In single-source settings, \cite{agarwal2022minimax} introduced a minimax regret framework to mitigate the impact of noise levels across source and target distributions. Recent extensions include multi-source linear regression \cite{mo2024minimax} and conditional average treatment effect estimation in causal inference \cite{zhang2024minimax}. However, these works restrict the constraint set to simplexes or polyhedrons, ensuring computational tractability. In Section \ref{subsec: other losses}, we show that such regret-based approaches become NP-hard for general convex constraints in the MSDA setting, limiting their ability to incorporate flexible prior information about the target distribution. In contrast, our proposed reward-based models not only mitigate the impact of heterogeneous noise levels across source domains, but also integrate the prior in a computationally efficient manner; see Section \ref{subsec: other losses} for the detailed discussions.

\vspace{1.5mm}

\noindent \textit{Maximin Effect.} Previous works
\cite{meinshausen2015maximin, buhlmann2015magging, rothenhausler2016confidence, guo2020inference} studied the estimation and inference for the maximin effect, a particular type of robust prediction model using linear models. 
This paper extends the concept of the maximin effect to encompass ML algorithms, moving beyond the linear models.
Moreover, our DRoL approach is designed to incorporate prior knowledge about the target distribution, resulting in a less conservative model.

\subsection{Notations}  For a positive integer $m$, define $[m] = \{1,..., m\}$. For real numbers $a$ and $b$, define $a\vee b = \max\{a,b\}$ and $a\wedge b = \min\{a, b\}$. We use $c$ and $C$ to denote generic positive constants that may vary from place to place. For positive sequences $a(n)$ and $b(n)$, we use $a(n)\lesssim b(n)$ 
to represent that there exists some universal constant $C > 0$ such that $a(n) \leq C\cdot b(n)$ for all 
$n\geq 1$, and denote $a(n) \asymp b(n)$ if $a(n)\lesssim b(n)$ and $b(n)\lesssim a(n)$. We use notations $a(n)\ll b(n)$ 
if $\limsup_{n\rightarrow\infty} (a(n)/b(n)) = 0$. For a vector $x \in \R^p$, we define its $\ell_q$ norm as $\|x\|_q = (\sum_{i=1}^p |x_i|^q)^{1/q}$ for $q \geq 0$. For a numeric value $a$, we use notation $a_p$ to represent a length $p$ vector whose every entry is $a$. For a matrix $A$, we use $\|A\|_F, \|A\|_2$ and $\|A\|_\infty$ to denote its Frobenius norm, spectral norm, and element-wise maximum norm, respectively. We use $I_p$ to denote the $p$-dimensional identity matrix. Let $n_l$ and $N$ represent the sample size for the $l$-th source domain and the target domain, respectively.

\section{Distributionally Robust Prediction Models: Definition and Identification}
\label{sec: formulation}
We consider the multi-source unsupervised domain adaptation (MSDA) setting, where we have labeled samples from $L$ source domains and only unlabeled data from the target domain. For each source domain $l\in [L]$, let $\{\X{l}_i, \Y{l}_i\}_{i\in [n_l]}$ be i.i.d. samples drawn from the distribution $\PP{l} = (\PP{l}_X, \PP{l}_{Y|X})$.  
For the target domain, characterized by $\QQ = (\QQ_X, \QQ_{Y|X})$, we observe only unlabeled covariate samples $\{X^\QQ_j\}_{j\in [N]}$. 
Our goal is to build a prediction model that robustly generalizes to the target domain, accommodating both covariate shift (where $\QQ_X$ may differ from any $\PP{l}_X$), and posterior drift (where $\QQ_{Y|X}$ may differ from any $\PP{l}_{Y|X}$). 


\subsection{Group Distributionally Robust Prediction Models}
\label{subsec: model}
In the absence of the labeled samples from the target domain, identifying its conditional outcome distribution $\QQ_{Y|X}$ becomes particularly challenging, especially when we allow $\QQ_{Y|X}$ to be different from the source domain's conditional outcome distributions. Instead of aiming at recovering $\QQ_{Y|X}$, we propose a distributionally robust prediction model that ensures reliable predictive performance across a range of distributions, potentially encompassing the target $\QQ$.

In contrast to the unidentifiable \(\QQ_{Y|X}\), the target covariate distribution \(\QQ_X\) is empirically accessible via
the available target covariate samples \(\{X_j^{\QQ}\}_{j\in [N]}\). Based on $\QQ_X$, we define the uncertainty class of distributions as
\begin{equation}
    \C(\QQ_X):= \left\{\left.\TT= (\QQ_X, \TT_{Y|X}) \; \right\vert \; \TT_{Y|X} = \sum_{l=1}^L q_l\cdot \PP{l}_{Y|X} \quad \text{with}\quad q\in \Delta^L
    \right\},
    \label{eq: region-Q full}
\end{equation}
where $\Delta^L= \{q \in \R^L \vert \sum_{l=1}^L q_l=1,\; \min_l q_l \geq 0\}$ denotes the $(L-1)$-dimensional simplex. Notably, if the true target conditional distribution $\QQ_{Y|X}$ is indeed a mixture of $\{\PP{l}_{Y|X}\}_{l\in[L]}$, then this uncertainty set $\C(\QQ_X)$ contains the actual target population $\QQ$.

Unlike previous Group DRO works \cite{sagawa2019distributionally, hu2018does, zhang2020coping}, which are designed for the regime  without target covariate information and thus define uncertainty sets over the joint distribution of $(X,Y)$, our proposed uncertainty set is tailored to the target domain by fixing \(\QQ_X\) and focusing solely on the conditional distribution $Y|X$. In the no covariate shift scenario, the uncertainty sets defined by Group DRO and our approach coincide. However, in the presence of  covariate shift, we utilize the available target covariate samples in the MSDA regime and focus solely on variations in the conditional distribution $Y|X$. 

Next, we introduce the \emph{reward function} or the negative loss function, which serves as the basis for defining our distributionally robust prediction model. We will discuss the benefits of using our particular reward function over alternative loss functions in Section \ref{subsec: other losses}. Given a distribution $\TT$ and a prediction model $f(\cdot)$, the reward function $\RR_\TT(f)$ is defined as
\begin{equation}
    \RR_{\TT}(f) := \E_{(X,Y)\sim\TT}[Y^2 - (Y - f(X))^2],
    \label{eq: reward}
\end{equation}
where $\E_{(X,Y)\sim\TT}$ denotes the expectation taken with respect to the data $(X,Y)$ following the distribution $\TT$. The reward function $\RR_\TT(f)$ evaluates the predictive performance of the model $f(\cdot)$ relative to the null model, evaluated on the distribution $\TT$. For the centered outcome variable (i.e., $\E_\TT[Y] = 0$), $\RR_\TT(f)$ corresponds to the variance of the outcome explained by the model $f(\cdot)$. 
Therefore, a larger value of $\RR_\TT(f)$ indicates better prediction performance of the model $f(\cdot)$. 

Building upon the uncertainty set $\C(\QQ_X)$, we consider the worst-case reward of a prediction model $f$ over this set, i.e. $\min_{\TT \in \C(\QQ_X)} \RR_\TT(f)$. We then introduce the robust prediction model $f^*$ to maximize this worst-case reward, defined as
\begin{equation}
    f^* := \argmax_{f\in \FF}\min_{\TT \in \C(\QQ_X)} \RR_\TT(f),
    \label{eq: f_star full}
\end{equation}
where $\FF$ denotes a pre-specified function class. Using the loss function terminology, we rewrite \eqref{eq: f_star full} into an equivalent minimax optimization as follows:
$$
f^* := \argmin_{f\in \FF} \max_{\TT \in \C(\QQ_X)}\E_\TT[\ell(X,Y; f)] 
\quad \text{with}\quad \ell(x,y; f) = (y-f(x))^2 - y^2.$$ 

Since $f^*$ optimizes the worst-case reward over the entire convex hull of $\{\PP{l}_{Y|X}\}_{l\in[L]}$, this approach may result in conservative prediction performance due to the broad uncertainty set. A common strategy to mitigate this conservatism is to incorporate prior knowledge about the target distribution, thereby refining the uncertainty set \cite{hu2018does, mohri2019agnostic, zhang2020coping,soma2022optimal}.
For instance, suppose prior expertise suggests that the target's $\QQ_{Y|X}$ is a mixture of the sources' $\{\PP{l}_{Y|X}\}_{l\in [L]}$ with mixture weights close to a pre-specified weight vector $q^{\rm prior}\in \Delta^L$. In this case, we can restrict the target mixture weights to lie within the refined region $\HH = \{q\in \Delta^L\mid\|q - q^{\rm prior}\|_2\leq \rho\}$, where $\rho>0$ is a user-specified parameter controlling the size of $\HH$.

Let $\HH\subseteq \Delta^L$ represent the prior knowledge regarding the sourcing mixture for the target domain. We define the refined uncertainty set as
\begin{equation}
    \C(\QQ_X, \HH):= \left\{\left.\TT = (\QQ_X, \TT_{Y|X})\;\right\vert \; \TT_{Y|X} = \sum_{l=1}^L q_l \cdot \PP{l}_{Y|X} \quad \text{with}\quad q\in \HH  \right\}.
    \label{eq: region-Q H}
\end{equation}
This uncertainty set $\C(\QQ_X, \HH)$ is a subset of $\C(\QQ_X)$ defined in \eqref{eq: region-Q full}, thereby narrowing the range of plausible distributions by leveraging the prior information encapsulated in $\HH$.
With the refined $\C(\QQ_X, \HH)$, we define the corresponding robust prediction model $\fH$ as:
\begin{equation}
    \fH := \argmax_{f\in \FF} \min_{\TT \in \C(\QQ_X, \HH)} \RR_\TT (f).
    \label{eq: f_star H}
\end{equation}

 The specification of $\C(\QQ_X, \HH)$ involves balancing predictiveness against the robustness of $\fH$. If $\C(\QQ_X, \HH)$ contains the target distribution $\QQ$, this smaller uncertainty set facilitates a more accurate prediction model. In contrast, a larger uncertainty set $\C(\QQ_X)$, which does not incorporate prior knowledge $\HH$, is more likely to contain $\QQ$ and thus produce a more robust but potentially conservative prediction model. We recommend specifying a reasonable $\HH$ based on domain expertise, with $\HH=\Delta^L$ serving as a default where prior knowledge is limited. We have explored how different specifications of $\HH$ affect the model predictiveness and robustness in the numerical studies in Sections \ref{sec: simu-H} and \ref{sec: real data}. We demonstrate in Sections \ref{sec: simu-H} and \ref{sec: real data} that integrating such prior information can significantly mitigate the conservativeness and improve the prediction accuracy of the robust prediction model.
 
We conclude this subsection with a remark on the connection with minimax fairness.
In scenarios without covariate shift (i.e. $\QQ_X = \PP{l}_X$ for all $l$), our robust model
\eqref{eq: f_star full} aligns with the minimax fairness framework \cite{martinez2020minimax, diana2021minimax}; see the details in Section A.2 of Supplements \cite{wang2025supplements}.
In the covariate shift setting,  our proposal serves as a distributionally robust model in the MSDA regime, but it remains unclear whether it can be directly related to minimax fairness.

\subsection{Identification for Distributionally Robust Prediction Models}
\label{subsec: identification}

In this subsection, we establish that $\fH$, initially formulated via a minimax optimization problem in \eqref{eq: f_star H}, shall be explicitly characterized as a weighted average of individual source models $\{\f{l}\}_{l\in[L]}$. 
\begin{Theorem}
Suppose that the function class $\FF$ is convex with $\f{l} \in \FF$ for all $l\in[L]$ and $\HH$ is a convex subset of $\Delta^L$, then $\fH$ defined in \eqref{eq: f_star H} is identified as:
    \begin{equation}
        \fH = \sum_{l=1}^L q_l^* \cdot \f{l} \quad \textrm{with} \quad 
        q^* = \argmin_{q\in \HH}q^\intercal \Gamma q,  
        \label{eq: identification}
    \end{equation}
where $\Gamma_{k,l} = \E_{\QQ_X}[\f{k}(X)\f{l}(X)]$ for $k,l\in [L]$.
    \label{thm: identification}
\end{Theorem}

This theorem provides a population-level identification result, demonstrating that $\fH$ is identified as a weighted aggregation of the true source models $\{\f{l}\}_{l\in [L]}$. It motivates our proposed data-dependent algorithm in Section \ref{subsec: alg}:
we apply the existing ML algorithms to estimate these individual source models $\{\f{l}\}_{l\in [L]}$ and then compute the optimal aggregation weights by solving the data-dependent version of the quadratic program in \eqref{eq: identification}. We finally estimate $\fH$ using estimated source models and optimal aggregation weights; see Algorithm \ref{alg: cs mm} in Section \ref{subsec: alg} for details.

\begin{figure}[H]
    \vspace{-1em}
    \centering
    \includegraphics[width=0.5\textwidth]{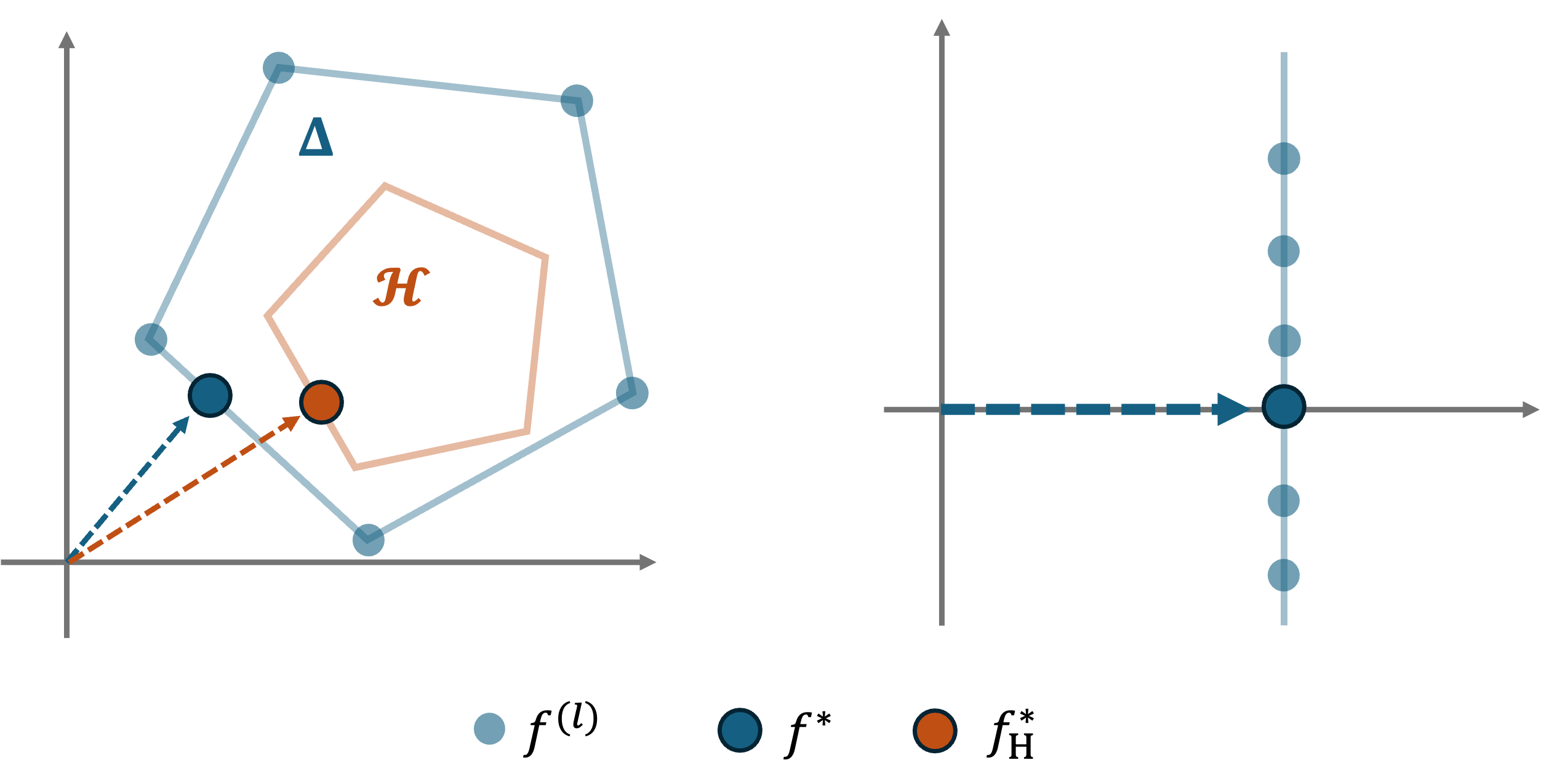}
    \vspace{-3.5mm}
    \caption{Illustration of $f^*$ (the blue point) and $\fH$ (the red point) for $p=2$, $L=5$, and the additive models $\f{l}(x)=\sum_{j=1}^{2}\f{l}_{j}(x_j)$ for ${l\in[L]}$. The left panel: $f^*$ is the point closest to the origin in the convex hull of $\{\f{l}\}_{l\in[L]}$, and $\fH$ is the point in the $\HH$-constrained set having the smallest distance to the origin; The right panel: consider the setting with shared first component $\f{1}_1 = \f{2}_1 = ... =\f{L}_1 = f_1$ and the second component being scattered around $0$; the distributional robust prediction model $f^*(x)=f_1(x_1)$ (blue point) retains only the shared component and shrinks the sign heterogeneous component to zero. }
    \label{fig: illus mag}
    \vspace{-1.5em}
\end{figure}

Next, we provide a geometric interpretation of the result established in Theorem \ref{thm: identification}. The quadratic objective $q^\intercal \Gamma q$ in \eqref{eq: identification} is equivalent to $\E_{X\sim \QQ_X}[\sum_{l=1}^L q_l\f{l}(X))]^2$, representing the distance from the aggregated model $\sum_{l=1}^L q_l \f{l}$ to the origin. 
Geometrically, $f^*$ corresponds to the point within the convex hull of $\{\f{l}\}_{l\in[L]}$ that is closest to the origin{, minimizing the second-order moment}, whereas $\fH$ denotes the point closest to the origin within the $\HH$-constrained set, as shown in the left panel of Figure \ref{fig: illus mag}. 

Moreover, the robust prediction model $\fH$ tends to capture shared associations across source domains while eliminating heterogeneous effects. To illustrate this, consider additive models with $p=2$ for simplicity. For each source $l$, suppose the individual source model is $\f{l}(x) = f_1^{(l)}(x_1) + f_2^{(l)}(x_2)$, where the first component $f_1^{(l)}(x_1)=f_1(x_1)$ is shared across all source domains, but the second component $f_2^{(l)}(x_2)$ varies with $l$ and is randomly scattered around $0$. In this scenario, our proposed robust model $f^*(x) = f_1(x_1)$ effectively captures the shared component across source domains while removing the heterogeneous effect, as depicted in the right panel of Figure \ref{fig: illus mag}. We numerically demonstrate this property in the case of high-dimensional sparse linear models in Section B.2 of Supplements \cite{wang2025supplements}.
This finding generalizes results from previous studies \cite{meinshausen2015maximin, guo2020inference}, which primarily focused on linear models.

We now highlight the distinctions between our identification strategy and those employed in Group DRO works \cite{sagawa2019distributionally, hu2018does}. Group DRO methods directly tackle the minimax problem $\min_{f\in \FF} \max_{1\leq l \leq L} {\E}_{\PP{l}}[\ell(X, Y; f)]$ by alternately solving the inner maximization and the outer minimization via neural networks. This procedure is feasible because each source domain provides paired data $(X,Y)$ drawn from $\PP{l} = (\PP{l}_X, \PP{l}_{Y|X})$.
In contrast, under the MSDA setting, directly applying such a strategy to solve the proposed minimax problem $$\min_{f\in \FF} \max_{\TT\in \C(\QQ_X)} {\E}_\TT[\ell(X,Y;f)] = \min_{f\in \FF} \max_{1\leq l\leq L} {\E}_{(\QQ_X, \PP{l}_{Y|X})}[\ell(X,Y;f)],$$ requires the paired data $(X,Y)$ drawn from $(\QQ_X, \PP{l}_{Y|X})$ for each $l$, which is impractical to obtain in the covariate shift regime. To overcome this challenge, Theorem \ref{thm: identification} identifies the distributionally robust model $\fH$ as a weighted aggregation of the source models $\{\f{l}\}_{l\in [L]}$, where these source models themselves are estimated using general ML algorithms.

\subsection{Exploration of Various Loss Functions for Distributionally Robust Models}
\label{subsec: other losses}
The choice of the loss function is crucial 
for constructing distributionally robust models in the MSDA setting.
In the following, we establish identification results for robust models that use different loss functions, including the standard squared error and regret loss \cite{agarwal2022minimax}, and discuss the advantages of using the reward function that we have mainly adopted in the present paper.

We start with the squared error loss, which leads to the following robust prediction model:
\begin{equation}
    f^{\rm sq}_{\HH} = \min_{f\in \FF}\max_{\TT\in \C(\QQ_X,\mathcal{H})} \E_{(X,Y)\sim \TT}[(Y - f(X))^2],
    \label{eq: sol - squared error}
\end{equation}
where $\FF$ denotes a pre-specified function class, and $\HH\subseteq \Delta^{L}$ encodes the prior information about the sourcing mixture in the target domain. 
For each source domain, we define the noise level as $[\sigma^{(l)}]^2 = \E[(\eps{l}_i)^2|\X{l}_i]$ with $\eps{l}_i = \Y{l}_i - \f{l}(\X{l}_i)$. We further define the vector $\bm{\sigma}^2 = \left((\sigma^{(1)})^2, ..., (\sigma^{(L)})^2\right)^\intercal$. The following proposition shows that $f^{\rm sq}_\HH$ admits the weighted average expression, analogous to the proposed $\fH$ as established in Theorem \ref{thm: identification}. 
\begin{Proposition}
    Suppose that the function class $\FF$ is convex with $\f{l}\in \FF$ for all $l\in [L]$ and $\HH$ is a convex subset of $\Delta^{L}$, then $f^{\rm sq}_\HH$ defined in \eqref{eq: sol - squared error} is identified as:
    \begin{equation}
        f^{\rm sq}_{\HH} = \sum_{l=1}^L q_l^{\rm sq}\cdot f^{(l)}\quad \textrm{with}\quad q^{\rm sq} = \argmin_{q\in \HH}\left\{ q^\intercal \Gamma q - q^\intercal (\gamma + \bm{\sigma}^2)\right\},
    \end{equation}
    where $\Gamma_{k,l}=\E_{\QQ_X}\E[\f{k}(X)\f{l}(X)]$ for $k,l\in [L],$ and $\gamma_{l} = \Gamma_{l,l}$ for $l\in [L]$.
    \label{prop: squared error}
\end{Proposition}
This proposition indicates that $f^{\rm sq}_{\HH}$ is also characterized by aggregating the source models $\{f^{(l)}\}_{l\in [L]}$. The key distinction from the identification Theorem \ref{thm: identification} is that the noise levels $\bm{\sigma}^2$ across different source domains affect the aggregation weights of $f^{\rm sq}_{\HH}$. We demonstrate the disadvantage of such aggregation in the following corollary by considering a simpler regime with $L=2$ and $\HH=\Delta^2$.
\begin{Corollary}
   Suppose $L=2$, and $\HH=\Delta^2$. Then $f^{\rm sq}$ defined in \eqref{eq: sol - squared error} is identified as:
    \[
    f^{\rm sq} = q_1 \f{1} + (1-q_1) \f{2} \quad {\rm with}\quad q_1 = 0\vee \left(\frac{1}{2} + \frac{\sigma_1^2 - \sigma_2^2}{\E_\QQ[(\f{1}(X)-\f{2}(X))^2]}\right) \wedge 1.
    \]
Consequently, when $\sigma_1^2 \gg \sigma_2^2$, $f^{\rm sq} = \f{1}$; when $\sigma_1^2 =\sigma_2^2$, $f^{\rm sq} = \frac{1}{2}\f{1} + \frac{1}{2}\f{2}$.
\label{coro: squared error}
\end{Corollary} 
The above corollary shows that $f^{\rm sq}$ prioritizes the group with the higher noise level, especially when one group exhibits a dominant noise over the other. When both groups have the same noise level, then $f^{\rm sq}$ becomes a simple average.

We now introduce the distributionally robust model based on the regret function in the MSDA setting, where the following regret function has been introduced in the work \cite{agarwal2022minimax}
$${\rm Regret}_\TT(f) = \E_\TT[Y - f(X)]^2 - \inf_{f'\in \FF}\E_\TT[Y-f'(X)]^2.$$
The regret ${\rm Regret}_\TT(f)$ measures the excess risk of the model $f$ compared to the optimal model for the distribution $\TT$.
Given the set $\HH \subseteq \Delta^L$ and the function class $\FF$, we define
\begin{equation}
\begin{aligned}
    f^{{\rm reg}}_{\HH} := \argmin_{f\in \FF}\max_{\TT\in \C(\QQ_X,\HH)} {\rm Regret}_\TT(f).
\end{aligned}
\label{eq: MRO}
\end{equation}
The following proposition establishes the identification of $f^{\rm reg}_{\HH}$.
\begin{Proposition}
    Suppose that the function class $\FF$ is convex with $\f{l} \in \FF$ for all $l\in[L]$, and $\HH$ is a convex subset of $\Delta^L$. Then $f^{\rm reg}_\HH$ is equivalent to:
    \begin{equation}
        f^{\rm reg}_\HH = \argmin_{f\in \FF}\left\{r:~~ \max_{q\in \HH}\E_{\QQ_X}\left[\left(\sum_{l=1}^L q_l \f{l}(X) - f(X)\right)^2\right]\leq r\right\}.
        \label{eq: chebyshev center}
    \end{equation}
    Therefore, $f^{\rm reg}_\HH$ corresponds to the center of the smallest circle enclosing all functions of the form $\{\sum_{l=1}^L q_l \f{l}(\cdot)\;\vert \; q \in \HH\}$, known as the Chebyshev center. For general convex subsets $\HH \subseteq \Delta^L$, solving \eqref{eq: chebyshev center} is NP-hard.
    \label{prop: identification MRO}
\end{Proposition}

This proposition reveals that identifying $f^{\rm reg}_\HH$ reduces to solving the Chebyshev center problem. For an arbitrary convex set $\HH\subseteq \Delta^L$, 
computing the Chebyshev center is computationally intractable \cite{eldar2008minimax}, except in special cases such as polyhedral \cite{milanese1985optimal} or finite sets \cite{xu2003solution}. 
For example, when $\HH$ is defined as the intersection of multiple ellipsoids, solving the optimization problem in \eqref{eq: chebyshev center} becomes NP-hard \cite{xia2021chebyshev}. 
Therefore, the computational intractability limits the regret-based models' usefulness when integrating prior information via a convex set $\HH$. In contrast, our reward-based approach, introduced in Section \ref{subsec: identification} effectively incorporates prior information while maintaining computational efficiency. As demonstrated in Sections \ref{sec: simu-H} and \ref{sec: real data}, this prior integration leads to the improved predictive performance of the proposed approach on the target domain. 

We summarize the key distinctions among the distributionally robust models employing the three different loss functions in Table \ref{tab:comparison}.
Notably, the proposed reward function is the only one that produces a robust prediction model both independent of noise levels and computationally feasible when incorporating general convex prior mixture information.

\begin{table}[!htp]
\vspace{-1em}
    \centering
    \resizebox{0.9\textwidth}{!}{%
    \begin{tabular}{@{}lccc@{}}
        \toprule
        \textbf{Loss type} & \textbf{Squared Error} & \textbf{Regret} & \textbf{Reward (Ours)} \\ \midrule
        \textbf{Aggregation independent of noise levels} & $\times$ & $\checkmark$ & $\checkmark$ \\
        \textbf{Computational feasibility with arbitrary convex set $\mathcal{H}$}
        & $\checkmark$ & $\times$ & $\checkmark$ \\ \bottomrule
    \end{tabular}
    }
    \caption{Comparison of minimax optimization with respect to the choices of different loss functions}
    \label{tab:comparison}
    \vspace{-1.5em}
\end{table}

We now move on to considering two special cases of $\HH$ where the regret-based $f^{\rm reg}_\HH$ is computationally feasible and will provide further insights for this regret-based approach.
\begin{Corollary}
We consider two specifications of $\HH$ as follows:
\begin{itemize}
    \item[(a)] When $\HH = \Delta^{L}$, $f^{\rm reg}$ is identified as:
    \begin{equation}
        f^{\rm reg}= \sum_{l=1}^L q_l^{\rm reg} \cdot \f{l} \quad \textrm{with} \quad 
        q^{\rm reg} = \argmin_{q\in \Delta^L}\left\{q^\intercal \Gamma q - q^\intercal\gamma \right\},
        \label{eq: identification MRO}
    \end{equation}
where $\Gamma_{k,l} = \E_{\QQ_X}[\f{k}(X)\f{l}(X)]$ for $k,l\in [L]$, and $\gamma_l = \Gamma_{l,l}$ for $l\in [L].$
    \item[(b)] When $\HH$ is defined as
\begin{equation}
    \HH := \left\{\left.q\in \Delta^L \right\vert \E_{\QQ_X}\left[\sum_{l=1}^L(q_l - q^{\rm prior}_l)\f{l}(X)\right]^2\leq \rho\right\},
    \label{eq: H issue}
\end{equation}
where $q^{\rm prior}\in \Delta^L$ denotes a prior weight vector and $\rho\geq 0$ controls the size of $\HH$. Suppose $q^{\rm prior}$ lies strictly within $\Delta^L$ and $\rho$ is small enough such that $\HH$ does not intersect the boundary of $\Delta^L$. Then $f^{\rm reg}_\HH = \sum_{l\in [L]}q^{\rm prior}_l\f{l}$, which is independent of the value of $\rho$. 
\end{itemize}
\label{coro: MRO issue}
\end{Corollary}

\begin{figure}[H]
\vspace{-1em}
    \centering
    \includegraphics[width=0.6\linewidth]{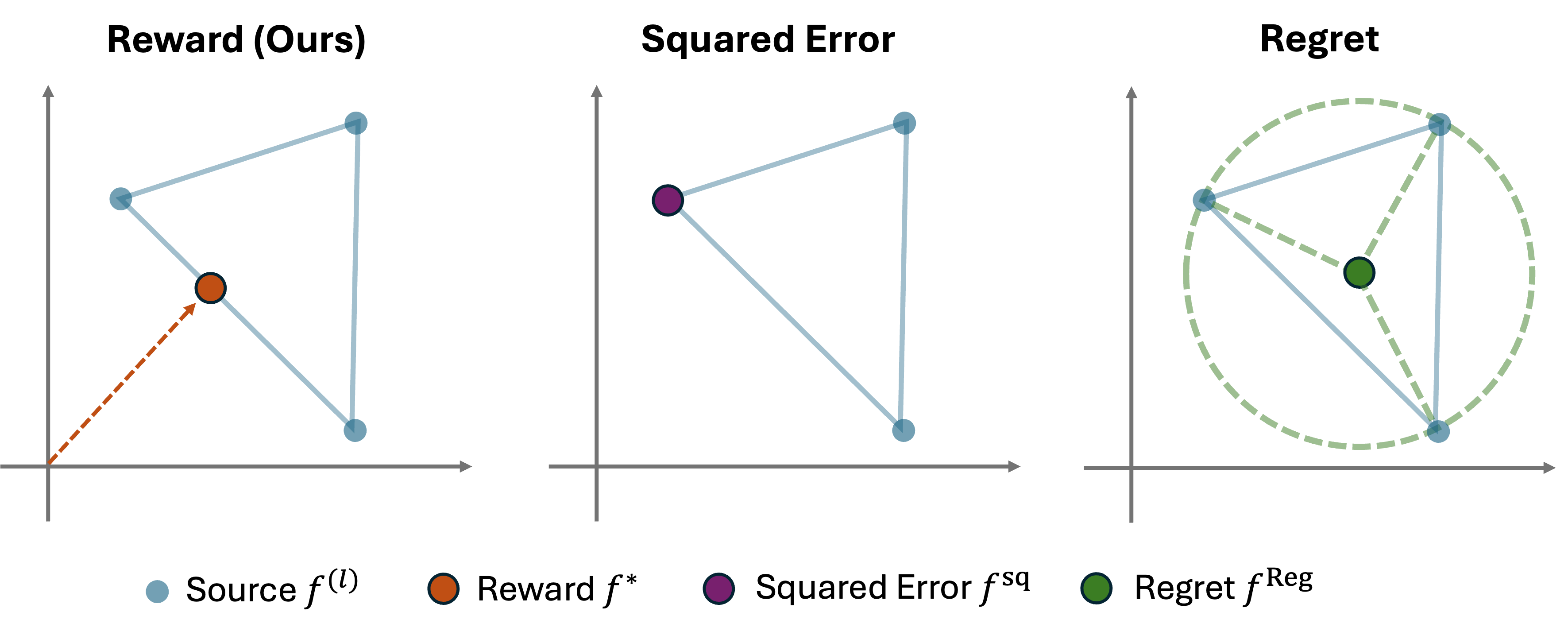}
    \vspace{-3.5mm}
    \caption{Illustration of robust prediction models utilizing reward (ours), squared error, and regret, for $L=3$ and $\HH=\Delta^3$. The left panel: $f^*$ is the point closest to the original within the convex hull of $\{\f{l}\}_{l\in[L]}$; The middle panel: $f^{\rm sq}$ corresponds to the source model with the largest noise level with the highest noise level when this noise is substantially higher than that in other sources; The right panel: $f^{\rm reg}$ is the center of the smallest circle enclosing all individual source models.}
    \label{fig:loss-compare}
    \vspace{-1.5em}
\end{figure}

Result (a) is closely related to the works of \cite{mo2024minimax} and \cite{zhang2024minimax}, who establish similar results for linear models and conditional average treatment effects in causal inference, respectively. \footnote{The result in \cite{zhang2024minimax} has been announced independently of ours at \url{https://arxiv.org/abs/2309.02211} (v3).} However, these works restrict the constraint set to simplexes or polyhedrons, whereas Proposition \ref{prop: identification MRO} shows that regret-based models become NP-hard for more general convex constraints.
For $\HH=\Delta^L$, we now compare its geometric interpretation with our proposed reward-based model $f^*$ and the squared-error based model $f^{\rm sq}$, that are established in Theorem \ref{thm: identification} and Proposition \ref{prop: squared error}, respectively.
To illustrate the main point, we consider a simple scenario with three source domains ($L=3$).
As depicted in Figure \ref{fig:loss-compare}, our proposed $f^*$ is the point closest to the origin within the convex hull of source models. In contrast, the model based on squared error is influenced by the heterogeneous noise levels across source domains - if one source exhibits significantly higher noise, $f^{\rm sq}$ essentially recovers the noisiest source's model. Meanwhile, the regret-based model $f^{\rm reg}$ corresponds to the center of the smallest circle that encloses all the source models.

In addition, for Result (b) in Corollary \ref{coro: MRO issue}, we consider that $\HH$ is specified as \eqref{eq: H issue}, which encodes the prior information that the conditional mean of the future target lies closely to the weighted source models $\sum_{l\in [L]} q_l^{\rm prior} \f{l}(X)$ for some pre-specified weights $q^{\rm prior}\in \Delta^L$.
Corollary \ref{coro: MRO issue} reveals that $f^{\rm reg}_\HH$ reduces exactly to $\sum_{l\in [L]} q_l^{\rm prior} \f{l}(X)$, independent of the radius $\rho$ of the set $\HH$. This implies that, in this case, the regret-based model $f^{\rm reg}_\HH$ is determined by the prior weights. However, in practice, $q^{\rm prior}$ may not be accurate, and one would expect the diameter $\rho$ to play a role in reflecting the uncertainty of this prior.

We conclude this subsection by observing that all of these loss or reward functions produce the same prediction model in the single-source setting, but they lead to different distributionally robust prediction models in the multi-source regime.

\subsection{Extensions of the Robust Models}
\label{subsec: extension}
We now extend the proposed robust prediction model $\fH$, defined in \eqref{eq: f_star H}, to a more general form. Previously, we characterized the uncertainty set \eqref{eq: region-Q H} via mixtures of the entire source conditional distributions. We relax that requirement and consider an uncertainty set defined only in terms of the conditional means. Specifically, we define
\begin{equation}
    \C^\prime(\QQ_X, \HH):= \left\{\left.\TT = (\QQ_X, \TT_{Y|X})\;\right\vert \; \E_{\TT}[Y|X] = \sum_{l=1}^L q_l \cdot \E_{\PP{l}}[Y|X],\quad q\in \HH  \right\}.
    \label{eq: region-Q H general}
\end{equation}
We observe that the newly defined uncertainty set  $\C^\prime(\QQ_X, \HH)$ is broader than the previous set $\C(\QQ_X, \HH)$ defined in \eqref{eq: region-Q H general}, with $\C(\QQ_X, \HH)\subseteq \C^\prime(\QQ_X, \HH)$. This is because $\C^\prime(\QQ_X,\HH)$ only requires a mixture of the conditional means rather than the entire conditional distributions. 

With a slight abuse of notation, we denote the robust prediction model based on this uncertainty set $\C^\prime(\QQ_X, \HH)$ also by $\fH$:
\[
\fH = \argmin_{f\in \FF} \max_{\TT \in C'(\QQ_X, \HH)} \RR_\TT(f).
\]
$\fH$ admits the same identification result as Theorem \ref{thm: identification}; see Supplement's Section A.1 \cite{wang2025supplements}.

Furthermore, while our reward function $\RR_\TT(f)$ was initially defined to evaluate the performance of the model $f(\cdot)$ relative to the null model, we extend the reward function to compare $f(\cdot)$ with any deterministic benchmark function. The identification theorem analogous to Theorem \ref{thm: identification} for this more general reward function is provided in Section A.1 of Supplements \cite{wang2025supplements}.

\section{Algorithms: Distributionally Robust Learning}
\label{sec: methodology}
In this section, we leverage the identification Theorem \ref{thm: identification} and devise an algorithm to estimate the robust prediction model $\fH$. For each source domain $l\in [L]$, we utilize the labeled data $\{\X{l}_{i}, \Y{l}_i\}_{i\in [n_l]}$ to fit the individual source model $\f{l}$ using a preferred ML method (e.g., random forests, boosting, or neural networks). The resulting fitted individual source model is denoted by $\hf{l}$. We then construct the plug-in estimator of the robust prediction model $\fH$ via aggregating these fitted individual source models as
\begin{equation}
    \tfH = \sum_{l=1}^L \widetilde{q}_l \cdot \hf{l}\quad \textrm{with} \quad \widetilde{q} = \argmin_{q\in \HH} q^\intercal \tGamma q,
    \label{eq: Magging plug-in}
\end{equation}  
where $\tGamma\in \R^{L\times L}$ is the data-dependent estimator of $\Gamma$ defined in \eqref{eq: identification} with $\tGamma_{k,l} = \frac{1}{N}\sum_{j=1}^{N} \hf{k}(\XQ_j)\hf{l}(\XQ_j)$ for $k,l\in [L]$.

The estimation error of $\tfH$ decomposes into two components: the error of estimating individual source models $\{\f{l}\}_{l\in [L]}$, and the error between the data-dependent aggregation weight $\widetilde{q}$ and the corresponding optimal weight $q^*$, defined in \eqref{eq: identification}. The error of estimating $\{\f{l}\}_{l\in [L]}$ can be minimized through tuning the ML methods used for each source domain. However, even with optimally tuned individual models, the error of estimating the optimal aggregation weight may still be significant. We shall propose in the following a bias-correction method for estimating the optimal aggregation weight, which further improves the estimation accuracy compared to simply plugging in machine learning prediction models.

The error of $\widetilde{q}$ depends on the difference $\widetilde{\Gamma}-\Gamma$ as indicated by \eqref{eq: Magging plug-in}. To improve the accuracy of weight $\widetilde{q}$, we introduce in the following sections a bias-correction step to improve the estimation of $\widetilde{\Gamma}$. Using the resulting bias-corrected estimator $\widehat{\Gamma}$, we construct the estimator for the robust prediction model $\fH$ as follows:
\begin{equation}
    \hfH = \sum_{l=1}^L \widehat{q}_l\cdot \hf{l} \quad \textrm{with} \quad \widehat{q} = \argmin_{q\in \HH} q^\intercal \hGamma q.
    \label{eq: Magging cs}
\end{equation}
We explain the main idea behind this bias-corrected approach in the following Section \ref{sec: main idea} and provide the detailed construction in Sections \ref{sec: no shift} (for the no covariate shift setting) and \ref{sec: covariate shift} (for the general covariate shift setting).  

\subsection{Bias Correction: Main Idea}
\label{sec: main idea}
We now explain the main idea behind the bias correction before providing the full details. For $k,l \in [L]$, the error of the plug-in estimator $\tGamma_{k,l}$ is decomposed as:
{\small
\begin{align*}
    \tGamma_{k,l} - \Gamma_{k,l} &= \left\{\frac{1}{N}\sum_{j=1}^{N}\hf{k}(\XQ_j)\hf{l}(\XQ_j) - \E_{\QQ_X}\left[\hf{k}(X)\hf{l}(X)\right]\right\} \\
    &\quad + \left\{\E_{\QQ_X}\left[\hf{k}(X)\hf{l}(X)\right] - \E_{\QQ_X}\left[\f{k}(X)\f{l}(X)\right]\right\}. 
    \label{eq: tGamma error decompose}
\end{align*}}
The first term on the right-hand side represents the finite-sample variability, vanishing at the parametric rate, whereas the second term captures the estimation error of the production $\hf{k}\hf{l}$. We further decompose the second term of the above decomposition as follows, 
{
\begin{equation}
\begin{aligned}
&\underbrace{\E_{\QQ_X}\left[\hf{k}(X)\left(\hf{l}(X)-\f{l}(X)\right)\right]}_{\textrm{Bias-kl}}+
    \underbrace{\E_{\QQ_X}\left[\hf{l}(X)\left(\hf{k}(X)-\f{k}(X)\right)\right]}_{\textrm{Bias-lk}} \\
    & - \underbrace{\E_{\QQ_X}\left[\left(\hf{k}(X)-\f{k}(X)\right)\left(\hf{l}(X) - \f{l}(X)\right)\right]}_{\textrm{Higher order bias}}.
\end{aligned}
\label{eq: error decomposition}
\end{equation}}
In this decomposition, the “Higher-order bias” term, being the product of $\widehat{f}^{(k)} - f^{(k)}$ and $\widehat{f}^{(l)} - f^{(l)}$, typically diminishes to zero at a faster rate and is negligible compared to the “Bias-kl” and “Bias-lk” terms, which are first-order error terms in either $\widehat{f}^{(k)} - f^{(k)}$ or $\widehat{f}^{(l)} - f^{(l)}$. In what follows, we discuss the intuitive idea for estimating the “Bias-kl” term; a similar approach can be applied to the “Bias-lk” term.

Suppose that the fitted individual models $\hf{k},\hf{l}$ are independent of the data $(X, Y)$ that is drawn from the distribution $(\QQ_X, \PP{l}_{Y|X})$. Then, taking the expectation yields
{
\[
\begin{aligned}
    \E_{(\QQ_X,\PP{l}_{Y|X})}\left[\hf{k}(X) \left(\hf{l}(X) - Y\right) \right] &=\E_{\QQ_X}\left[\hf{k}(X) \left(\hf{l}(X) - \E_{\PP{l}_{Y|X}}[Y|X]\right)\right]\\
    &= \E_{\QQ_X}\left[\hf{k}(X) \left(\hf{l}(X) - \f{l}(X)\right)\right],
\end{aligned}
\]}
which is exactly the "Bias-kl" term in \eqref{eq: error decomposition}.
Therefore, if we have access to paired data $\{X_i,Y_i\}_{i\in [m]}$ i.i.d. drawn from the distribution $(\QQ_X, \PP{l}_{Y|X})$ that is independent of fitted individual source models $\hf{k},\hf{l}$, we can estimate the ``Bias-kl'' term by computing 
{
\begin{equation}\frac{1}{m}\sum_{i=1}^{m}\hf{k}(X_i) \left(\hf{l}(X_i) - Y_i\right) \approx \textrm{Bias-kl}.
\label{eq: intuitive expression}
\end{equation}}
This provides an intuitive way to correct the bias of $\widetilde{\Gamma}$.

 We employ the sampling-splitting technique to ensure the independence between the fitted individual source models and the data used for bias correction as required in \eqref{eq: intuitive expression}. Specifically, for each source $l\in [L]$, we randomly partition the data $\{\X{l}_i, \Y{l}_i\}_{i\in [n_l]}$
into two disjoint subsets $\A_l$ and $\B_l$. Without loss of generality, we consider $\A_l = \{1,2,..., \lfloor n_l/2\rfloor \}$ and $\B_l = [n_l] \setminus \A_l$.
The models fitted on one subset are thereby independent of the data in the other subset. 
We denote the fitted individual source models obtained from data in $\A_l$ and $\B_l$ by $\hf{l}_\A$ and $\hf{l}_\B$, respectively. 
With a slight abuse of notation, we define the fitted individual source model $\hf{l}(\X{l}_i)$ for a sample $\X{l}_i$ from the $l$-th source domain and $\hf{l}(\XQ_j)$ for a sample from the target domain as follows:
\begin{equation}
    \hf{l}(\X{l}_i) = \begin{cases}
        \hf{l}_\A(\X{l}_i) & \textrm{if $i\in \B_l$} \\
        \hf{l}_\B(\X{l}_i) & \textrm{if $i\in \A_l$}
    \end{cases},
    \;\textrm{and}\quad
    \hf{l}(\XQ_j) = \frac{1}{2}\hf{l}_\A(\XQ_j) + \frac{1}{2}\hf{l}_\B(\XQ_j).
    \label{eq: sample-split hfl}
\end{equation}
This construction ensures that the function $\hf{l}(\cdot)$ is independent of the sample that the function is applied to, and such a property is essential for conducting the bias correction for generic machine learning algorithms.

In the following Subsections \ref{sec: no shift} and \ref{sec: covariate shift}, we build on the intuition from \eqref{eq: intuitive expression} to construct the bias-corrected estimator $\hGamma$ of $\Gamma$ for settings without and with covariate shifts, respectively. 

\subsection{No Covariate Shift Setting}
\label{sec: no shift}
We first consider the setting without covariate shift, where $\PP{l}_X=\QQ_X$ for all $l\in[L]$. In this case, each source domain's data $\{\X{l}_i,\Y{l}_i\}_{i\in [n_l]}$, drawn from $(\PP{l}_X, \PP{l}_{Y|X})$, are in fact i.i.d. samples from $(\QQ_X, \PP{l}_{Y|X})$. We then apply \eqref{eq: intuitive expression} to construct the bias-corrected estimator for $\Gamma$. Specifically, for $ k,l\in [L]$, we correct the plug-in estimator $\tGamma_{k,l} = \frac{1}{N}\sum_{j=1}^{N} \hf{k}(\XQ_j)\hf{l}(\XQ_j)$ by subtracting estimates of the dominant bias terms. The resulting bias-corrected estimator is $\hGamma_{k,l}=\tGamma_{k,l}-\widehat{\mathcal{D}}_{k,l} - \widehat{\mathcal{D}}_{l,k}$, where
\begin{equation}
\begin{aligned}
    \widehat{\mathcal{D}}_{k,l} &=\frac{1}{n_l}\sum_{i=1}^{n_l} \hf{k}(\X{l}_i)\left(\hf{l}(\X{l}_i) - \Y{l}_i \right),\\
    \widehat{\mathcal{D}}_{l,k} &=\frac{1}{n_l}\sum_{i=1}^{n_l} \hf{l}(\X{k}_i)\left(\hf{k}(\X{k}_i) - \Y{k}_i \right).
\end{aligned}
    \label{eq: Gamma-debias-NoCS}
\end{equation}
Here,
$\widehat{\mathcal{D}}_{k,l}$, $\widehat{\mathcal{D}}_{l,k}$ estimate the "Bias-kl", "Bias-lk" terms in \eqref{eq: error decomposition}, respectively.

\subsection{
Covariate Shift Setting}
\label{sec: covariate shift}
In the covariate shift scenario, the target covariate distribution \(\QQ_X\) may differ from the source distributions \(\{\PP{l}_X\}_{l\in[L]}\). Consequently, we do not have direct access to the data sampled from \((\QQ_X, \PP{l}_{Y|X})\) for bias correction as required by \eqref{eq: intuitive expression}. To address this, we adopt an importance weighting strategy, a well-established approach for handling covariate shifts \cite{gretton2009covariate, sugiyama2012density, menon2016linking}. The key idea is to assign weights to the source data
$\{\X{l}_i,\Y{l}_i\}_{i\in [n_l]}$ for each source $l\in [L]$, so that the reweighted source distribution closely resembles $(\QQ_X, \PP{l}_{Y|X})$. Then we leverage \eqref{eq: intuitive expression} to estimate the bias components with the reweighted source distribution. 

For each source $l\in[L]$, suppose that $\QQ_X$ is absolutely continuous with respect to $\PP{l}_X$ and define the density ratio $\w{l}(x) = d\QQ_X(x)/ d\PP{l}_X(x)$. A crucial step in our approach is to estimate these density ratios $\{\w{l}\}_{l\in[L]}$.
To ensure independence between the density ratio estimators and the data used for bias correction, we employ the sample-splitting technique, analogous to that used for constructing the individual source model estimators $\{\hf{l}\}_{l\in [L]}$ in \eqref{eq: sample-split hfl}. 
Specifically, we construct the density ratio estimator $\hw{l}_\A$ using the covariate data from the split $\A_l$ combined with the target covariate $\{\XQ_j\}_{j\in [N]}$, and $\hw{l}_\B$ using the data from $\B_l$ together with the target covariate data.
In Supplements' Section A.4 \cite{wang2025supplements},
we provide details on the construction of \(\widehat{\omega}^{(l)}_\A\) and \(\widehat{\omega}^{(l)}_\B\) via the Bayes formula. Our proposal is also compatible with alternative density ratio estimation methods \cite[e.g.,]{kanamori2009least, gretton2009covariate, nguyen2010estimating}.
We then define the sample-split density ratio estimator $\hw{l}(\X{l}_i)$ for a sample $\X{l}_i$ from the $l$-th source domain:
\begin{equation}
    \hw{l}(\X{l}_i) = \begin{cases}
        \hw{l}_\A(\X{l}_i) & \textrm{if \;$i\in \B_l$}\\
        \hw{l}_\B(\X{l}_i) & \textrm{if \;$i\in \A_l$}\\
    \end{cases}.
    \label{eq: sample-split hwl}
\end{equation}
This construction ensures that the estimated density ratio function $\hw{l}(\cdot)$ is independent of the sample that the function is applied to.
For $k,l\in [L]$, we construct the bias-corrected matrix estimator of the matrix $\Gamma$ as follows:
\begin{equation}
    \hGamma_{k,l} = \tGamma_{k,l} - \widehat{\mathcal{D}}_{k,l} - \widehat{\mathcal{D}}_{l,k},
    \label{eq: hGamma}
\end{equation} 
where the plug-in estimator remains $\tGamma_{k,l} = \frac{1}{N}\sum_{j=1}^{N} \hf{k}(\XQ_j)\hf{l}(\XQ_j)$, and the terms $\widehat{\mathcal{D}}_{k,l}$, $\widetilde{\mathcal{D}}_{l,k}$ are used to estimate "Bias-kl", "Bias-lk" terms in \eqref{eq: error decomposition}, respectively. In the covariate shift setting, these bias terms are defined as:
{\begin{equation}
\begin{aligned}
    \widehat{\mathcal{D}}_{k,l} &=\frac{1}{n_l}\sum_{i=1}^{n_l} \hw{l}(\X{l}_i)\hf{k}(\X{l}_i)\left(\hf{l}(\X{l}_i) - \Y{l}_i \right),\\
    \widehat{\mathcal{D}}_{l,k} &= \frac{1}{n_k}\sum_{i=1}^{n_k} \hw{k}(\X{k}_i)\hf{l}(\X{k}_i)\left(\hf{k}(\X{k}_i) - \Y{k}_i \right).
\end{aligned}
\label{eq: Gamma-debias-CS}
\end{equation}}
The above construction is analogous to the one in the no covariate shift setting (see \eqref{eq: Gamma-debias-NoCS}), with the key distinction that the source data $\{X^{(l)}_i, \Y{l}_i\}_{i\in [n_l]}$ and $\{X^{(k)}_i, \Y{k}_i\}_{i\in [n_k]}$ are reweighted by the corresponding density ratio estimators $\widehat{w}^{(l)}$ and $\widehat{w}^{(k)}$. In effect, the no covariate shift case is a special instance of this formulation with $\widehat{w}^{(l)}(\cdot) \equiv \widehat{w}^{(k)}(\cdot) \equiv 1$. 

\subsection{Algorithm}
\label{subsec: alg}
We summarize the proposed bias-corrected estimator for the robust prediction model \(\fH\) in the general covariate shift setting in Algorithm \ref{alg: cs mm}, and we refer to our methodology as \textbf{D}istributionally \textbf{Ro}bust \textbf{L}earning (DRoL).  
The plug-in version of the DRoL algorithm is provided in Supplements' Section A.5 \cite{wang2025supplements}
for completeness. Importantly, our DRoL algorithms are specifically designed for the MSDA setting with the unlabeled target covariate data, which aggregates the ML prediction models trained on all source domains. Our proposal may not apply to Group DRO settings, where target covariate data are unavailable.

\begin{algorithm}[H]

    \DontPrintSemicolon
    \SetAlgoLined
    \SetNoFillComment
    \LinesNotNumbered 
    \caption{Bias-corrected {DRoL}}
    \KwData{Labeled data $\{\X{l}_i, \Y{l}_i\}_{i\in [n_l]}$  on each source domain $l\in[L]$; {Unlabeled data $\{\XQ_j\}_{j\in [N]}$ on the target domain;}
    Prior knowledge about target mixture $\HH$ (with the default $\HH=\Delta^{L}$).}
    \KwResult{Bias-corrected {DRoL} estimator $\hfH$}

    \For{$l=1,...,L$}{
        Construct the sample-split individual source model $\hf{l}$ as in \eqref{eq: sample-split hfl};

        Construct the sample-split density ratio estimator $\hw{l}$ as in \eqref{eq: sample-split hwl};
    }

    \For{$k,l=1,...,L$}{
        Compute the bias estimators $\widehat{\mathcal{D}}_{k,l}$ and $\widehat{\mathcal{D}}_{k,l}$  as in \eqref{eq: Gamma-debias-CS};

        Compute the bias-corrected estimator $\hGamma_{k,l}$ as in \eqref{eq: hGamma};
    }
    
    Construct the data-dependent optimal aggregation weight as $\widehat{q} = \argmin_{q\in \mathcal{H}} q^\intercal \hGamma q;$

    Return $\widehat{f}_\HH = \sum_{l=1}^L \widehat{q}_l \cdot \hf{l}.$

    \label{alg: cs mm}
\end{algorithm}

Notably, Algorithm \ref{alg: cs mm} is designed to be privacy-preserving and can be viewed within the federated learning framework, as it does not require sharing raw data across different source domains. While the density ratio estimation step involves transmitting target covariate information to each source group, this requirement is eliminated in scenarios where the density ratios are already known (as in the no covariate shift setting). Overall, the DRoL algorithm protects data privacy by relying primarily on the trained prediction and density ratio models from each source rather than on the raw data. This approach complies with data privacy regulations such as the HIPAA Privacy Rule \cite{topaloglu2021pursuit} in the United States, which safeguards patient information and prohibits the sharing of patient-level data between sites.

\section{Theoretical Justification}
\label{sec: Theoretical}
In this section, we establish the convergence rates for the plug-in estimator $\tfH$ defined in \eqref{eq: Magging plug-in} and the bias-corrected estimator $\hfH$ defined in \eqref{eq: Magging cs}.
We start with the following condition. 
\begin{Assumption}
The number of source domains $L$ is finite. The matrix $\Gamma$ defined in \eqref{eq: identification} is positive definite with
    $\eigenGamma > 0$. There exists some positive constant $\sigma^2_{\varepsilon}>0$ such that
    $ \max_{l\in [L]}\max_{i\in [n_l]} \E[(\eps{l}_i)^2|\X{l}_i] \leq \sigma^2_{\varepsilon}.$
    \label{ass: eigen}
\end{Assumption}
We assume the finite count of source domains to simplify the presentation, although our analysis can be extended to allow for growing $L$. The requirement of $\Gamma$ being positive definite is ensured if the individual source models $\{\f{l}\}_{l\in [L]}$ are linearly independent. As for the noise level, we do not impose homogeneity across or within sources but only require its second moment to be bounded by some constant.

For a prediction model $f(\cdot)$, we define it $\ell_q$ with respect to the target covariate distribution $\QQ_{X}$ as $\|f\|_{\ell_q(\QQ)} := \left(\E_{\QQ_X}[f(X)^q]\right)^{1/q}$ for $q\geq 1$. We further define the scale of individual source models as
\begin{equation}
M := \max_{l\in[L]}\max\left\{\|\f{l}\|_{\ell_2(\QQ)} ,\;\|\f{l}\|_{\ell_4(\QQ)}\right\}.
\label{eq: f norm}
\end{equation}
Note that we do not assume $M$ to be bounded; it is allowed to grow with the dimension $p$. For instance, in a high-dimensional linear setting where \(f^{(l)}(x)=x^\intercal\beta^{(l)}\) with \(\beta^{(l)}\in\R^p\), we have \(M \asymp  \max_{l\in [L]}\|\beta^{(l)}\|_2\), which may increase with $p$.

For each source domain $l\in[L]$, given the labeled data, we obtain the fitted individual source model $\hf{l}$ using the ML method of choice.
Next, we assume that each $\hf{l}$ achieves a certain convergence rate in estimating the conditional mean model $\f{l}$; these rates will later be used to establish convergence results for our proposed DRoL estimators of the robust prediction model, as defined in \eqref{eq: Magging plug-in} and \eqref{eq: Magging cs}. Let $n = \min_{l\in[L]} n_l$ denote the smallest sample size among $L$ source domains. 
In the following assumption, we capture the convergence rate of the estimators $\{\hf{l}\}_{l\in [L]}$.

\begin{Assumption} With probability larger than $1-\tau_n$ with $\tau_n\rightarrow 0$, there exists a positive sequence $\delta_n>0$ such that $
\max_{l\in [L]}\max\{\|\hf{l} - \f{l}\|_{\ell_2(\QQ)}, \|\hf{l} - \f{l}\|_{\ell_4(\QQ)}\}\leq \delta_n. 
$ 
\label{ass: plug-in}
\end{Assumption}
Without further specification, we consider that each fitted individual model $\hf{l}$ is a consistent estimator of the true $\f{l}$ (i.e. $\delta_n\to 0$).
Consistency has been established for various ML methods. For example, in high-dimensional sparse linear regression, estimators such as the Lasso or the Dantzig selector satisfy Assumption \ref{ass: plug-in} with $\delta_n= C\sqrt{s_\beta \log p/n}$, where $s_\beta = \max_{l\in [L]}\|\beta^{(l)}\|_0$ denotes the largest sparsity level of $\beta^{(l)}$ for $l\in [L]$ \cite{{candes2007dantzig, bickel2009simultaneous}}. In the context of random forests, \cite{biau2008consistency, biau2012analysis, scornet2015consistency, meinshausen2006quantile} have established consistency results, with \cite{biau2012analysis} showing that under their conditions $\delta_n^2 = C\cdot n^{-0.75/ (S\log 2+0.75)}$, assuming that only $S$ out of the $p$ features play a role in the model. Similarly, \cite{schmidt2020nonparametric} and \cite{farrell2021deep} have demonstrated that certain neural networks satisfy Assumption \ref{ass: plug-in}; in particular, \cite{farrell2021deep} prove that if $f^{(l)}$ belongs to a Sobolev ball $\mathcal{W}^{\alpha, \infty}([-1,1]^p)$ with smoothness level $\alpha$, then one may take $\delta_n^2 = C\cdot(n^{-\alpha/(\alpha+p)} \log^8 n + \log\log n/n)$.

The following theorem establishes the convergence rate for the plug-in DRoL estimator $\tfH$ as defined in \eqref{eq: Magging plug-in}.
\begin{Theorem} Suppose Assumptions \ref{ass: eigen} and \ref{ass: plug-in} hold. With probability larger than $1-{1}/{t^2}- \tau_n$ for $t>1$ {and $\tau_n\to 0$}, 
the plug-in estimator defined in \eqref{eq: Magging plug-in} satisfies:
{   \begin{equation}
        \begin{aligned}
            \left\|\tfH - \fH\right\|_{\ell_2(\QQ)} &\lesssim \delta_n + t M \cdot \min\left\{\delta_n^2 + \frac{M^2}{\sqrt{N}} + \Err_0, \;\rho_\HH\right\},\; \textrm{with}\quad \Err_0 = M\delta_n,
        \end{aligned}
    \label{eq: error plug-in}
    \end{equation}}
where $M$ is the scale of the individual source models as defined in \eqref{eq: f norm}, $\delta_n$ measures the estimation errors of fitted individual models defined in Assumption \ref{ass: plug-in}, and $\rho_\HH := \max_{q, q' \in \HH} \|q - q'\|_2$ is the diameter of the set $\HH$ that encodes the prior information.
    \label{thm: Magging plug-in}
\end{Theorem}

As shown in \eqref{eq: error plug-in}, the convergence rate of $\|\tfH - \fH\|_{\ell_2(\QQ)}$ consists of two parts. The first term, $\delta_n$ stems from the estimation error in each fitted individual model $\hf{l}$.  
The second term $\min\left\{\delta_n^2 + \frac{M^2}{\sqrt{N}} + \Err_0, \rho_{\HH}\right\}$ captures the error in estimating the optimal aggregation weights.  The intuition behind taking the minimum is as follows: if the diameter \(\rho_{\HH}\) of the set \(\HH\) is relatively large, the error in the aggregation weight mainly comes from the uncertainty in estimating $\Gamma$; however, if \(\rho_{\HH}\) is smaller than the statistical precision achieved in estimating the weights, then the constraint imposed by the set \(\HH\) itself limits the error.

For clarity, when $\delta_n\ll M$ -- that is, when the estimation error of $\hf{l}$ is small relative to the scale of $\f{l}$ -- and when the constraint set \(\HH\) is sufficiently large so that \(\rho_\HH\) does not impose any restriction when computing the aggregation weight, Theorem \ref{thm: Magging plug-in} simplifies as
\begin{equation}
\left\|\tfH - \fH\right\|_{\ell_2(\QQ)} \lesssim \delta_n+ M^2\delta_n+\frac{M^3}{\sqrt{N}}.
\label{eq: sim plug-in}
\end{equation}
In cases where $M$ is large (e.g., in high-dimensional settings where $M = \max_{l\in [L]}\| \beta^{(l)}\|_2$ grows with the dimension $p$), the second term dominates the convergence rate. In the following Theorem \ref{thm: Magging correct}, we show that the proposed bias-corrected estimator can substantially reduce this dominating term, $M^2\delta_n$. 

\subsection{Rate Improvement with Bias Correction}
\label{sec: theory-correct}
We now focus on the bias-corrected DRoL estimator \(\widehat{f}_{\HH}\) and show that the bias correction leads to an improved convergence rate relative to its plug-in counterpart. To facilitate this analysis, we introduce convergence rate assumptions for the density ratio estimators.

\begin{Assumption}
There exist positive constants $c_1, c_2>0$ such that $\w{l}(x)\in [c_1,c_2]$ for all $l\in [L]$ and $x\in \R^p$. Furthermore, with probability at least $1-\tau_n$ with $\tau_n\rightarrow 0,$ there exists a positive sequence ${\eta}_{\omega}$ such that for all $\max_{l\in [L]}\|\hw{l}-\w{l} \|_{\ell_4(\QQ)}\leq {\eta}_{\omega}$.
\label{ass: correct}
\end{Assumption}

The first part of Assumption \ref{ass: correct} imposes a boundedness condition on $w^{(l)}$, which is standard in the density ratio estimation literature (see, e.g., Assumptions 13.1 and 14.7 in \cite{sugiyama2012density}). This condition is also consistent with the positivity (or overlap) assumption commonly invoked in causal inference \cite[e.g.]{rosenbaum1983central, chen2008semiparametric}. 
The second part captures the convergence rate of the density ratio estimators $\{\hw{l}\}_{l\in [L]}$ via the sequence $\eta_\omega$, where $\eta_\omega$ may or may not converge to $0$.

Here, we give concrete examples of $\eta_\omega$.
In the special setting with no covariate shift, we set $\hw{l}\equiv\w{l}(\cdot)\equiv1$ so that $\eta_\omega=0$.
In high-dimensional regimes with covariate shift, we further establish that $\eta_\omega\lesssim \sqrt{{s_{\gamma}\log p}/{(n+N)}}$, where $s_{\gamma} = \max_{l\in [L]}\|\gamma^{(l)}\|_0$ denotes the maximum sparsity for the high-dimensional logistic density ratio model; see Corollary 3
in Section A.4 of Supplements \cite{wang2025supplements}.
Moreover, we present a relaxed version of Assumption \ref{ass: correct} in Section C.5 of Supplements \cite{wang2025supplements}, where instead of assuming the pointwise boundedness of $\omega^{(l)}$, we control the convergence rate of the relative error \(\|\widehat{\omega}^{(l)}/\omega^{(l)} - 1\|_{\ell_4(\QQ)}\).

We now present the convergence rate for the bias-corrected DRoL estimator $\hfH$.
\begin{Theorem} 
Suppose Assumptions \ref{ass: eigen}, \ref{ass: plug-in}, and \ref{ass: correct} hold. With probability at least $1-{1}/{t^2} - 2\tau_n$ for $t>1$ and $\tau_n\to 0$, the estimator $\hfH$ defined in \eqref{eq: Magging cs} satisfies:
{   \begin{equation}
    \begin{aligned}
    \left\|\hfH - \fH\right\|_{\ell_2(\QQ)} \lesssim \delta_n + t M\cdot \min\left\{\delta_n^2 + \frac{M^2}{\sqrt{N}} + \Err_1 + \Err_2,\; \rho_\HH\right\},\\
    \textrm{with}\quad \Err_1 = \frac{(M+\delta_n)\delta_n}{\sqrt{n \wedge N}} + \frac{M+\delta_n}{\sqrt{n}},\quad
    \Err_2 = (M+\delta_n)(\delta_n + \frac{1}{\sqrt{n}})\eta_\omega,
    \end{aligned} 
    \label{eq: error correct}
    \end{equation}}
where $M$ is the scale of the individual source models defined in \eqref{eq: f norm}, $\delta_n$ and $\eta_\omega$ measure the estimation errors of the fitted individual models and the density ratio estimators defined in Assumptions \ref{ass: plug-in} and \ref{ass: correct}, respectively, and \(\rho_{\HH} := \max_{q, q' \in \HH} \|q - q'\|_2\) is the diameter of the set \(\HH\) that encodes the prior information.
    \label{thm: Magging correct}
\end{Theorem}

In comparison to Theorem \ref{thm: Magging plug-in}, the bound here replaces the term \(\Err_0=M\delta_n\) in \eqref{eq: error plug-in} with \(\Err_1 + \Err_2\). The term \(\Err_1\) represents the bias estimation error with a known density ratio, while \(\Err_2\) captures the bias estimation error due to the density ratio estimation. Generally, $\Err_1 + \Err_2$ is smaller than $\Err$, indicating the effectiveness of bias correction.

Next, we simplify this theorem and discuss how the bias-corrected \(\widehat{f}_{\HH}\) achieves a smaller estimation error compared to the plug-in estimator. We consider the scenario where $\delta_n\ll M$ (i.e., the estimation error of $\widehat{f}^{(l)}$ is small relative to the scale of $f^{(l)}$) and the constraint set \(\HH\) is sufficiently large so that \(\rho_{\HH}\) does not impose a restriction on the aggregation weight estimation. Then, Theorem \ref{thm: Magging correct} simplifies to
\begin{equation*}
    \left\|\hfH - \fH\right\|_{\ell_2(\QQ)} \lesssim \delta_n+ M^2\delta_n \cdot (r_n+\eta_{\omega})+\frac{M^3}{\sqrt{N}},\; \textrm{with}\quad r_n=\frac{1}{\sqrt{n \wedge N}}+\frac{\delta_n}{M}+\frac{1}{\delta_n\sqrt{n}}.
\end{equation*}
By comparing this rate to that of the plug-in estimator in \eqref{eq: sim plug-in}, we observe that the bias-correction step effectively shrinks the term $M^2\delta_n$ by the multiplicative factor $r_n + \eta_\omega$. Notably, $r_n$ diminishes to zero whenever $\delta_n/M\rightarrow 0$ and the convergence rate of the individual estimators $\{\widehat{f}^{(l)}\}_{l\in[L]}$ is slower than $n^{-1/2}$, which holds for most ML estimators, including high-dimensional regression, random forests, and deep neural networks. Furthermore, if the density ratio estimators are consistent with $\eta_\omega\to 0$, the overall multiplicative factor $r_n + \eta_\omega$ vanishes, leading to a strictly improved convergence rate.  In the special case of no covariate shift, where there is no need to estimate density ratios (i.e., $\eta_\omega = 0$), the bias‐corrected estimator \(\widehat{f}_{\HH}\) improves precisely by the factor $r_n$.

Notably, even if the density ratio estimators are inconsistent, with $\eta_\omega$ remaining at a constant level, the factor $r_n + \eta_\omega$ is still controlled, ensuring that the bias-corrected estimator \(\widehat{f}_{\HH}\) is at least as good as the plug-in estimator. This finding highlights the robustness of our proposed bias‐correction approach with respect to potential mis-specification in the density ratio model. In summary, our analysis demonstrates that the bias-correction step yields an estimator with a smaller estimation error, and it attains a strictly better rate when the density ratio estimators are consistent. We also illustrate the numerical advantages of the bias-corrected estimator in Section \ref{sec: simu-cs}, where it consistently outperforms the naive plug-in estimator. 

In the discussion above, we quantified the estimation error of the DRoL estimator $\hfH$ in terms of its $\ell_2(\QQ)$ distance to the actual $\fH$. We now translate this result to the objective function by quantifying the difference in rewards.
\begin{Theorem}
    For the target distribution $\QQ$ with the conditional outcome model defined as $f^\QQ(X^\QQ):=\E[Y^{\QQ}\mid X^{\QQ}]$, we have
    \[
    \left|\RR_\QQ(\hfH)-\RR_{\QQ}(\fH) \right|\leq 2\|f^\QQ-\fH\|_{\ell_2(\QQ)} \cdot \|\hfH - \fH\|_{\ell_2(\QQ)} + \|\hfH - \fH\|_{\ell_2(\QQ)}^2,
    \]
where the reward function $\RR_{\QQ}(\fH)$ is defined in \eqref{eq: reward}.
\label{prop: reward-diff}
\end{Theorem}
This theorem shows that the convergence rate of the reward difference \(|\RR_\QQ(\widehat{f}_{\HH}) - \RR_{\QQ}(f_{\HH})|\) is directly determined by the convergence rate of \(\|\widehat{f}_{\HH} - f_{\HH}\|_{\ell_2(\QQ)}\). In particular, if \(\widehat{f}_{\HH}\) is a consistent estimator of \(f_{\HH}\) and the distance \(\|f^\QQ - f_{\HH}\|_{\ell_2(\QQ)}\) is bounded by a constant, then the reward of our bias-corrected estimator converges to the reward of the population-level model \(f_{\HH}\). It is worth noting that Theorem \ref{prop: reward-diff} also applies to the plug-in DRoL estimator (with \(\widehat{f}_{\HH}\) replaced by \(\widetilde{f}_{\HH}\)).

Theorems \ref{thm: Magging plug-in}, \ref{thm: Magging correct}, and \ref{prop: reward-diff} together provide a theoretical framework for analyzing the proposed robust prediction model estimators, addressing the estimation error in both the model parameters and the objective function value. This framework can be applied to concrete settings by verifying the estimation errors of individual source models and density ratio estimators, as specified in Assumptions \ref{ass: plug-in} and \ref{ass: correct}. In particular, in Section A.4 of Supplements \cite{wang2025supplements},
we achieve a rate of $n^{-1/2}$ in the context of low-dimensional linear regression and a rate of $\sqrt{s\log p/n}$ in high-dimensional settings (with $s$ denoting the sparsity level).

\section{Simulations}
\label{sec: simus}
We evaluate the effectiveness of our proposed DRoL method through a series of simulation studies\footnote{The code for replicating results in Sections 5 and 6 is available at \url{https://github.com/zywang0701/DRoL}.}. The experiments are organized as follows. 
In Section \ref{sec: simu-varyL}, we compare DRoL with various methods, including the ERM approach, the unsupervised domain adaptation technique, and Group DRO, in terms of the worst-case reward.
In Section \ref{sec: simu-H}, we investigate the impact of different specifications of prior information on DRoL’s performance.
Finally, Section \ref{sec: simu-cs} demonstrates the advantages of our bias-corrected estimator compared to a plug-in estimator that directly utilizes ML algorithms without bias correction.

Next, we describe the simulated data setup. The simulation involves generating labeled data from $L$ source domains and unlabeled data from the target domain. For each source domain $l\in [L]$, the labeled data $\{\X{l}_i, \Y{l}_i\}_{i\in [n_l]}$ are i.i.d. generated as follows:
{\small
\[
\X{l}_i\sim \N(0_5, I_5),\quad \Y{l}_i = \f{l}(X_i) + \varepsilon^{(l)}_i, \quad \textrm{with}\quad \varepsilon^{(l)}_i\sim \N(0, 1).
\]}
We now specify the form of the individual source model $\f{l}$ for each $l\in [L]$:
\begin{equation}
    \f{l}(x) = \sin(x^\intercal \beta^{(l)}) + x^\intercal A^{(l)} x - {\rm Tr}(A^{(l)}),
    \label{eq: fl simus}
\end{equation}
where $\beta^{(l)}\in \R^5$ is generated by drawing each entry independently from ${\rm Unif}(-1,1)$ and $A^{(l)}\in \R^{5\times 5}$ is generated by drawing each entry independently from ${\rm Unif}(-0.5, 0.5)$.
These parameters $\beta^{(l)}$ and $A^{(l)}$, for $l\in [L]$, are generated randomly once and then fixed throughout the 200 simulation rounds, ensuring consistency across replications. For the target domain, the unlabeled data $\{\XQ_j\}_{j\in [N]}$ are i.i.d. drawn from $\QQ_X=\N\left((1,-1,0.5, 0, 0)^\intercal, I_5\right)$.

\subsection{Comparison of Worst-case Reward} 
\label{sec: simu-varyL}

We evaluate our proposed DRoL approach -- implemented via Algorithm \ref{alg: cs mm} with the default \(\HH=\Delta^L\) -- against several alternative methods by comparing their worst-case rewards over all distributions in the uncertainty class \(\C(\QQ_X)\), defined in \eqref{eq: region-Q full}. Below is a brief description of each method. Except for \texttt{GroupDRO} (which has to be implemented via gradient-based neural networks), all methods, including our DRoL approach, are implemented using XGBoost with 5-fold cross-validation. We note that, unlike standard approaches that evaluate performance based on squared error loss, we focus on the reward function; the rationale is detailed in Section \ref{subsec: other losses}. 
\begin{itemize}
    \item \texttt{ERM}. This method pools all source data and maximizes the empirical reward:
    {\small
    $$ \argmax_{f\in \FF} \frac{1}{\sum_{i=1}^L n_l}\sum_{l=1}^L \sum_{i=1}^{n_l}\left[(\Y{l}_i)^2 - (\Y{l}_i-f(\X{l}_i))^2\right].$$}
    \item \texttt{ImpWeight}. A common technique in unsupervised domain adaptation \cite{sugiyama2007direct, gretton2009covariate, nguyen2010estimating} is given by reweighting
    each $l$-th source data sample by an estimated $\hw{l}(x)$ of importance (density) ratio $\omega^{(l)}(x)={d\QQ_X(x)}/{d\PP{l}(x)}$. In our implementation, we estimate these density ratios using logistic regression, and then maximize the empirical reward over the pooled, reweighted data:
    {\small
    $$
    \argmax_{f\in \FF} \frac{1}{\sum_{l=1}^L \sum_{i=1}^{n_l}\widehat{\omega}^{(l)}(\X{l}_i)}\sum_{l=1}^L \sum_{i=1}^{n_l}\left\{\widehat{\omega}^{(l)}(\X{l}_i)\left[(\Y{l}_i)^2 - (\Y{l}_i-f(\X{l}_i))^2\right]\right\}.
    $$}
    \item \texttt{GroupDRO}. This approach maximizes the worst-case (group-wise) empirical reward across the source domains:
    {\small
    $$
    \argmax_{f\in \FF}\min_{1\leq l\leq L} \frac{1}{n_l}\sum_{i=1}^{n_l}\left[(\Y{l}_i)^2 - (\Y{l}_i-f(\X{l}_i))^2\right],
    $$}
    implemented via the nested gradient-based algorithm for Group DRO, as detailed in Algorithm 1 of the work \cite{sagawa2019distributionally}.
\end{itemize}

\begin{figure}[htp!]
\vspace{-1em}
    \centering
    \includegraphics[width=0.65\textwidth]{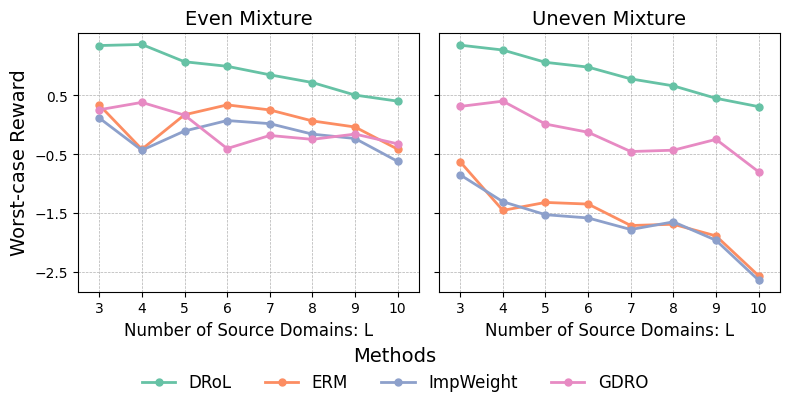}
    \vspace{-3.5mm}
    \caption{
    Comparison of worst-case reward for \texttt{DRoL}, \texttt{ERM}, and \texttt{ImpWeight}, and \texttt{GroupDRO} with the number of source domains $L$ varied across $\{3,...,10\}$. 
    The left panel corresponds to even mixture scheme, where the source domain sample size is $n_l=1000$ for all $l\in [L]$. The right panel stands for the uneven mixture scheme, with $n_1=500\cdot L$ for the first source, and $n_l=500$ for all $l\geq 2$.
    } 
    \vspace{-1.5em}
    \label{fig: compare_mixture}
\end{figure}

We consider scenarios with a varying number of source domains, where $L$ ranges in $\{3,4,..., 10\}$. In all experiments, the sample size for the unlabeled target data is fixed at $N=20,000$. In contrast, the source domain sample sizes follow two schemes: (i) Even Mixture: $n_l=1000$ for all $l\in [L]$; (ii) Uneven Mixture: $n_1=500\cdot L$ for the first source, and $n_l=500$ for all $l\geq 2$. 
The results based on 200 simulation rounds are presented in Figure \ref{fig: compare_mixture}. 
Recall that a higher reward indicates better predictive performance.
Our proposed \texttt{DRoL} method consistently achieves better worst-case reward than the other three methods across all values of $L$.

Comparing the left and right panels of Figure \ref{fig: compare_mixture}, we observe that the prediction models produced by \texttt{ERM} and \texttt{ImpWeight} vary markedly between the even and uneven mixture schemes. This suggests that their performance is highly sensitive to the relative proportions of each source’s sample size when the data are pooled. Moreover, even though \texttt{ImpWeight} leverages unlabeled target data, unlike \texttt{ERM}, their worst-case rewards over the uncertainty class \(\C(\QQ_X)\) are quite similar. In contrast, both \texttt{GroupDRO} and our \texttt{DRoL} produce robust prediction models that perform consistently under both the even and uneven mixture schemes.
Importantly, since \texttt{GroupDRO} is designed for a different setting where target data are not available, and as a result, it does not incorporate unlabeled target data into its formulation. Consequently, its robust prediction model is less tailored to the target domain compared to that of \texttt{DRoL}, which leverages the unlabeled target data to improve performance.

\subsection{Impact of Prior Information} 
\label{sec: simu-H}
We investigate how different specifications of the constraint set \(\HH\) -- which encodes prior knowledge about the target distribution -- affect the predictive performance of our DRoL approach. In practice, users may determine the constraint \(\HH\) based on their domain expertise. In this subsection, we assume that in addition to the unlabeled target data $\{\XQ_j\}_{j\in [N]}$, a small set of labeled target samples \(\{X_k^{\QQ}, Y_k^{\QQ}\}_{k\in N_0}\) (with $N_0 \ll N$) is available. 
Now, we describe how the labeled target data is utilized in our experiment. First, for each source domain, we fit the individual source model $\widehat{f}^{(l)}$. Next, we determine the optimal aggregation of these fitted individual models that fits the limited labeled target data. The optimal weight vector, denoted by $\widehat{q}^{\rm label}$, is obtained via
{\small
\begin{equation}
    \widehat{q}^{\rm label} = \argmin_{q\in \Delta^L} \frac{1}{N_0}\sum_{k=1}^{N_0} \left[\YQ_k - \sum_{l=1}^L q_l \cdot \hf{l}(\XQ_k)\right]^2.
    \label{eq: gamma prior}
\end{equation}
}
We incorporate this information by defining the constraint set
$\mathcal{H} = \{q\in \Delta^L \mid \|q - \widehat{q}^{\rm label}\|_2\leq \rho\}$,
where $\rho\geq0$ is the parameter controlling the size of $\HH$. We then run Algorithm \ref{alg: cs mm} using this constraint set and refer to the resulting method as \texttt{DRoL-Label}.
We emphasize that, in the absence of specific domain expertise, this procedure represents one practical way to derive the prior information $\mathcal{H}$ from a limited amount of labeled target data. More formal approaches that leverage small amounts of labeled target data with theoretical guarantees are provided in \cite{xiong2023distributionally}.

In the following experiments, we assume that the target conditional mean satisfies
$\E_{\QQ}[Y|X] = \sum_{l=1}^L q_l^\QQ\cdot \f{l}(X)$, 
where $q^\QQ\in \Delta^L$ is the (unknown) true mixture weight vector with $q^\QQ = (0.6, \frac{0.4}{3},\frac{0.4}{3}, \frac{0.4}{3})^\intercal$. The small set of labeled target samples  \(\{X_k^{\QQ}, Y_k^{\QQ}\}_{k\in N_0}\) are generated as follows:
{\small
\[
\XQ_k\sim \QQ_X,\quad \YQ_k= \sum_{l=1}^L \gamma_l^\QQ\cdot \f{l}(\XQ_k) + \varepsilon^\QQ_k
,\quad
\textrm{with} \quad\varepsilon^\QQ_k\sim \N(0,1),
\]} where $\varepsilon^\QQ_k$ is independent of $X_k^\QQ$ and the individual source models $\{\f{l}\}_{l\in [L]}$ are specified in \eqref{eq: fl simus}. 
We vary the labeled target data sample size $N_0 \in \{20, 50, 100\}$ and the parameter $\rho \in [0, 0.9]$, while fixing the number of source domains at $L=4$, each with a sample size of $n_l = 1000$, and the unlabeled target data sample size at $N=20,\!000$. 
Additionally, we compare \texttt{DRoL-Label} with the following methods, each method evaluated by the reward on the target distribution $\QQ$.
\begin{itemize}
    \item \texttt{DRoL-Unif}. This method constructs the constraint set around the uniform weight vector $q^{\rm unif} = (0.25, 0.25, 0.25, 0.25)^\intercal$ i.e., $\HH = \{q\in \Delta^L\mid \|q - q^{\rm unif}\|_2\leq \rho\}$.
    \item \texttt{DRoL-Default}. Without prior information, the proposed DRoL sets $\HH = \Delta^L$ as a default.
    \item \texttt{TargetOnly}.  This approach fits a prediction model using only the limited labeled target data, without incorporating any source domain data.
\end{itemize}
\begin{figure}[htp!]
\vspace{-1em}
    \centering
    \includegraphics[width=0.9\textwidth]{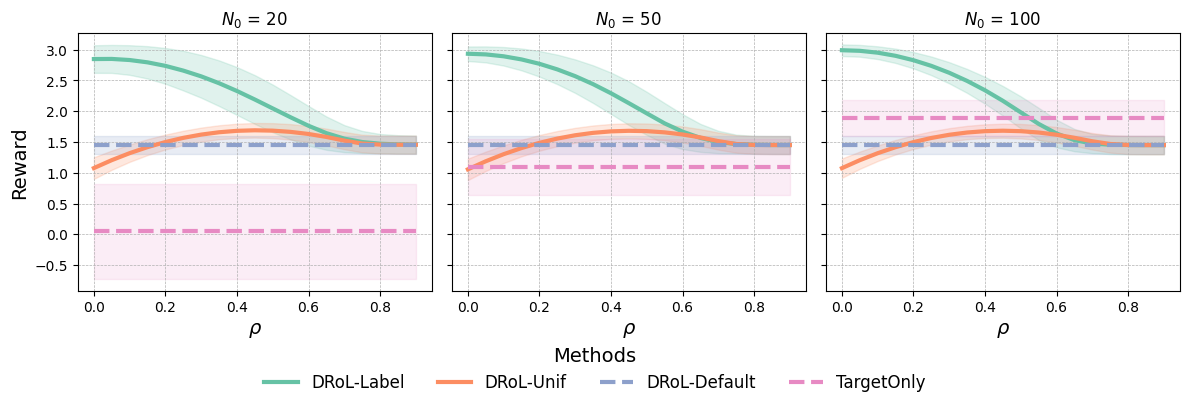}
    \vspace{-3.5mm}
    \caption{
    Comparison of different methods in terms of the reward evaluated on the target distribution \(\QQ\). The plotted curves represent the mean reward computed over 200 simulation rounds, while the shard error bands indicate the 10th and 90th percentile variability across these rounds. Here, the target conditional outcome model \(\QQ_{Y|X}\) is generated as a mixture of the source conditional outcome models \(\{\PP{l}_{Y|X}\}_{l\in [4]}\) with mixture weights \(\gamma^\QQ = \left(0.6,\frac{0.4}{3},\frac{0.4}{3},\frac{0.4}{3}\right)^\intercal\). The experiment fixes the number of source domains at $L=4$ (with $n_l=2000$ samples per source) and the unlabeled target sample size at $N=20,\!000$. We vary the number of labeled target samples $N_0 \in \{20,\, 50,\, 100\}$ (which are used only in \texttt{DRoL-Label} and \texttt{TargetOnly}) and the parameter $\rho \in [0, 0.9]$ that controls the size of the constraint set \(\HH\).
    } 
    \label{fig: prior}
    \vspace{-1.5em}
\end{figure}

Figure \ref{fig: prior} presents the results of 200 simulation rounds. The plotted curves depict the mean reward across these rounds, and the shaded error bands indicate the variability between the 10th and 90th percentiles.  We begin by comparing \texttt{DRoL-Label} with \texttt{DRoL-Default}. The results clearly demonstrate that even a small amount of labeled target data provides valuable insights into the target mixture, enabling \texttt{DRoL-Label} to consistently outperform \texttt{DRoL-Default} across all values of $N_0$ and $\rho$. As $\rho$ increases and the constraint set \(\HH\) expands to the full simplex $\Delta^L$, the performance of \texttt{DRoL-Label} is similar to that of \texttt{DRoL-Default}.

Next, we examine \texttt{DRoL-Unif},
which simulates a situation where one naively assumes that the target domain is an equally weighted average of the source domains. When the constraint set \(\HH\) is tight (i.e., when $\rho$ is small), this naive prior centered at ${\gamma}^{\rm unif}$ leads to poorer performance compared to \texttt{DRoL-Default} (which does not use any prior information). However, as $\rho$ increases and the constraint set expands, the negative impact of the inaccurate prior diminishes, and the performance of \texttt{DRoL-Unif} gradually improves until it eventually converges to that of \texttt{DRoL-Default}. This pattern highlights that incorporating a mistakenly specified prior can degrade the performance of DRoL, underscoring the importance for practitioners to carefully set the constraint \(\HH\) to balance predictiveness and robustness.

We also include the method \texttt{TargetOnly} as a benchmark, which builds the prediction model solely from the limited labeled target data. Our experiments reveal that when labeled target samples are too few to fit a satisfactory model (e.g., $N_0=20$ or $50$), the proposed DRoL methods substantially outperform \texttt{TargetOnly}. However, as $N_0$ increases to $100$, the performance of \texttt{TargetOnly} 
starts to outperform \texttt{DRoL-Default} (which does not incorporate prior information). Nonetheless, \texttt{DRoL-Label}, which effectively combines information from both the source domains and the labeled target data, 
delivers better performance compared to \texttt{TargetOnly} for our considered range of $N_0$.

\subsection{Effectiveness of Bias Correction}
\label{sec: simu-cs}
We assess the impact of bias correction in our proposed DRoL Algorithm \ref{alg: cs mm} by comparing it with that of a plug-in approach that simply aggregates the ML-fitted individual source models without bias correction. In our experiments, we fix the number of source domains at $L=3$ and vary the source sample sizes $n_l$ over $\{100, 300, 500, 1000\}$, while keeping the unlabeled target sample size fixed at $N = 20,\!000$. We compare three variants of DRoL, all using the default constraint set \(\HH = \Delta^L\):
\begin{itemize}
    \item \texttt{PlugIn}: Aggregates the individual source models with the aggregation weights computed directly from the ML-fitted models, without any bias correction, as described in \eqref{eq: Magging plug-in}.
    \item \texttt{Logistic}: Estimates the density ratios $\{\hw{l}\}_{l\in [L]}$ using logistic regression, and employs these estimates for bias correction, as in Algorithm \ref{alg: cs mm}.
    \item \texttt{Oracle}: Uses the true density ratios $\{\omega^{(l)}\}_{l\in [L]}$ for bias correction, as in Algorithm \ref{alg: cs mm}.
\end{itemize}

\begin{figure}[!ht]
\vspace{-1em}
    \centering
\includegraphics[width=0.85\textwidth]{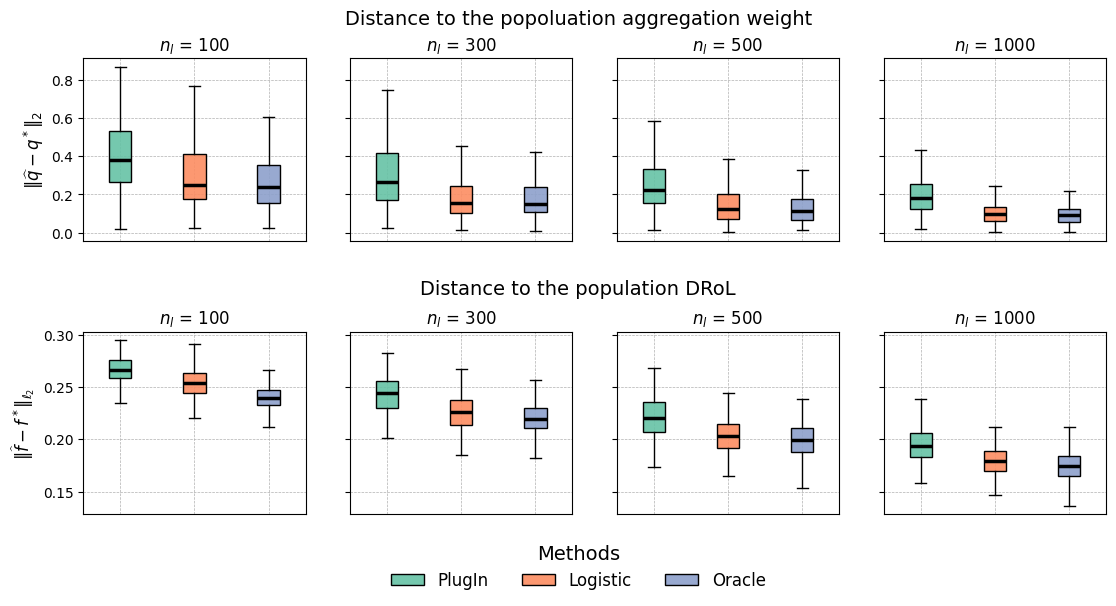}
\vspace{-3.5mm}
    \caption{Comparison of variants of DRoL estimators. \texttt{PlugIn} denotes the plug-in DRoL estimator with \(\HH = \Delta^L\). In contrast, \texttt{Logistic} and \texttt{Oracle} are bias-corrected DRoL estimators implemented via Algorithm \ref{alg: cs mm} with \(\HH = \Delta^L\), where density ratios are estimated using logistic regression or set to their true values, respectively. In this experiment, the number of source domains is fixed at $L=3$, with per-source sample sizes varied over $n_l\in \{100, 300, 500, 1000\}$ and the unlabeled target sample size fixed at $N = 20,\!000$. The top panel displays the error $\|\widehat{q} - q^*\|_2$ between the estimated aggregation weight and the true aggregation weight at the population level. The bottom panel shows the error \(\|\widehat{f} - f^*\|_{\ell_2(\QQ)}\) between the estimated model and the true DRoL model.}
    \label{fig: boxplot}
    \vspace{-1.5em}
\end{figure}

Figure \ref{fig: boxplot} summarizes the results over 200 simulation runs. In the top panel, we report the error $\|\widehat{q} - q^*\|_2$ between the estimated weights $\widehat{q}$ and the true population aggregation weight $q^*$. The bottom panel displays the model estimation error \(\|\widehat{f} - f^*\|_{\ell_2(\QQ)}\), where $\widehat{f}$ is the DRoL model estimated by each method and $f^*$ is the true DRoL model at the population level.
The results demonstrate that the bias-corrected method \texttt{Logistic} outperforms the \texttt{PlugIn}, both in terms of aggregation weight accuracy and overall model estimation error, with improvements becoming more pronounced as the source sample size increases. As expected, the \texttt{Oracle} method, which leverages the true density ratios, achieves the best performance among the three approaches,  and our \texttt{Logistic} method is comparable to it.



\section{Real Data}
\label{sec: real data}
We evaluate the proposed DRoL approach using the Beijing PM2.5 Air Pollution dataset, that was initially analyzed in \cite{zhang2017cautionary} \footnote{The data is publicly available at \url{https://archive.ics.uci.edu/dataset/501/beijing+multi+site+air+quality+data}}. This dataset contains PM2.5 concentration measurements collected from 2013 to 2016 at 12 nationally controlled air-quality monitoring sites in Beijing, along with various meteorological covariates, including temperature, pressure, dew point, precipitation, wind direction, and wind speed 
\footnote{
The wind direction variable is further categorized into five groups: northwest, northeast, calm and variable, southwest, and southeast.}.

In particular, direct measurement of PM2.5 concentrations is resource intensive, requiring specialized equipment that is rarely available outside of professional monitoring sites. In contrast, meteorological covariates are widely accessible and exhibit strong associations with PM2.5 levels \cite{tai2010correlations, zhang2017cautionary,xu2018impact,chen2020influence}, making it practically useful in predicting PM2.5 with meteorological covariates. 
However, the relationship between meteorological variables and PM2.5 may vary significantly across districts due to localized environmental factors (e.g., urbanization, industrial activity). As a result, models trained solely on data from monitoring sites fail to generalize to districts with distinct meteorological patterns. 

Constructing models that generalize well across 
the city of Beijing is critical, since the PM2.5 forecasting may be used to provide early warnings to citizens and inform regulatory actions to mitigate air pollution.
This challenge aligns naturally with the Multi-source Unsupervised Domain Adaptation (MSDA) framework. Here, labeled source domains correspond to air monitoring sites (with paired PM2.5 and meteorological data), while unlabeled target domains represent districts where only meteorological measurements are available. Our goal is to leverage labeled source data and unlabeled target covariates to build robust models that generalize well to other locations in the city, considering the distribution shifts caused by spatial heterogeneity.

In this study, we select $L=5$
representative air monitoring sites to serve as the source domains, namely:
Aotizhongxin (eastern traffic corridor), Dongsi (central urban), Tiantan (southern cultural hub), Guanyuan (northwestern urban), and Wanliu (northwestern residential), and they represent a diverse cross-section of Beijing's official air quality monitoring network. The remaining seven monitoring sites are used as target domains to evaluate the proposed approach.

For each year $\in \{2013,...2016\}$ and for each season (denoted as SP for Spring, SU for Summer, AU for Autumn, and WI for Winter)\footnote{Spring: March–May; Summer: June–August; Autumn: September–November; Winter: December–February (of the following year).}, we develop predictive models based on the five source domains ($L=5$) and evaluate their performance using the worst-case reward (i.e., the minimum reward) across the seven target sites. Consistent with the experiments described in Section \ref{sec: simu-varyL}, we compare our proposed \texttt{DRoL} method (implemented via Algorithm \ref{alg: cs mm} with the default constraint \(\HH=\Delta^L\)) against \texttt{ERM}, \texttt{ImpWeight}, and \texttt{GroupDRO} that have been described in Section \ref{sec: simu-varyL}.
\begin{figure}[!htp]
\vspace{-1em}
    \centering
    \includegraphics[width=0.85\linewidth]{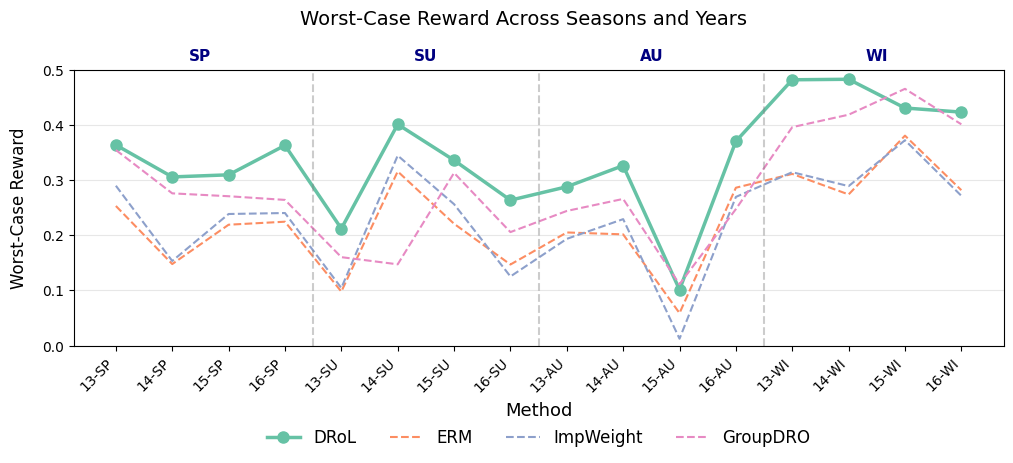}
    \vspace{-3.5mm}
    \caption{Comparison of methods in terms of the worst-case reward evaluated over seven target domains across seasons and years. The \texttt{DRoL} method is implemented by Algorithm \ref{alg: cs mm} (using default $\HH=\Delta^L$). The descriptions of methods \texttt{ERM}, \texttt{ImpWeight}, and \texttt{GroupDRO} are provided in Section \ref{sec: simu-varyL}.}
    \label{fig:air-worstcase}
    \vspace{-1.5em}
\end{figure}

Figure \ref{fig:air-worstcase} summarizes the results for each year and season. Our proposed \texttt{DRoL} method achieves the highest worst-case reward across all years and seasons, with the only exception of Winter 2015, where its performance is slightly lower than that of \texttt{GroupDRO}. While \texttt{GroupDRO} also yields a robust prediction model, the usage of unlabeled target data in the proposed \texttt{DRoL} is tailored to each specific target domain using covariate information, which explains its superior performance.
Finally, methods like \texttt{ImpWeight} and \texttt{ERM}, which do not account for potential conditional outcome distribution shifts across source and target domains, deliver the poorest performance in most cases for this application. Although \texttt{ImpWeight} makes use of unlabeled target data, its benefits are limited compared to our approach.

\begin{figure}[!htp]
\vspace{-1em}
    \centering
    \includegraphics[width=0.85\linewidth]{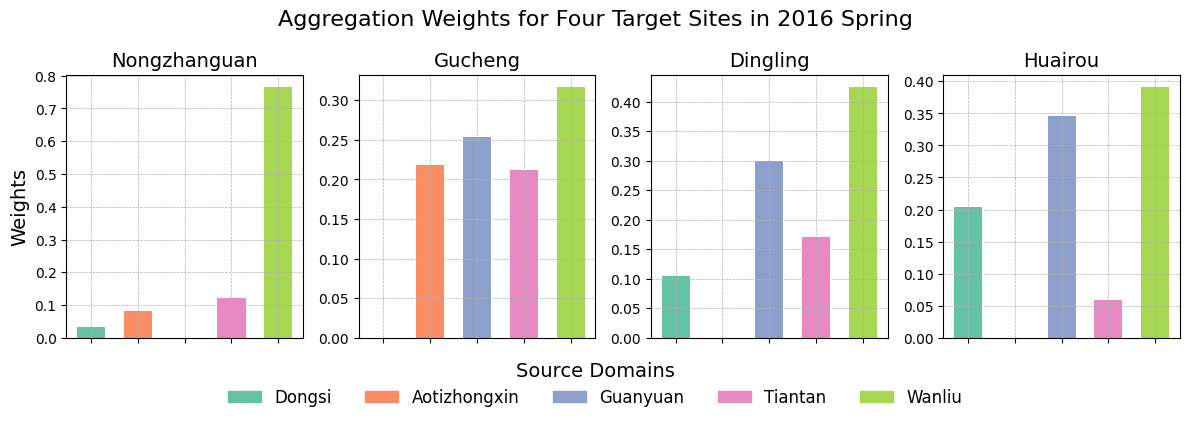}
    \vspace{-3.5mm}
    \caption{Aggregation weights for four  
    target air-monitoring sites (Nongzhuanguan, Gucheng, Dingling, Huairou) in Spring 2016. The weights are computed by Algorithm \ref{alg: cs mm} using the default constraint set $\HH=\Delta^L$.
    }
    \label{fig:air-weight}
    \vspace{-1.5em}
\end{figure}

Moreover, the aggregation weights produced by the proposed \texttt{DRoL} method naturally reflect the relative importance of each source domain for a given target domain. For illustrative purposes, Figure \ref{fig:air-weight} displays the aggregation weights computed for four target air-monitoring sites during Spring 2016, enabling a clear visualization of each source air-monitoring site's contribution to building a robust prediction model for each target site.

Next, we investigate how different specifications of the constraint $\HH$, which encodes prior information about the target distribution, affect the performance of our DRoL approach.
For \texttt{DRoL-Label}, we consider that 5\% of the target domain data is labeled (about 100 labeled samples in this real application), while the remaining 95\% of the target domain data is unlabeled. We set $\HH = \{q\in \Delta^L \mid \|q - \widehat{q}^{\rm label}\|_2 \leq \rho\}$, where $\widehat{q}^{\rm label}$ is the aggregation weight estimated using the limited labeled target data (as specified in \eqref{eq: gamma prior}) and $\rho\in [0,1]$ controls the size of \(\HH\). For \texttt{DRoL-Unif}, we use a uniform prior by setting $\HH = \{q\in \Delta^L \mid \|q - q^{\rm unif}\|_2 \leq \rho\}$, with \(q^{\rm unif} = (0.2,0.2,0.2,0.2,0.2)^\intercal\). In contrast, \texttt{DRoL-Default} adopts no prior information by setting \(\HH = \Delta^L\). Additionally, we include the \texttt{TargetOnly} benchmark, which fits models solely using the small amount of labeled target data.

\begin{figure}[!htp]
\vspace{-1em}
    \centering
    \includegraphics[width=0.95\linewidth]{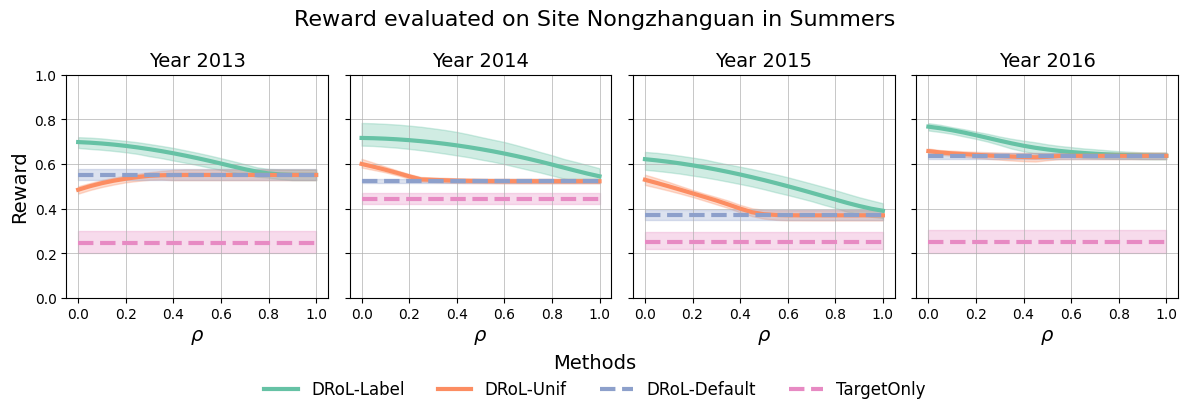}
    \vspace{-3.5mm}
    \caption{
    Comparison of different methods in terms of the reward evaluated on the target air-monitoring site Nongzhanguan during the summer season. In each simulation, the models are trained using a randomly selected 90\% subsample of the original dataset.
    The plotted curves are the mean rewards computed over 200 subsampled data sets, and the shaded error bands represent the 10th and 90th percentiles of the rewards across 200 subsampled data sets. 
    \texttt{DRoL-Label}, \texttt{DRoL-Unif}, \texttt{DRoL-Default} are implemented via Algorithm \ref{alg: cs mm} with different specifications of the constraint set $\HH$.
    \texttt{DRoL-Label} sets $\HH = \{q\in \Delta^L\mid \|q-\widehat{q}^{\rm label}\|_2\leq \rho\}$, where $\widehat{q}^{\rm label}$ is the aggregation weight fitted from a small set of labeled target data (5\% of the target dataset), as described in \eqref{eq: gamma prior}. \texttt{DRoL-Unif} sets $\HH = \{q\in \Delta^L\mid \|q-q^{\rm unif}\|_2\leq \rho\}$ with a uniform weight vector $q^{\rm unif}$. \texttt{DRoL-Default} uses the default $\HH = \Delta^L$, i.e., no prior information. \texttt{TargetOnly} fits models using the limited labeled data from the target air-monitoring site.}
    \label{fig:air-prior}
    \vspace{-1.5em}
\end{figure}

Figure \ref{fig:air-prior} illustrates the comparison using data from the target air-monitoring site Nongzhanguan during the summer season. To assess the uncertainty of each method, we subsample 90\% of the data without replacement 200 times. The plotted curves represent the mean reward computed over these 200 subsamples, with shaded error bands indicating the 10th and 90th percentile variability.
When comparing \texttt{DRoL-Label} (which leverages prior information estimated from the limited labeled target data) with \texttt{DRoL-Default} (which does not use any prior information), it is clear that accurately specified prior information can enhance the predictive performance of DRoL. Moreover, \texttt{DRoL-Unif} constrains the solution around the uniform vector \(q^{\rm unif} = (0.2, 0.2, 0.2, 0.2, 0.2)^\intercal\), failing to capture the target domain’s characteristics. As a result, it may perform better or worse than \texttt{DRoL-Unif} in different years.
Finally, the \texttt{TargetOnly} benchmark, which fits a model solely on the limited labeled target data, performs the worst among these methods and fails to deliver a reliable prediction for the target domain. 
These findings highlight DRoL's ability to integrate prior information, which can improve its predictive performance on a specific target domain when the prior information captures the true target distribution.

%
%

\begin{acks}[Acknowledgments]
The authors extend their gratitude to Koulik Khamaru for the valuable discussions on the NP-hardness of the regret-based approach.
The research of Z. Wang and Z. Guo was partly supported by the NSF DMS 2015373 and NIH R01GM140463 and R01LM013614; Z.Guo also acknowledges financial support for visiting the Institute of Mathematical Research (FIM) at ETH Zurich.
P. B\"{u}hlmann received funding from the European Research Council (ERC) under the European Union’s Horizon 2020 research and innovation program (grant agreement No. 786461)
\end{acks}
%

\begin{supplement}
\stitle{Supplement to ``Distributionally Robust Machine Learning with Multi-Source Data''.} 
\sdescription{The supplementary material contains additional methods, numerical results, and proof deferred due to space constraints.}
\end{supplement}


\bibliographystyle{imsart-number} 
\bibliography{ref}       

\begin{thebibliography}{71}

\bibitem{agarwal2022minimax}
\begin{binproceedings}[author]
\bauthor{\bsnm{Agarwal},~\bfnm{Alekh}\binits{A.}} \AND \bauthor{\bsnm{Zhang},~\bfnm{Tong}\binits{T.}}
(\byear{2022}).
\btitle{Minimax regret optimization for robust machine learning under distribution shift}.
In \bbooktitle{Conference on Learning Theory}
\bpages{2704--2729}.
\bpublisher{PMLR}.
\end{binproceedings}
\endbibitem

\bibitem{athey2018approximate}
\begin{barticle}[author]
\bauthor{\bsnm{Athey},~\bfnm{Susan}\binits{S.}}, \bauthor{\bsnm{Imbens},~\bfnm{Guido~W}\binits{G.~W.}} \AND \bauthor{\bsnm{Wager},~\bfnm{Stefan}\binits{S.}}
(\byear{2018}).
\btitle{Approximate residual balancing: debiased inference of average treatment effects in high dimensions}.
\bjournal{Journal of the Royal Statistical Society Series B: Statistical Methodology}
\bvolume{80}
\bpages{597--623}.
\end{barticle}
\endbibitem

\bibitem{ben2006analysis}
\begin{barticle}[author]
\bauthor{\bsnm{Ben-David},~\bfnm{Shai}\binits{S.}}, \bauthor{\bsnm{Blitzer},~\bfnm{John}\binits{J.}}, \bauthor{\bsnm{Crammer},~\bfnm{Koby}\binits{K.}} \AND \bauthor{\bsnm{Pereira},~\bfnm{Fernando}\binits{F.}}
(\byear{2006}).
\btitle{Analysis of representations for domain adaptation}.
\bjournal{Advances in neural information processing systems}
\bvolume{19}.
\end{barticle}
\endbibitem

\bibitem{ben2013robust}
\begin{barticle}[author]
\bauthor{\bsnm{Ben-Tal},~\bfnm{Aharon}\binits{A.}}, \bauthor{\bsnm{Den~Hertog},~\bfnm{Dick}\binits{D.}}, \bauthor{\bsnm{De~Waegenaere},~\bfnm{Anja}\binits{A.}}, \bauthor{\bsnm{Melenberg},~\bfnm{Bertrand}\binits{B.}} \AND \bauthor{\bsnm{Rennen},~\bfnm{Gijs}\binits{G.}}
(\byear{2013}).
\btitle{Robust solutions of optimization problems affected by uncertain probabilities}.
\bjournal{Management Science}
\bvolume{59}
\bpages{341--357}.
\end{barticle}
\endbibitem

\bibitem{biau2012analysis}
\begin{barticle}[author]
\bauthor{\bsnm{Biau},~\bfnm{G{\'e}rard}\binits{G.}}
(\byear{2012}).
\btitle{Analysis of a random forests model}.
\bjournal{The Journal of Machine Learning Research}
\bvolume{13}
\bpages{1063--1095}.
\end{barticle}
\endbibitem

\bibitem{biau2008consistency}
\begin{barticle}[author]
\bauthor{\bsnm{Biau},~\bfnm{G{\'e}rard}\binits{G.}}, \bauthor{\bsnm{Devroye},~\bfnm{Luc}\binits{L.}} \AND \bauthor{\bsnm{Lugosi},~\bfnm{G{\"a}bor}\binits{G.}}
(\byear{2008}).
\btitle{Consistency of random forests and other averaging classifiers.}
\bjournal{Journal of Machine Learning Research}
\bvolume{9}.
\end{barticle}
\endbibitem

\bibitem{bickel2009simultaneous}
\begin{barticle}[author]
\bauthor{\bsnm{Bickel},~\bfnm{Peter~J}\binits{P.~J.}}, \bauthor{\bsnm{Ritov},~\bfnm{Ya’acov}\binits{Y.}} \AND \bauthor{\bsnm{Tsybakov},~\bfnm{Alexandre~B}\binits{A.~B.}}
(\byear{2009}).
\btitle{Simultaneous analysis of Lasso and Dantzig selector}.
\bjournal{The Annals of Statistics}
\bvolume{37}
\bpages{1705--1732}.
\end{barticle}
\endbibitem

\bibitem{buhlmann2015magging}
\begin{barticle}[author]
\bauthor{\bsnm{B{\"u}hlmann},~\bfnm{Peter}\binits{P.}} \AND \bauthor{\bsnm{Meinshausen},~\bfnm{Nicolai}\binits{N.}}
(\byear{2015}).
\btitle{Magging: maximin aggregation for inhomogeneous large-scale data}.
\bjournal{Proceedings of the IEEE}
\bvolume{104}
\bpages{126--135}.
\end{barticle}
\endbibitem

\bibitem{buhlmann2011statistics}
\begin{bbook}[author]
\bauthor{\bsnm{B{\"u}hlmann},~\bfnm{Peter}\binits{P.}} \AND \bauthor{\bparticle{van~de} \bsnm{Geer},~\bfnm{Sara}\binits{S.}}
(\byear{2011}).
\btitle{Statistics for high-dimensional data: methods, theory and applications}.
\bpublisher{Springer Science \& Business Media}.
\end{bbook}
\endbibitem

\bibitem{candes2007dantzig}
\begin{barticle}[author]
\bauthor{\bsnm{Candes},~\bfnm{Emmanuel}\binits{E.}} \AND \bauthor{\bsnm{Tao},~\bfnm{Terence}\binits{T.}}
(\byear{2007}).
\btitle{The Dantzig selector: Statistical estimation when p is much larger than n}.
\bjournal{Annals of statistics}
\bvolume{35}
\bpages{2313--2404}.
\end{barticle}
\endbibitem

\bibitem{cao2021data}
\begin{barticle}[author]
\bauthor{\bsnm{Cao},~\bfnm{Longbing}\binits{L.}}, \bauthor{\bsnm{Yang},~\bfnm{Qiang}\binits{Q.}} \AND \bauthor{\bsnm{Yu},~\bfnm{Philip~S}\binits{P.~S.}}
(\byear{2021}).
\btitle{Data science and AI in FinTech: An overview}.
\bjournal{International Journal of Data Science and Analytics}
\bvolume{12}
\bpages{81--99}.
\end{barticle}
\endbibitem

\bibitem{chen2008semiparametric}
\begin{barticle}[author]
\bauthor{\bsnm{Chen},~\bfnm{Xiaohong}\binits{X.}}, \bauthor{\bsnm{Hong},~\bfnm{Han}\binits{H.}} \AND \bauthor{\bsnm{Tarozzi},~\bfnm{Alessandro}\binits{A.}}
(\byear{2008}).
\btitle{Semiparametric efficiency in GMM models in auxiliary data}.
\bjournal{The Annals of Statistics}
\bvolume{36}
\bpages{808--843}.
\end{barticle}
\endbibitem

\bibitem{chen2020influence}
\begin{barticle}[author]
\bauthor{\bsnm{Chen},~\bfnm{Ziyue}\binits{Z.}}, \bauthor{\bsnm{Chen},~\bfnm{Danlu}\binits{D.}}, \bauthor{\bsnm{Zhao},~\bfnm{Chuanfeng}\binits{C.}}, \bauthor{\bsnm{Kwan},~\bfnm{Mei-po}\binits{M.-p.}}, \bauthor{\bsnm{Cai},~\bfnm{Jun}\binits{J.}}, \bauthor{\bsnm{Zhuang},~\bfnm{Yan}\binits{Y.}}, \bauthor{\bsnm{Zhao},~\bfnm{Bo}\binits{B.}}, \bauthor{\bsnm{Wang},~\bfnm{Xiaoyan}\binits{X.}}, \bauthor{\bsnm{Chen},~\bfnm{Bin}\binits{B.}}, \bauthor{\bsnm{Yang},~\bfnm{Jing}\binits{J.}} \betal{et~al.}
(\byear{2020}).
\btitle{Influence of meteorological conditions on PM2. 5 concentrations across China: A review of methodology and mechanism}.
\bjournal{Environment international}
\bvolume{139}
\bpages{105558}.
\end{barticle}
\endbibitem

\bibitem{deng2020distributionally}
\begin{barticle}[author]
\bauthor{\bsnm{Deng},~\bfnm{Yuyang}\binits{Y.}}, \bauthor{\bsnm{Kamani},~\bfnm{Mohammad~Mahdi}\binits{M.~M.}} \AND \bauthor{\bsnm{Mahdavi},~\bfnm{Mehrdad}\binits{M.}}
(\byear{2020}).
\btitle{Distributionally robust federated averaging}.
\bjournal{Advances in neural information processing systems}
\bvolume{33}
\bpages{15111--15122}.
\end{barticle}
\endbibitem

\bibitem{diana2021minimax}
\begin{binproceedings}[author]
\bauthor{\bsnm{Diana},~\bfnm{Emily}\binits{E.}}, \bauthor{\bsnm{Gill},~\bfnm{Wesley}\binits{W.}}, \bauthor{\bsnm{Kearns},~\bfnm{Michael}\binits{M.}}, \bauthor{\bsnm{Kenthapadi},~\bfnm{Krishnaram}\binits{K.}} \AND \bauthor{\bsnm{Roth},~\bfnm{Aaron}\binits{A.}}
(\byear{2021}).
\btitle{Minimax group fairness: Algorithms and experiments}.
In \bbooktitle{Proceedings of the 2021 AAAI/ACM Conference on AI, Ethics, and Society}
\bpages{66--76}.
\end{binproceedings}
\endbibitem

\bibitem{duan2012domain}
\begin{barticle}[author]
\bauthor{\bsnm{Duan},~\bfnm{Lixin}\binits{L.}}, \bauthor{\bsnm{Xu},~\bfnm{Dong}\binits{D.}} \AND \bauthor{\bsnm{Tsang},~\bfnm{Ivor Wai-Hung}\binits{I.~W.-H.}}
(\byear{2012}).
\btitle{Domain adaptation from multiple sources: A domain-dependent regularization approach}.
\bjournal{IEEE Transactions on neural networks and learning systems}
\bvolume{23}
\bpages{504--518}.
\end{barticle}
\endbibitem

\bibitem{eldar2008minimax}
\begin{barticle}[author]
\bauthor{\bsnm{Eldar},~\bfnm{Yonina~C}\binits{Y.~C.}}, \bauthor{\bsnm{Beck},~\bfnm{Amir}\binits{A.}} \AND \bauthor{\bsnm{Teboulle},~\bfnm{Marc}\binits{M.}}
(\byear{2008}).
\btitle{A minimax Chebyshev estimator for bounded error estimation}.
\bjournal{IEEE transactions on signal processing}
\bvolume{56}
\bpages{1388--1397}.
\end{barticle}
\endbibitem

\bibitem{farrell2021deep}
\begin{barticle}[author]
\bauthor{\bsnm{Farrell},~\bfnm{Max~H}\binits{M.~H.}}, \bauthor{\bsnm{Liang},~\bfnm{Tengyuan}\binits{T.}} \AND \bauthor{\bsnm{Misra},~\bfnm{Sanjog}\binits{S.}}
(\byear{2021}).
\btitle{Deep neural networks for estimation and inference}.
\bjournal{Econometrica}
\bvolume{89}
\bpages{181--213}.
\end{barticle}
\endbibitem

\bibitem{ganin2016domain}
\begin{barticle}[author]
\bauthor{\bsnm{Ganin},~\bfnm{Yaroslav}\binits{Y.}}, \bauthor{\bsnm{Ustinova},~\bfnm{Evgeniya}\binits{E.}}, \bauthor{\bsnm{Ajakan},~\bfnm{Hana}\binits{H.}}, \bauthor{\bsnm{Germain},~\bfnm{Pascal}\binits{P.}}, \bauthor{\bsnm{Larochelle},~\bfnm{Hugo}\binits{H.}}, \bauthor{\bsnm{Laviolette},~\bfnm{Fran{\c{c}}ois}\binits{F.}}, \bauthor{\bsnm{March},~\bfnm{Mario}\binits{M.}} \AND \bauthor{\bsnm{Lempitsky},~\bfnm{Victor}\binits{V.}}
(\byear{2016}).
\btitle{Domain-adversarial training of neural networks}.
\bjournal{Journal of machine learning research}
\bvolume{17}
\bpages{1--35}.
\end{barticle}
\endbibitem

\bibitem{gretton2009covariate}
\begin{barticle}[author]
\bauthor{\bsnm{Gretton},~\bfnm{Arthur}\binits{A.}}, \bauthor{\bsnm{Smola},~\bfnm{Alex~J}\binits{A.~J.}}, \bauthor{\bsnm{Huang},~\bfnm{Jiayuan}\binits{J.}}, \bauthor{\bsnm{Schmittfull},~\bfnm{Marcel}\binits{M.}}, \bauthor{\bsnm{Borgwardt},~\bfnm{Karsten~M}\binits{K.~M.}} \AND \bauthor{\bsnm{Sch{\"o}llkopf},~\bfnm{Bernhard}\binits{B.}}
(\byear{2009}).
\btitle{Covariate Shift by Kernel Mean Matching}.
\bpages{131--160}.
\end{barticle}
\endbibitem

\bibitem{guo2020inference}
\begin{barticle}[author]
\bauthor{\bsnm{Guo},~\bfnm{Zijian}\binits{Z.}}
(\byear{2023}).
\btitle{Statistical inference for maximin effects: Identifying stable associations across multiple studies}.
\bjournal{Journal of the American Statistical Association}
\bpages{1--32}.
\end{barticle}
\endbibitem

\bibitem{hu2015deep}
\begin{binproceedings}[author]
\bauthor{\bsnm{Hu},~\bfnm{Junlin}\binits{J.}}, \bauthor{\bsnm{Lu},~\bfnm{Jiwen}\binits{J.}} \AND \bauthor{\bsnm{Tan},~\bfnm{Yap-Peng}\binits{Y.-P.}}
(\byear{2015}).
\btitle{Deep transfer metric learning}.
In \bbooktitle{Proceedings of the IEEE conference on computer vision and pattern recognition}
\bpages{325--333}.
\end{binproceedings}
\endbibitem

\bibitem{hu2018does}
\begin{binproceedings}[author]
\bauthor{\bsnm{Hu},~\bfnm{Weihua}\binits{W.}}, \bauthor{\bsnm{Niu},~\bfnm{Gang}\binits{G.}}, \bauthor{\bsnm{Sato},~\bfnm{Issei}\binits{I.}} \AND \bauthor{\bsnm{Sugiyama},~\bfnm{Masashi}\binits{M.}}
(\byear{2018}).
\btitle{Does distributionally robust supervised learning give robust classifiers?}
In \bbooktitle{International Conference on Machine Learning}
\bpages{2029--2037}.
\bpublisher{PMLR}.
\end{binproceedings}
\endbibitem

\bibitem{huang2012estimation}
\begin{barticle}[author]
\bauthor{\bsnm{Huang},~\bfnm{Jian}\binits{J.}} \AND \bauthor{\bsnm{Zhang},~\bfnm{Cun-Hui}\binits{C.-H.}}
(\byear{2012}).
\btitle{Estimation and selection via absolute penalized convex minimization and its multistage adaptive applications}.
\bjournal{The Journal of Machine Learning Research}
\bvolume{13}
\bpages{1839--1864}.
\end{barticle}
\endbibitem

\bibitem{kanamori2009least}
\begin{barticle}[author]
\bauthor{\bsnm{Kanamori},~\bfnm{Takafumi}\binits{T.}}, \bauthor{\bsnm{Hido},~\bfnm{Shohei}\binits{S.}} \AND \bauthor{\bsnm{Sugiyama},~\bfnm{Masashi}\binits{M.}}
(\byear{2009}).
\btitle{A least-squares approach to direct importance estimation}.
\bjournal{The Journal of Machine Learning Research}
\bvolume{10}
\bpages{1391--1445}.
\end{barticle}
\endbibitem

\bibitem{koh2021wilds}
\begin{binproceedings}[author]
\bauthor{\bsnm{Koh},~\bfnm{Pang~Wei}\binits{P.~W.}}, \bauthor{\bsnm{Sagawa},~\bfnm{Shiori}\binits{S.}}, \bauthor{\bsnm{Marklund},~\bfnm{Henrik}\binits{H.}}, \bauthor{\bsnm{Xie},~\bfnm{Sang~Michael}\binits{S.~M.}}, \bauthor{\bsnm{Zhang},~\bfnm{Marvin}\binits{M.}}, \bauthor{\bsnm{Balsubramani},~\bfnm{Akshay}\binits{A.}}, \bauthor{\bsnm{Hu},~\bfnm{Weihua}\binits{W.}}, \bauthor{\bsnm{Yasunaga},~\bfnm{Michihiro}\binits{M.}}, \bauthor{\bsnm{Phillips},~\bfnm{Richard~Lanas}\binits{R.~L.}}, \bauthor{\bsnm{Gao},~\bfnm{Irena}\binits{I.}} \betal{et~al.}
(\byear{2021}).
\btitle{Wilds: A benchmark of in-the-wild distribution shifts}.
In \bbooktitle{International Conference on Machine Learning}
\bpages{5637--5664}.
\bpublisher{PMLR}.
\end{binproceedings}
\endbibitem

\bibitem{komiya1988elementary}
\begin{barticle}[author]
\bauthor{\bsnm{Komiya},~\bfnm{Hidetoshi}\binits{H.}}
(\byear{1988}).
\btitle{Elementary proof for Sion's minimax theorem}.
\bjournal{Kodai mathematical journal}
\bvolume{11}
\bpages{5--7}.
\end{barticle}
\endbibitem

\bibitem{long2018transferable}
\begin{barticle}[author]
\bauthor{\bsnm{Long},~\bfnm{Mingsheng}\binits{M.}}, \bauthor{\bsnm{Cao},~\bfnm{Yue}\binits{Y.}}, \bauthor{\bsnm{Cao},~\bfnm{Zhangjie}\binits{Z.}}, \bauthor{\bsnm{Wang},~\bfnm{Jianmin}\binits{J.}} \AND \bauthor{\bsnm{Jordan},~\bfnm{Michael~I}\binits{M.~I.}}
(\byear{2018}).
\btitle{Transferable representation learning with deep adaptation networks}.
\bjournal{IEEE transactions on pattern analysis and machine intelligence}
\bvolume{41}
\bpages{3071--3085}.
\end{barticle}
\endbibitem

\bibitem{malinin2021shifts}
\begin{barticle}[author]
\bauthor{\bsnm{Malinin},~\bfnm{Andrey}\binits{A.}}, \bauthor{\bsnm{Band},~\bfnm{Neil}\binits{N.}}, \bauthor{\bsnm{Chesnokov},~\bfnm{German}\binits{G.}}, \bauthor{\bsnm{Gal},~\bfnm{Yarin}\binits{Y.}}, \bauthor{\bsnm{Gales},~\bfnm{Mark~JF}\binits{M.~J.}}, \bauthor{\bsnm{Noskov},~\bfnm{Alexey}\binits{A.}}, \bauthor{\bsnm{Ploskonosov},~\bfnm{Andrey}\binits{A.}}, \bauthor{\bsnm{Prokhorenkova},~\bfnm{Liudmila}\binits{L.}}, \bauthor{\bsnm{Provilkov},~\bfnm{Ivan}\binits{I.}}, \bauthor{\bsnm{Raina},~\bfnm{Vatsal}\binits{V.}} \betal{et~al.}
(\byear{2021}).
\btitle{Shifts: A dataset of real distributional shift across multiple large-scale tasks}.
\bjournal{arXiv preprint arXiv:2107.07455}.
\end{barticle}
\endbibitem

\bibitem{mansour2008domain}
\begin{barticle}[author]
\bauthor{\bsnm{Mansour},~\bfnm{Yishay}\binits{Y.}}, \bauthor{\bsnm{Mohri},~\bfnm{Mehryar}\binits{M.}} \AND \bauthor{\bsnm{Rostamizadeh},~\bfnm{Afshin}\binits{A.}}
(\byear{2008}).
\btitle{Domain adaptation with multiple sources}.
\bjournal{Advances in neural information processing systems}
\bvolume{21}.
\end{barticle}
\endbibitem

\bibitem{martinez2020minimax}
\begin{binproceedings}[author]
\bauthor{\bsnm{Martinez},~\bfnm{Natalia}\binits{N.}}, \bauthor{\bsnm{Bertran},~\bfnm{Martin}\binits{M.}} \AND \bauthor{\bsnm{Sapiro},~\bfnm{Guillermo}\binits{G.}}
(\byear{2020}).
\btitle{Minimax pareto fairness: A multi objective perspective}.
In \bbooktitle{International Conference on Machine Learning}
\bpages{6755--6764}.
\bpublisher{PMLR}.
\end{binproceedings}
\endbibitem

\bibitem{meinshausen2015maximin}
\begin{barticle}[author]
\bauthor{\bsnm{Meinshausen},~\bfnm{Nicolai}\binits{N.}} \AND \bauthor{\bsnm{B{\"u}hlmann},~\bfnm{Peter}\binits{P.}}
(\byear{2015}).
\btitle{Maximin effects in inhomogeneous large-scale data}.
\bjournal{The Annals of Statistics}
\bvolume{43}
\bpages{1801--1830}.
\end{barticle}
\endbibitem

\bibitem{meinshausen2006quantile}
\begin{barticle}[author]
\bauthor{\bsnm{Meinshausen},~\bfnm{Nicolai}\binits{N.}} \AND \bauthor{\bsnm{Ridgeway},~\bfnm{Greg}\binits{G.}}
(\byear{2006}).
\btitle{Quantile regression forests.}
\bjournal{Journal of machine learning research}
\bvolume{7}.
\end{barticle}
\endbibitem

\bibitem{menon2016linking}
\begin{binproceedings}[author]
\bauthor{\bsnm{Menon},~\bfnm{Aditya}\binits{A.}} \AND \bauthor{\bsnm{Ong},~\bfnm{Cheng~Soon}\binits{C.~S.}}
(\byear{2016}).
\btitle{Linking losses for density ratio and class-probability estimation}.
In \bbooktitle{International Conference on Machine Learning}
\bpages{304--313}.
\bpublisher{PMLR}.
\end{binproceedings}
\endbibitem

\bibitem{milanese1985optimal}
\begin{barticle}[author]
\bauthor{\bsnm{Milanese},~\bfnm{Mario}\binits{M.}} \AND \bauthor{\bsnm{Tempo},~\bfnm{Roberto}\binits{R.}}
(\byear{1985}).
\btitle{Optimal algorithms theory for robust estimation and prediction}.
\bjournal{IEEE Transactions on Automatic Control}
\bvolume{30}
\bpages{730--738}.
\end{barticle}
\endbibitem

\bibitem{mo2024minimax}
\begin{barticle}[author]
\bauthor{\bsnm{Mo},~\bfnm{Weibin}\binits{W.}}, \bauthor{\bsnm{Tang},~\bfnm{Weijing}\binits{W.}}, \bauthor{\bsnm{Xue},~\bfnm{Songkai}\binits{S.}}, \bauthor{\bsnm{Liu},~\bfnm{Yufeng}\binits{Y.}} \AND \bauthor{\bsnm{Zhu},~\bfnm{Ji}\binits{J.}}
(\byear{2024}).
\btitle{Minimax Regret Learning for Data with Heterogeneous Subgroups}.
\bjournal{arXiv preprint arXiv:2405.01709}.
\end{barticle}
\endbibitem

\bibitem{mohri2019agnostic}
\begin{binproceedings}[author]
\bauthor{\bsnm{Mohri},~\bfnm{Mehryar}\binits{M.}}, \bauthor{\bsnm{Sivek},~\bfnm{Gary}\binits{G.}} \AND \bauthor{\bsnm{Suresh},~\bfnm{Ananda~Theertha}\binits{A.~T.}}
(\byear{2019}).
\btitle{Agnostic federated learning}.
In \bbooktitle{International Conference on Machine Learning}
\bpages{4615--4625}.
\bpublisher{PMLR}.
\end{binproceedings}
\endbibitem

\bibitem{namkoong2017variance}
\begin{barticle}[author]
\bauthor{\bsnm{Namkoong},~\bfnm{Hongseok}\binits{H.}} \AND \bauthor{\bsnm{Duchi},~\bfnm{John~C}\binits{J.~C.}}
(\byear{2017}).
\btitle{Variance-based regularization with convex objectives}.
\bjournal{Advances in neural information processing systems}
\bvolume{30}.
\end{barticle}
\endbibitem

\bibitem{negahban2012unified}
\begin{barticle}[author]
\bauthor{\bsnm{Negahban},~\bfnm{Sahand~N}\binits{S.~N.}}, \bauthor{\bsnm{Ravikumar},~\bfnm{Pradeep}\binits{P.}}, \bauthor{\bsnm{Wainwright},~\bfnm{Martin~J}\binits{M.~J.}} \AND \bauthor{\bsnm{Yu},~\bfnm{Bin}\binits{B.}}
(\byear{2012}).
\btitle{A unified framework for high-dimensional analysis of M-estimators with decomposable regularizers}.
\end{barticle}
\endbibitem

\bibitem{nguyen2010estimating}
\begin{barticle}[author]
\bauthor{\bsnm{Nguyen},~\bfnm{XuanLong}\binits{X.}}, \bauthor{\bsnm{Wainwright},~\bfnm{Martin~J}\binits{M.~J.}} \AND \bauthor{\bsnm{Jordan},~\bfnm{Michael~I}\binits{M.~I.}}
(\byear{2010}).
\btitle{Estimating divergence functionals and the likelihood ratio by convex risk minimization}.
\bjournal{IEEE Transactions on Information Theory}
\bvolume{56}
\bpages{5847--5861}.
\end{barticle}
\endbibitem

\bibitem{perone2019unsupervised}
\begin{barticle}[author]
\bauthor{\bsnm{Perone},~\bfnm{Christian~S}\binits{C.~S.}}, \bauthor{\bsnm{Ballester},~\bfnm{Pedro}\binits{P.}}, \bauthor{\bsnm{Barros},~\bfnm{Rodrigo~C}\binits{R.~C.}} \AND \bauthor{\bsnm{Cohen-Adad},~\bfnm{Julien}\binits{J.}}
(\byear{2019}).
\btitle{Unsupervised domain adaptation for medical imaging segmentation with self-ensembling}.
\bjournal{NeuroImage}
\bvolume{194}
\bpages{1--11}.
\end{barticle}
\endbibitem

\bibitem{qin1998inferences}
\begin{barticle}[author]
\bauthor{\bsnm{Qin},~\bfnm{Jing}\binits{J.}}
(\byear{1998}).
\btitle{Inferences for case-control and semiparametric two-sample density ratio models}.
\bjournal{Biometrika}
\bvolume{85}
\bpages{619--630}.
\end{barticle}
\endbibitem

\bibitem{quinonero2008dataset}
\begin{bbook}[author]
\bauthor{\bsnm{Quinonero-Candela},~\bfnm{Joaquin}\binits{J.}}, \bauthor{\bsnm{Sugiyama},~\bfnm{Masashi}\binits{M.}}, \bauthor{\bsnm{Schwaighofer},~\bfnm{Anton}\binits{A.}} \AND \bauthor{\bsnm{Lawrence},~\bfnm{Neil~D}\binits{N.~D.}}
(\byear{2008}).
\btitle{Dataset shift in machine learning}.
\bpublisher{Mit Press}.
\end{bbook}
\endbibitem

\bibitem{ren2018generalized}
\begin{barticle}[author]
\bauthor{\bsnm{Ren},~\bfnm{Chuan-Xian}\binits{C.-X.}}, \bauthor{\bsnm{Xu},~\bfnm{Xiao-Lin}\binits{X.-L.}} \AND \bauthor{\bsnm{Yan},~\bfnm{Hong}\binits{H.}}
(\byear{2018}).
\btitle{Generalized conditional domain adaptation: A causal perspective with low-rank translators}.
\bjournal{IEEE transactions on cybernetics}
\bvolume{50}
\bpages{821--834}.
\end{barticle}
\endbibitem

\bibitem{rosenbaum1983central}
\begin{barticle}[author]
\bauthor{\bsnm{Rosenbaum},~\bfnm{Paul~R}\binits{P.~R.}} \AND \bauthor{\bsnm{Rubin},~\bfnm{Donald~B}\binits{D.~B.}}
(\byear{1983}).
\btitle{The central role of the propensity score in observational studies for causal effects}.
\bjournal{Biometrika}
\bvolume{70}
\bpages{41--55}.
\end{barticle}
\endbibitem

\bibitem{rothenhausler2016confidence}
\begin{binproceedings}[author]
\bauthor{\bsnm{Rothenh{\"a}usler},~\bfnm{Dominik}\binits{D.}}, \bauthor{\bsnm{Meinshausen},~\bfnm{Nicolai}\binits{N.}} \AND \bauthor{\bsnm{B{\"u}hlmann},~\bfnm{Peter}\binits{P.}}
(\byear{2016}).
\btitle{Confidence intervals for maximin effects in inhomogeneous large-scale data}.
In \bbooktitle{Statistical Analysis for High-Dimensional Data: The Abel Symposium 2014}
\bpages{255--277}.
\bpublisher{Springer}.
\end{binproceedings}
\endbibitem

\bibitem{sagawa2019distributionally}
\begin{barticle}[author]
\bauthor{\bsnm{Sagawa},~\bfnm{Shiori}\binits{S.}}, \bauthor{\bsnm{Koh},~\bfnm{Pang~Wei}\binits{P.~W.}}, \bauthor{\bsnm{Hashimoto},~\bfnm{Tatsunori~B}\binits{T.~B.}} \AND \bauthor{\bsnm{Liang},~\bfnm{Percy}\binits{P.}}
(\byear{2019}).
\btitle{Distributionally robust neural networks for group shifts: On the importance of regularization for worst-case generalization}.
\bjournal{arXiv preprint arXiv:1911.08731}.
\end{barticle}
\endbibitem

\bibitem{saito2018maximum}
\begin{binproceedings}[author]
\bauthor{\bsnm{Saito},~\bfnm{Kuniaki}\binits{K.}}, \bauthor{\bsnm{Watanabe},~\bfnm{Kohei}\binits{K.}}, \bauthor{\bsnm{Ushiku},~\bfnm{Yoshitaka}\binits{Y.}} \AND \bauthor{\bsnm{Harada},~\bfnm{Tatsuya}\binits{T.}}
(\byear{2018}).
\btitle{Maximum classifier discrepancy for unsupervised domain adaptation}.
In \bbooktitle{Proceedings of the IEEE conference on computer vision and pattern recognition}
\bpages{3723--3732}.
\end{binproceedings}
\endbibitem

\bibitem{schmidt2020nonparametric}
\begin{barticle}[author]
\bauthor{\bsnm{Schmidt-Hieber},~\bfnm{Johannes}\binits{J.}}
(\byear{2020}).
\btitle{Nonparametric regression using deep neural networks with ReLU activation function}.
\bjournal{The Annals of Statistics}
\bvolume{48}
\bpages{1875--1897}.
\end{barticle}
\endbibitem

\bibitem{schulam2015framework}
\begin{barticle}[author]
\bauthor{\bsnm{Schulam},~\bfnm{Peter}\binits{P.}} \AND \bauthor{\bsnm{Saria},~\bfnm{Suchi}\binits{S.}}
(\byear{2015}).
\btitle{A framework for individualizing predictions of disease trajectories by exploiting multi-resolution structure}.
\bjournal{Advances in neural information processing systems}
\bvolume{28}.
\end{barticle}
\endbibitem

\bibitem{scornet2015consistency}
\begin{barticle}[author]
\bauthor{\bsnm{Scornet},~\bfnm{Erwan}\binits{E.}}, \bauthor{\bsnm{Biau},~\bfnm{G{\'e}rard}\binits{G.}} \AND \bauthor{\bsnm{Vert},~\bfnm{Jean-Philippe}\binits{J.-P.}}
(\byear{2015}).
\btitle{Consistency of random forests}.
\bjournal{Annals of Statistics}
\bvolume{43}
\bpages{1716--1741}.
\end{barticle}
\endbibitem

\bibitem{sinha2017certifying}
\begin{barticle}[author]
\bauthor{\bsnm{Sinha},~\bfnm{Aman}\binits{A.}}, \bauthor{\bsnm{Namkoong},~\bfnm{Hongseok}\binits{H.}}, \bauthor{\bsnm{Volpi},~\bfnm{Riccardo}\binits{R.}} \AND \bauthor{\bsnm{Duchi},~\bfnm{John}\binits{J.}}
(\byear{2017}).
\btitle{Certifying some distributional robustness with principled adversarial training}.
\bjournal{arXiv preprint arXiv:1710.10571}.
\end{barticle}
\endbibitem

\bibitem{sion1958general}
\begin{barticle}[author]
\bauthor{\bsnm{Sion},~\bfnm{Maurice}\binits{M.}}
(\byear{1958}).
\btitle{On general minimax theorems.}
\end{barticle}
\endbibitem

\bibitem{soma2022optimal}
\begin{barticle}[author]
\bauthor{\bsnm{Soma},~\bfnm{Tasuku}\binits{T.}}, \bauthor{\bsnm{Gatmiry},~\bfnm{Khashayar}\binits{K.}} \AND \bauthor{\bsnm{Jegelka},~\bfnm{Stefanie}\binits{S.}}
(\byear{2022}).
\btitle{Optimal algorithms for group distributionally robust optimization and beyond}.
\bjournal{arXiv preprint arXiv:2212.13669}.
\end{barticle}
\endbibitem

\bibitem{sugiyama2007direct}
\begin{barticle}[author]
\bauthor{\bsnm{Sugiyama},~\bfnm{Masashi}\binits{M.}}, \bauthor{\bsnm{Nakajima},~\bfnm{Shinichi}\binits{S.}}, \bauthor{\bsnm{Kashima},~\bfnm{Hisashi}\binits{H.}}, \bauthor{\bsnm{Buenau},~\bfnm{Paul}\binits{P.}} \AND \bauthor{\bsnm{Kawanabe},~\bfnm{Motoaki}\binits{M.}}
(\byear{2007}).
\btitle{Direct importance estimation with model selection and its application to covariate shift adaptation}.
\bjournal{Advances in neural information processing systems}
\bvolume{20}.
\end{barticle}
\endbibitem

\bibitem{sugiyama2012density}
\begin{bbook}[author]
\bauthor{\bsnm{Sugiyama},~\bfnm{Masashi}\binits{M.}}, \bauthor{\bsnm{Suzuki},~\bfnm{Taiji}\binits{T.}} \AND \bauthor{\bsnm{Kanamori},~\bfnm{Takafumi}\binits{T.}}
(\byear{2012}).
\btitle{Density ratio estimation in machine learning}.
\bpublisher{Cambridge University Press}.
\end{bbook}
\endbibitem

\bibitem{tai2010correlations}
\begin{barticle}[author]
\bauthor{\bsnm{Tai},~\bfnm{Amos~PK}\binits{A.~P.}}, \bauthor{\bsnm{Mickley},~\bfnm{Loretta~J}\binits{L.~J.}} \AND \bauthor{\bsnm{Jacob},~\bfnm{Daniel~J}\binits{D.~J.}}
(\byear{2010}).
\btitle{Correlations between fine particulate matter (PM2. 5) and meteorological variables in the United States: Implications for the sensitivity of PM2. 5 to climate change}.
\bjournal{Atmospheric environment}
\bvolume{44}
\bpages{3976--3984}.
\end{barticle}
\endbibitem

\bibitem{topaloglu2021pursuit}
\begin{barticle}[author]
\bauthor{\bsnm{Topaloglu},~\bfnm{Mustafa~Y}\binits{M.~Y.}}, \bauthor{\bsnm{Morrell},~\bfnm{Elisabeth~M}\binits{E.~M.}}, \bauthor{\bsnm{Rajendran},~\bfnm{Suraj}\binits{S.}} \AND \bauthor{\bsnm{Topaloglu},~\bfnm{Umit}\binits{U.}}
(\byear{2021}).
\btitle{In the pursuit of privacy: the promises and predicaments of federated learning in healthcare}.
\bjournal{Frontiers in Artificial Intelligence}
\bpages{147}.
\end{barticle}
\endbibitem

\bibitem{tzeng2017adversarial}
\begin{binproceedings}[author]
\bauthor{\bsnm{Tzeng},~\bfnm{Eric}\binits{E.}}, \bauthor{\bsnm{Hoffman},~\bfnm{Judy}\binits{J.}}, \bauthor{\bsnm{Saenko},~\bfnm{Kate}\binits{K.}} \AND \bauthor{\bsnm{Darrell},~\bfnm{Trevor}\binits{T.}}
(\byear{2017}).
\btitle{Adversarial discriminative domain adaptation}.
In \bbooktitle{Proceedings of the IEEE conference on computer vision and pattern recognition}
\bpages{7167--7176}.
\end{binproceedings}
\endbibitem

\bibitem{van2014asymptotically}
\begin{barticle}[author]
\bauthor{\bparticle{van~de} \bsnm{Geer},~\bfnm{Sara}\binits{S.}}, \bauthor{\bsnm{B{\"u}hlmann},~\bfnm{Peter}\binits{P.}}, \bauthor{\bsnm{Ritov},~\bfnm{Ya’acov}\binits{Y.}} \AND \bauthor{\bsnm{Dezeure},~\bfnm{Ruben}\binits{R.}}
(\byear{2014}).
\btitle{On asymptotically optimal confidence regions and tests for high-dimensional models}.
\bjournal{The Annals of Statistics}
\bvolume{42}
\bpages{1166--1202}.
\end{barticle}
\endbibitem

\bibitem{vershynin2010introduction}
\begin{barticle}[author]
\bauthor{\bsnm{Vershynin},~\bfnm{Roman}\binits{R.}}
(\byear{2010}).
\btitle{Introduction to the non-asymptotic analysis of random matrices}.
\bjournal{arXiv preprint arXiv:1011.3027}.
\end{barticle}
\endbibitem

\bibitem{wang2025supplements}
\begin{barticle}[author]
\bauthor{\bsnm{Wang},~\bfnm{Zhenyu}\binits{Z.}}, \bauthor{\bsnm{B{\"u}hlmann},~\bfnm{Peter}\binits{P.}} \AND \bauthor{\bsnm{Guo},~\bfnm{Zijian}\binits{Z.}}
(\byear{2025}).
\btitle{Supplement to ``Distributionally Robust Learning For Multi-source Unsupervised Domain Adaptation'''}.
\end{barticle}
\endbibitem

\bibitem{weiss2016survey}
\begin{barticle}[author]
\bauthor{\bsnm{Weiss},~\bfnm{Karl}\binits{K.}}, \bauthor{\bsnm{Khoshgoftaar},~\bfnm{Taghi~M}\binits{T.~M.}} \AND \bauthor{\bsnm{Wang},~\bfnm{DingDing}\binits{D.}}
(\byear{2016}).
\btitle{A survey of transfer learning}.
\bjournal{Journal of Big data}
\bvolume{3}
\bpages{1--40}.
\end{barticle}
\endbibitem

\bibitem{xia2021chebyshev}
\begin{barticle}[author]
\bauthor{\bsnm{Xia},~\bfnm{Yong}\binits{Y.}}, \bauthor{\bsnm{Yang},~\bfnm{Meijia}\binits{M.}} \AND \bauthor{\bsnm{Wang},~\bfnm{Shu}\binits{S.}}
(\byear{2021}).
\btitle{Chebyshev center of the intersection of balls: complexity, relaxation and approximation}.
\bjournal{Mathematical Programming}
\bvolume{187}
\bpages{287--315}.
\end{barticle}
\endbibitem

\bibitem{xiong2023distributionally}
\begin{barticle}[author]
\bauthor{\bsnm{Xiong},~\bfnm{Xin}\binits{X.}}, \bauthor{\bsnm{Guo},~\bfnm{Zijian}\binits{Z.}} \AND \bauthor{\bsnm{Cai},~\bfnm{Tianxi}\binits{T.}}
(\byear{2023}).
\btitle{Distributionally robust transfer learning}.
\bjournal{arXiv preprint arXiv:2309.06534}.
\end{barticle}
\endbibitem

\bibitem{xu2003solution}
\begin{barticle}[author]
\bauthor{\bsnm{Xu},~\bfnm{Sheng}\binits{S.}}, \bauthor{\bsnm{Freund},~\bfnm{Robert~M}\binits{R.~M.}} \AND \bauthor{\bsnm{Sun},~\bfnm{Jie}\binits{J.}}
(\byear{2003}).
\btitle{Solution methodologies for the smallest enclosing circle problem}.
\bjournal{Computational Optimization and Applications}
\bvolume{25}
\bpages{283--292}.
\end{barticle}
\endbibitem

\bibitem{xu2018impact}
\begin{barticle}[author]
\bauthor{\bsnm{Xu},~\bfnm{Yanling}\binits{Y.}}, \bauthor{\bsnm{Xue},~\bfnm{Wenbo}\binits{W.}}, \bauthor{\bsnm{Lei},~\bfnm{Yu}\binits{Y.}}, \bauthor{\bsnm{Zhao},~\bfnm{Yang}\binits{Y.}}, \bauthor{\bsnm{Cheng},~\bfnm{Shuiyuan}\binits{S.}}, \bauthor{\bsnm{Ren},~\bfnm{Zhenhai}\binits{Z.}} \AND \bauthor{\bsnm{Huang},~\bfnm{Qing}\binits{Q.}}
(\byear{2018}).
\btitle{Impact of meteorological conditions on PM2. 5 pollution in China during winter}.
\bjournal{Atmosphere}
\bvolume{9}
\bpages{429}.
\end{barticle}
\endbibitem

\bibitem{zhang2020coping}
\begin{barticle}[author]
\bauthor{\bsnm{Zhang},~\bfnm{Jingzhao}\binits{J.}}, \bauthor{\bsnm{Menon},~\bfnm{Aditya}\binits{A.}}, \bauthor{\bsnm{Veit},~\bfnm{Andreas}\binits{A.}}, \bauthor{\bsnm{Bhojanapalli},~\bfnm{Srinadh}\binits{S.}}, \bauthor{\bsnm{Kumar},~\bfnm{Sanjiv}\binits{S.}} \AND \bauthor{\bsnm{Sra},~\bfnm{Suvrit}\binits{S.}}
(\byear{2020}).
\btitle{Coping with label shift via distributionally robust optimisation}.
\bjournal{arXiv preprint arXiv:2010.12230}.
\end{barticle}
\endbibitem

\bibitem{zhang2017cautionary}
\begin{barticle}[author]
\bauthor{\bsnm{Zhang},~\bfnm{Shuyi}\binits{S.}}, \bauthor{\bsnm{Guo},~\bfnm{Bin}\binits{B.}}, \bauthor{\bsnm{Dong},~\bfnm{Anlan}\binits{A.}}, \bauthor{\bsnm{He},~\bfnm{Jing}\binits{J.}}, \bauthor{\bsnm{Xu},~\bfnm{Ziping}\binits{Z.}} \AND \bauthor{\bsnm{Chen},~\bfnm{Song~Xi}\binits{S.~X.}}
(\byear{2017}).
\btitle{Cautionary tales on air-quality improvement in Beijing}.
\bjournal{Proceedings of the Royal Society A: Mathematical, Physical and Engineering Sciences}
\bvolume{473}
\bpages{20170457}.
\end{barticle}
\endbibitem

\bibitem{zhang2024minimax}
\begin{barticle}[author]
\bauthor{\bsnm{Zhang},~\bfnm{Yi}\binits{Y.}}, \bauthor{\bsnm{Huang},~\bfnm{Melody}\binits{M.}} \AND \bauthor{\bsnm{Imai},~\bfnm{Kosuke}\binits{K.}}
(\byear{2024}).
\btitle{Minimax Regret Estimation for Generalizing Heterogeneous Treatment Effects with Multisite Data}.
\bjournal{arXiv preprint arXiv:2412.11136}.
\end{barticle}
\endbibitem

\bibitem{zhuang2020comprehensive}
\begin{barticle}[author]
\bauthor{\bsnm{Zhuang},~\bfnm{Fuzhen}\binits{F.}}, \bauthor{\bsnm{Qi},~\bfnm{Zhiyuan}\binits{Z.}}, \bauthor{\bsnm{Duan},~\bfnm{Keyu}\binits{K.}}, \bauthor{\bsnm{Xi},~\bfnm{Dongbo}\binits{D.}}, \bauthor{\bsnm{Zhu},~\bfnm{Yongchun}\binits{Y.}}, \bauthor{\bsnm{Zhu},~\bfnm{Hengshu}\binits{H.}}, \bauthor{\bsnm{Xiong},~\bfnm{Hui}\binits{H.}} \AND \bauthor{\bsnm{He},~\bfnm{Qing}\binits{Q.}}
(\byear{2020}).
\btitle{A comprehensive survey on transfer learning}.
\bjournal{Proceedings of the IEEE}
\bvolume{109}
\bpages{43--76}.
\end{barticle}
\endbibitem

\end{thebibliography}

\newpage
\appendix
\setcounter{page}{1}

\begin{center}
    \Large Appendix to ``Distributionally Robust Learning For Multi-source Unsupervised Domain Adaptation''
\end{center}
The Appendix is organized as follows. 

\begin{itemize}
    \item Appendix \ref{sec: appendix-method_theory} contains additional methods and theoretical results.
    \begin{itemize}
        \item Subsection \ref{subsec: appendix-extension} extends the reward function to compare $f(\cdot)$ with any deterministic benchmark function and establish the identification result for the distributionally robust model defined with this extended reward function.
        \item Subsection \ref{subsec: appendix-fairness} demonstrates the connection between our distributionally robust model with minimax fairness where there is no covariate shift across source and target domains.
        \item Subsection \ref{subsec: appendix-prop} demonstrates that a sufficiently accurate prior information $\HH$ will recover the actual conditional outcome model $f^\QQ(x) = \E_{\QQ}[Y|X=x]$, 
        \item Subsection \ref{sec: high-d regime} considers the high-dimensional setting to illustrate our main theoretical results regarding the convergence rate of the DRoL estimators.
        \item Subsection \ref{subsec: alg-plug-in} provides the algorithm of the plug-in DRoL estimator.
    \end{itemize}
    \item Appendix \ref{sec: appendix-exp} contains the additional numerical results and real data applications. 
    \begin{itemize}
        \item Subsection \ref{subsec: appendix-simu-fairness} conducts a simulation study to compare the proposed reward function and the squared error for the minimax fairness in the no covariate shift setting.
        \item Subsection \ref{subsec: appendix-exp-highd} demonstrates that DRoL is able to capture shared associations in the high-dimensional setting. 
    \end{itemize}  
    \item Appendix \ref{sec: appendix-proof-main} contains proofs for the results presented in the main body of the paper. 
    \item The proofs of technical lemmas are deferred to Appendix \ref{sec: proof-lemmas}.
\end{itemize}

\section{Additional Methods and Theories}
\label{sec: appendix-method_theory}

\subsection{Extension of Robust Prediction Models}
\label{subsec: appendix-extension}
We generalize the reward function $\RR_\TT(f)$ introduced in \eqref{eq: reward} as follows. By introducing a predefined deterministic function $h(\cdot)$, we define:
\begin{equation}
    \RR_{\TT}(f; h) := \E_{(X,Y)\sim\TT}[(Y-h(X))^2 - (Y - f(X))^2].
    \label{eq: reward general}
\end{equation}
It is clear that the original reward function $\RR_\TT(f)$ is a special case of $\RR_\TT(f; h)$ with $h \equiv 0$, i.e., $\RR_\TT(f) = \RR_\TT(f; 0)$. The generalized reward function $\RR_\TT(f; h)$ compares the prediction accuracy of the model $f(\cdot)$ against that of the predefined model $h(\cdot)$. Consequently, $\RR_\TT(f; h)$ measures the variance explained by the model $f$ relative to the model $h$. The choice of $h$ is typically informed by expert knowledge. As an alternative example, when a small number of labeled target observations is available, the work \citep{xiong2023distributionally} focuses on the linear regression setting and selects $h(\cdot)$ as the regression model fitted on the limited labeled target data.

With the generalized uncertainty set $\C^\prime (\QQ_X, \HH)$, defined in \eqref{eq: region-Q H general}, and reward function $\RR_\TT(f;h)$, we define the distributionally robust prediction model as follows:
\begin{equation}
    f_{\HH;h}^* := \argmax_{f\in \FF} \min_{\TT \in \C^\prime(\QQ_X, \HH)} \RR_\TT (f; h).
    \label{eq: f_star H general}
\end{equation}
The following theorem establishes that the generalized model $f_{\HH;h}^*$ also admits a weighted average formulation. 
\begin{Theorem}
Suppose that the function class $\FF$ is convex with $\f{l} \in \FF$ for all $l\in[L]$ and $\HH$ is a convex subset of $\Delta^L$. For any deterministic function $h: \mathcal{X} \to \R$, $f_{\HH;h}^*$ defined in \eqref{eq: f_star H general} is identified as:
    \begin{equation}
        f_{\HH;h}^* = \sum_{l=1}^L q_l^* \cdot \f{l} \quad \textrm{with} \quad 
        q^* = \argmin_{q\in \HH}q^\intercal \Gamma^h q 
        \label{eq: identification - extension}
    \end{equation}
where $\Gamma_{k,l}^h = \E_{\QQ_X}[\left(\f{k}(X)-h(X)\right)\left(\f{l}(X) - h(X)\right)]$ for $k,l\in [L]$.
    \label{thm: identification general}
\end{Theorem}
Compared to Theorem \ref{thm: identification}, the primary distinction in Theorem \ref{thm: identification general} lies in the definition of $\Gamma^h$. Specifically, when $h(\cdot) \equiv 0$, $\Gamma^h$ in \eqref{eq: identification - extension} reduces to $\Gamma$ in \eqref{eq: identification}. Geometrically, the generalized model $f_{\HH;h}^*$ represents the point within the $\HH$-constrained set that is closest to the model $h(\cdot)$, whereas the original model $f_\HH^*$ corresponds to the point within the $\HH$-constrained set that is closest to the origin. Notably, the identification theorem for the predefined function $h$ is also addressed by the work of \citep{xiong2023distributionally}, which primarily focuses on the linear case.

\subsection{Connection to Minimax Fairness} 
\label{subsec: appendix-fairness}
Ensuring fairness is a critical aspect of designing machine learning algorithms, particularly to prevent models that perform well on average from disadvantaging certain sub-populations. To address such concerns, recent works \cite{diana2021minimax, martinez2020minimax} have introduced minimax fairness approaches that focus on optimizing the worst-case performance across different source groups (e.g., various demographic groups). For example, if data $(X,Y)$ are collected from $L$ distinct sources, the minimax-fair model is obtained by solving:
\[
f^{\rm fair} = \argmin_{f\in \FF} \max_{1\leq l\leq L}\E_{(X,Y)\sim \PP{l}}[\ell(X,Y;f)],
\]
where \(\FF\) is a predetermined function class and $\ell(x,y;f)$ is the chosen loss function. 

Under the multi-source unsupervised domain adaptation (MSDA) framework, when there is no covariate shift between the source and target domains (i.e., when \(\QQ_X = \PP{l}X\) for all l), our proposed robust prediction model in \eqref{eq: f_star H} essentially coincides with the minimax fairness formulation:
\[
\begin{aligned}
    f^{*} &= \argmax_{f\in \FF}\max_{\TT\in \C(\QQ_X)} \RR_\TT(f) \\
    &= \argmax_{f\in \FF}\max_{1\leq l\leq L}\E_{(X,Y)\sim (\QQ_X, \PP{l}_{Y|X})}[-\ell(X,Y;f)]\\
    &= \argmin_{f\in \FF}\max_{1\leq l\leq L} \E_{(X,Y)\sim \PP{l}}[\ell(X,Y;f)],
\end{aligned}
\]
with $\ell(x,y;f) = (y-f(x))^2 - y^2.$

\subsection{Accurate Prior Information Recovers the Actual Model}
\label{subsec: appendix-prop}
We consider the scenario that $\QQ_{Y|X} = \sum_{l=1}^L q^0_l \cdot \PP{l}_{Y|X}$ for some (unknown) $q^0 \in \Delta^L$ and the corresponding conditional outcome $f^\QQ(x) = \E_{\QQ}[Y|X=x]$ admits the form 
$\sum_{l=1}^L q^0_l\cdot \f{l}(x).$
The following proposition establishes that a sufficiently accurate set $\HH$ can lead to the identification of $f^\QQ$. 
\begin{Proposition} Suppose that $\QQ_{Y|X} = \sum_{l=1}^L q^0_l \cdot \PP{l}_{Y|X}$ for some $q^0 \in \Delta^L.$
If $\max_{q\in \HH}\|q - q^0\|_2 \leq \rho$, then $
    \left(\E_{\QQ_X}[(\fH(X)-f^\QQ(X))^2]\right)^{1/2} \leq \rho \sqrt{L} \max_{l\in [L]}\|\f{l}\|_{\ell_2(\QQ)} $.
    \label{prop: shrink H}
\end{Proposition}

The proposition above demonstrates that $\fH$ converges to the true conditional outcome model $f^\QQ$ when the subset $\HH$ converges to the true mixing weight $q^0$. If we lack prior information about the potential mixture of the conditional outcome distribution, the most natural and secure approach is to fit a robust prediction model that guarantees all potential mixture weights by setting $\HH = \Delta^L$. However, with richer prior information, we can obtain a {better} predictive model or even recover the true conditional outcome model, but at the risk of reduced robustness if the prior information is incorrect.

\subsection{High-dimensional Sparse Linear Model}
\label{subsec: appendix-density-bayes}
\label{sec: high-d regime}
In this subsection, we focus on the high-dimensional sparse regression setting, which acts as an essential and illustrative example of our main theoretical results regarding the convergence rate of the DRoL estimators. In this setting, we assume that the conditional outcome models are linear models with $\f{l}(x) = x^\intercal \beta^{(l)}$ for $\beta^{(l)} \in \R^p$ and each source group $l\in [L]$. We also assume that the density ratio model $\w{l}(x)$ can be correctly characterized by sparse logistic regression, which will be detailed presented later.

\subsubsection{Methodology}
For each source group $l\in [L]$, we use LASSO to estimate $\beta^{(l)}$ and obtain the estimators $\{\widehat{\beta}^{(l)}, \widehat{\beta}^{(l)}_\A,  \widehat{\beta}^{(l)}_\B\}$. They are fitted on the full data and subset data belonging to $\A_l$ and $\B_l$, respectively. $\widehat{\beta}^{(l)}$ (with full data) is defined as follows:
{
\begin{equation}
    \widehat{\beta}^{(l)} = \argmin_{\beta \in \R^p}\left\{\frac{1}{2n_l} \sum_{i=1}^{n_l} (\Y{l}_i - X^{(l)\intercal}_{i} \beta)^2 + 
A\sqrt{\frac{\log p}{n_l}} \sum_{j=2}^p \frac{\|\X{l}_{\cdot j}\|_2}{\sqrt{n_l}} |\beta_j| \right\},
\label{eq: beta_full}
\end{equation}
} where the first column of $X^{(l)}$ is set as the constant 1, $X^{(l)}_{\cdot j}$ denotes the $j$-th column of $X^{(l)}$, and $A>\sqrt{2}$ is a pre-specified constant. Similarly, we can define $\widehat{\beta}^{(l)}_\A$ and $\widehat{\beta}^{(l)}_\B$ on the split data subset $\A_l$ and $\B_l$, respectively.
\begin{equation}
    \begin{aligned}
    \widehat{\beta}^{(l)}_\A &= \argmin_{\beta \in \R^p}\left\{\frac{1}{2|\A_l|} \sum_{i=1}^{|\A_l|} (\Y{l}_{\A,i} - X^{(l)\intercal}_{\A, i} \beta)^2 + 
    A\sqrt{\frac{\log p}{|\A_l|}} \sum_{j=2}^p \frac{\|\X{l}_{\A, \cdot j}\|_2}{\sqrt{|\A_l|}} |\beta_j| \right\}, \\
    \widehat{\beta}^{(l)}_\B &= \argmin_{\beta \in \R^p}\left\{\frac{1}{2n_l} \sum_{i=1}^{n_l} (\Y{l}_{\B,i} - X^{(l)\intercal}_{\B, i} \beta)^2 + 
    A\sqrt{\frac{\log p}{n_l}} \sum_{j=2}^p \frac{\|\X{l}_{\B, \cdot j}\|_2}{\sqrt{n_l}} |\beta_j| \right\},
\end{aligned}
\label{eq: beta_split}
\end{equation}
where $\{\X{l}_\A, \Y{l}_\A\}$ denotes data belonging to the subset $\A_l$ in  $\{\X{l}, \Y{l}\}$, and similar representation holds for $\{\X{l}_\B, \Y{l}_\B\}$.

As for the density ratio models $\{\w{l}(x)\}_{l\in [L]}$, we start with the introduction about the estimation approach by Bayes Formula, and then we will present the details for the high-dimensional setting.
Notably, our DRoL framework is compatible with various existing density ratio estimations; see \cite{sugiyama2012density} for a thorough review.

For each source group $l \in [L]$, we merge its covariates observations $\{{X}^{(l)}_i\}_{i\in [n_l]}$ with the target covariates $\{\XQ_j\}_{j\in [N]}$ and denote the merged covariates as $\{\widetilde{X}^{(l)}_i\}_{i\in [n_l + N]}$, where the first $n_l$ observations are from the $l$-th source group and the remaining $N$ observations are from the target group. We also define an indicator random variable $G^{(l)}_i\in \{0,1\}$ for $i \in [n_l+N]$: $\G{l}_i = 0$ if $\widetilde{X}^{(l)}_i$ is drawn from the $l$-th source group and $\G{l}_i = 1$ if $\widetilde{X}^{(l)}_i$ is drawn from the target distribution. The density of $l$-th source covariates and target covariates are written as $d\PP{l}_X(x) = p(x|\G{l}_i=0)$ and $d\QQ_X(x) = p(x|\G{l}_i=1)$. We apply the Bayes formula and obtain the following expression of the density ratio, 
\[
\w{l}(x) = \frac{d\QQ_X(x)}{d\PP{l}_X(x)} =  
\frac{p(\G{l}_i=0)}{p(\G{l}_i=1)} \frac{p(\G{l}_i = 1|x)}{p(\G{l}_i = 0|x)}.
\]
The ratio ${p(\G{l}_i=0)}/{p(\G{l}_i=1)}$ can be approximated by the ratio of sample size. 
The class posterior probability $p(\G{l}_i|x)$ may be approximated using a probabilistic classifier, like the logistic regression and the random forests classifier. Here, we consider the binary-class logistic regression with 
\begin{equation}
p(\G{l}_i = 1|x) = h(x^\intercal \gamma^{(l)})\quad \textrm{with} \quad h(z) = \exp(z)/(1+\exp(z))
\label{eq: density ratio}
\end{equation}
where $\gamma^{(l)}$ denotes the regression vector.

Similarly to estimation of $\beta^{(l)}$, we randomly partition each source data of the $l$-th group into two disjoint subsets $\A_l$ and $\B_l$. Then, we estimate $\gamma^{(l)}$ by $\widehat{\gamma}^{(l)}_\A$ and $\widehat{\gamma}^{(l)}_\B$ based on the split data $\A_l$ and $\B_l$, respectively, together with the target covariates. Examples include the MLE estimator in low dimensions and the penalized log-likelihood estimators in high dimensions. The details of the latter case will be presented soon. 
We then further estimate the density ratio as follows: for $l\in [L]$,
\begin{equation}
\hw{l}_\A(x) = \frac{|\A_l|}{N} \frac{h(x^\intercal \widehat{\gamma}^{(l)}_\A)}{1-h(x^\intercal \widehat{\gamma}^{(l)}_\A)}, \quad \textrm{and} \quad 
\hw{l}_\B(x) = \frac{n_l}{N} \frac{h(x^\intercal \widehat{\gamma}^{(l)}_\B)}{1-h(x^\intercal \widehat{\gamma}^{(l)}_\B)}.
    \label{eq: density ratio estimation}
\end{equation}
It is important to highlight that the consistency of our proposed DRoL estimator does not rely on the correct specification of the density ratio model, such as \eqref{eq: density ratio}. However, a correctly specified density ratio model leads to a faster convergence rate.

We start with the definition of $\widehat{\gamma}^{(l)}_\A$, and then we define $\widehat{\gamma}^{(l)}_\B$ following the similar argument.
{
\begin{equation}
\begin{aligned}
\widehat{\gamma}^{(l)}_\A = \argmin_{\gamma \in \R^p}\left\{-\frac{1}{|\A_l|+N} \sum_{i=1}^{|\A_l|+N} \left[\log(1+\exp(\widetilde{X}^{(l)\intercal}_{\A, i}\gamma^{(l)})) - \G{l}_{\A,i}(\widetilde{X}^{(l)\intercal}_{\A, i}\gamma^{(l)})\right]\right. \\
\left.+\lambda^{(l)}_\A \sum_{j=2}^p \frac{\|\widetilde{X}^{(l)}_{\A, \cdot j}\|_2}{\sqrt{|\A_l|+N}} |\gamma_j|\right\},
\label{eq: penalized MLE}
\end{aligned}
\end{equation}}
where $\widetilde{X}^{(l)}_\A = (X^{(l)\intercal}_\A, X_\QQ^\intercal)^\intercal$, $\widetilde{X}^{(l)}_{\A,\cdot j}$ denotes the $j$-th column of $\widetilde{X}^{(l)}_\A$, the first column of $\widetilde{X}^{(l)}_\A$ is set as the constant 1, and the tuning parameter $\lambda^{(l)}_\A$ is of the order $\sqrt{\log p/(|\A_l|+N)}$. The indicator random variable $G^{(l)}_{\A,i}$ is set as $G^{(l)}_{\A,i}=0$ if $\widetilde{X}^{(l)}_{\A,i}$ is drawn from the $l$-th source group's subset $\A_l$ and is set as $G^{(l)}_{\A,i}=1$ if  $\widetilde{X}^{(l)}_{\A,i}$ is drawn from the target distribution. By replacing $\A_l$ with $\B_l$ in \eqref{eq: penalized MLE}, we can obtain the density ratio estimator $\widehat{\gamma}^{(l)}_\B $ as follows:
{
\begin{equation}
\begin{aligned}
    \widehat{\gamma}^{(l)}_\B = \argmin_{\gamma \in \R^p}\left\{-\frac{1}{n_l+N} \sum_{i=1}^{n_l+N} \left[\log(1+\exp(\widetilde{X}^{(l)\intercal}_{\B, i}\gamma^{(l)})) - \G{l}_{\B,i}(\widetilde{X}^{(l)\intercal}_{\B, i}\gamma^{(l)})\right] \right.\\
    \left.+ \lambda^{(l)}_\B \sum_{j=2}^p \frac{\|\widetilde{X}^{(l)}_{\B, \cdot j}\|_2}{\sqrt{n_l+N}} |\gamma_j|\right\},
\label{eq: penalized MLE-2}
\end{aligned}
\end{equation}}
where $\widetilde{X}^{(l)}_\B = (X^{(l)\intercal}_\B, X_\QQ^\intercal)^\intercal$, $\widetilde{X}^{(l)}_{\B,\cdot j}$ denotes the $j$-th column of $\widetilde{X}^{(l)}_\B$, the first column of $\widetilde{X}^{(l)}_\B$ is set as the constant 1, and the tuning parameter $\lambda^{(l)}_\B$ is of the order $\sqrt{\log p/(n_l+N)}$. The indicator random variable $G^{(l)}_{\B,i}$ is set as $G^{(l)}_{\B,i}=0$ if $\widetilde{X}^{(l)}_{\B,i}$ is drawn from the $l$-th source group's subset $\B_l$ and is set as $G^{(l)}_{\B,i}=1$ if  $\widetilde{X}^{(l)}_{\B,i}$ is drawn from the target distribution.

\subsubsection{Theoretical analysis}
In this subsection, we present and compare the convergence rate of naive plug-in DRoL estimator $\tfH$ and bias-corrected counterpart $\hfH$, along with the technical assumptions. 

Under regularity conditions, the LASSO estimators $\{\widehat{\beta}^{(l)}, \widehat{\beta}^{(l)}_\A,  \widehat{\beta}^{(l)}_\B\}$ defined in \eqref{eq: beta_full} and \eqref{eq: beta_split} have been shown to satisfy the following rate of convergence: with probability greater than $1-p^{-c}-\exp(-c n_l)$ for some positive constant $c>0$, 
\begin{equation}
    \max\left\{\|\widehat{\beta}^{(l)} - \beta^{(l)}\|_2, \|\widehat{\beta}^{(l)}_\A - \beta^{(l)}\|_2, \|\widehat{\beta}^{(l)}_\B - \beta^{(l)}\|_2\right\} \lesssim \sqrt{s_\beta \log p/n_l}; 
    \label{eq: beta-rate}
\end{equation}
see Theorem 7.2 in \cite{bickel2009simultaneous} and Lemma 6.10 in \cite{buhlmann2011statistics}.

We follow the literature results and assume that with a high probability greater than $1-p^{-c}-\exp(-c (n_l+n_{\QQ}))$ for some positive constant $c>0$, 
the estimators $\widehat{\gamma}^{(l)}_\A, \widehat{\gamma}^{(l)}_\B$ defined in \eqref{eq: penalized MLE} and \eqref{eq: penalized MLE-2} satisfy
\begin{equation}
    \max\left\{\|\widehat{\gamma}^{(l)}_\A - \gamma^{(l)}\|_2, \|\widehat{\gamma}^{(l)}_\B - \gamma^{(l)}\|_2\right\} \lesssim \sqrt{s_\gamma \log p / (n+ N)};
    \label{eq: gamma-rate}
\end{equation}
see Section 4.4 of \cite{negahban2012unified} and Theorem 9 in \cite{huang2012estimation} for detailed statements.

We introduce the following assumptions for the high-dimensional setting. 
\begin{Assumption}
The target covariates observations $\{\XQ_j\}_{j\in [N]}$ are i.i.d p-dimensional sub-Gaussian random vectors.
For every group $l\in [L]$, with probability larger than $1-p^{-c_0}$ for some positive constant $c_0>0$, the density ratio model $h(\cdot)$ defined in \eqref{eq: density ratio} satisfies $\min\{h(\widetilde{X}_{i}^{(l)\intercal}\gamma^{(l)}), 1-h(\widetilde{X}_{i}^{(l)\intercal}\gamma^{(l)})\}\geq c_{\rm min}$,
with $\widetilde{X}^{(l)} = ((X^{(l)})^{\intercal}, X_\QQ^\intercal)^\intercal$ and $c_{\rm min}\in (0,\frac{1}{2})$ denoting some positive constant.
    \label{ass: linear}
\end{Assumption}
{Assumption} 4 requires the class conditional probability to be bounded away from $0$ and $1$, which is part of the assumption used to establish the consistency of the high-dimensional density ratio model. Such an assumption is commonly assumed for the theoretical analysis of the high-dimensional logistic regression; see condition (iv) of Theorem 3.3 of \cite{van2014asymptotically} and Assumption 6 in \cite{athey2018approximate}.

We now compare the convergence rate for $\tfH$ and $\hfH$ in the high-dimensional linear setting.

\begin{Corollary}
    Suppose Assumptions 1 and \ref{ass: linear} hold, the sparsity levels $s_{\beta}:= \max_{l\in[L]}\|\beta^{(l)}\|_0$ and $s_{\gamma}:= \max_{l\in[L]}\|\gamma^{(l)}\|_0$ satisfy $\sqrt{s_\beta \log p/n} \to 0$ and $\sqrt{s_\gamma \log p / (n+ N)} \to 0$, and the estimators $\{\widehat{\beta}^{(l)},\widehat{\beta}^{(l)}_\A, \widehat{\beta}^{(l)}_\B\}$ and $\{\widehat{\gamma}^{(l)}_\A, \widehat{\gamma}^{(l)}_\B\}$ satisfy \eqref{eq: beta-rate} and \eqref{eq: gamma-rate}, respectively. Then with probability larger than $1-p^{-c}-\exp(-cn)$ for some positive constant $c>0$, 
    $\delta_n\lesssim \sqrt{s_\beta \log p/n}$ and $ {\eta}_\omega \lesssim \sqrt{s_\gamma \log p / (n+N)}$.
    Additionally, we assume that $M = \max_{l \in [L]}\|\beta^{(l)}\|_2 \gg \sqrt{s_\beta \log p /n}$,
    then with probability larger than $1-p^{-c}-\exp(-cn)-{1}/{t^2}$ for some $c>0$ and $t>1$, the bias-corrected DRoL estimator satisfies 
\begin{equation}   
\begin{aligned}
    &\left\|\hfH - \fH\right\|_{\ell_2(\QQ)} 
    \lesssim 
    \sqrt{\frac{s_\beta \log p}{n}} + tM\cdot\min\left\{
    M\sqrt{\frac{s_\beta \log p}{n}} \cdot r_n + \frac{M^2}{\sqrt{N}}\;,\; \rho_\HH\right\} \\
    &\quad \textrm{where} \quad r_n = 
    \frac{1}{\sqrt{s_\beta \log p}} + \frac{1}{\sqrt{n\wedge N}} + \frac{1}{M}\sqrt{\frac{s_\beta \log p}{n}} +  \sqrt{\frac{s_\gamma \log p}{n+N}}.
\end{aligned}
\end{equation} 
In contrast, with probability larger than $1-p^{-c}-\exp(-cn)-1/{t^2}$ for some $c>0$ and $t>1$, the plug-in DRoL estimator satisfies 
\begin{equation*}
    \left\|\tfH - \fH\right\|_{\ell_2(\QQ)} \lesssim 
    \sqrt{\frac{s_\beta \log p}{n}} + tM\cdot\min\left\{M\sqrt{\frac{s_\beta \log p}{n}} +\frac{M^2}{\sqrt{N}}\;,\;\rho_\HH\right\}.
\end{equation*}
\label{cor: high-dim}
\end{Corollary}
By comparing the bias-corrected DRoL and the plug-in DRoL, we observe that the bias-correction step shrinks the inner $\sqrt{s_\beta \log p/n}$ by a factor $r_n\rightarrow 0$. {We now point out a regime where the bias-corrected estimator improves the rate of convergence for the high-dimensional setting: the norm $M = \max_{l \in [L]}\|\beta^{(l)}\|_2$ diverges to infinity with $n$ and $p$, the diameter $\rho_{\HH}$ is larger than $M\sqrt{{s_\beta \log p}/{n}}$, and the sample size of the unlabelled data in the target domain is larger than $n M^2/(s_\beta \log p).$} 

Considering the setting where $M$ is at the constant order, and both the diameter $\rho_\HH$ and target sample size $N$ are sufficiently large, satisfying $n/(s_\beta \log p)\lesssim N$ and $\sqrt{s_\beta \log p/n}\lesssim \rho_\HH$, we observe that both the bias-corrected DRoL estimator $\hfH$ and its plug-in counterpart $\tfH$ achieves the rate of $\sqrt{s_\beta \log p/n}$. Combining Theorem \ref{prop: reward-diff}, we can obtain the convergence rate for their reward function, which is $\sqrt{s_\beta \log p/n}$ as well.

\subsubsection{Extension to low-dimensional regime}
The preceding methodology and theoretical analysis in the high-dimensional regime can be trivially extended to the low-dimensional regression by assuming the dimension $p$ fixed. 

For the methodology part, we apply the MLE to estimate $\beta^{(l)}$ and $\gamma^{(l)}$ for each group $l\in [L]$. We still use the notations $\{\widehat{\beta}^{(l)}, \widehat{\beta}^{(l)}_\A,  \widehat{\beta}^{(l)}_\B\}$, $\{\widehat{\gamma}^{(l)}_\A, \widehat{\gamma}^{(l)}_\B\}$ to represent the estimators for $\beta^{(l)}$ and $\gamma^{(l)}$. However, they are defined without penalization terms as in \eqref{eq: beta_full},\eqref{eq: beta_split},\eqref{eq: penalized MLE}, and \eqref{eq: penalized MLE-2}.

For the theoretical part, since the dimension $p$ is assumed to be fixed, based on Corollary \ref{cor: high-dim}, we derive the following result: both the bias-corrected DRoL estimator and its plug-in counterpart are upper bounded by $\frac{1}{\sqrt{n}} + tM\cdot\min\left\{
    \frac{M}{\sqrt{n}} + \frac{M^2}{\sqrt{N}}\;,\; \rho_\HH\right\}$. 

    Since the dimension $p$ is assumed to be fixed, we have the norm $M=\max_{l\in [L]}\|\beta^{(l)}\|_2$ being at the constant scale as well. Considering the setting where $n \lesssim N$ and $\rho_\HH$ is larger $n^{-1/2}$, based on Corollary \ref{cor: high-dim}, we can establish the convergence rates for both the bias-corrected DRoL estimator $\hfH$ and its plug-in estimator $\tfH$, that they are both upper bounded by $n^{-1/2}$. Moreover, combining the result in Theorem \ref{prop: reward-diff}, we further obtain the convergence rate for the reward function, which is $n^{-1/2}$ as well.

\subsection{Additional Algorithms}
\label{subsec: alg-plug-in}

The plug-in DRoL estimator is encapsulated in Algorithm \ref{alg: plug-in}. 
It is essential to highlight that Algorithm \ref{alg: plug-in} achieves a higher level of privacy preservation compared to its bias-corrected counterpart, Algorithm \ref{alg: cs mm}. During the whole process, it only requires the transmission of the trained machine learning estimators $\{\hf{l}\}_{l\in [L]}$ to the target site. Moreover, it is easier to implement, but its convergence rate may be slower, as discussed in the main body of our work.

\begin{algorithm}[h]
    \DontPrintSemicolon
    \SetAlgoLined
    \SetNoFillComment
    \LinesNotNumbered 
    \caption{Plug-in {DRoL}}
    \KwData{Source data $\{(\X{l}_i, \Y{l}_i)\}_{i\in [n_l]}$ for $l\in[L]$; {target covariates $\{\XQ_j\}_{j\in [N]}$;}
    the convex subset $\HH$ (with the default value $\HH=\Delta^{L}$).}
    \KwResult{Plug-in {DRoL} estimator $\hfH$}

    \For{$l=1,...,L$}{
        Construct the initial machine learning estimators $\hf{l}$ 
    }
    
    \For{$k,l=1,...,L$}{ 
        Compute the plug-in estimator $\tGamma_{k,l} = \frac{1}{N}\sum_{j=1}^{N} \hf{k}(\XQ_j)\hf{l}(\XQ_j)$
    }

    Construct the data-dependent optimal weight as $\widetilde{q} = \argmin_{q\in \mathcal{H}} q^\intercal \tGamma q$

    Return $\widetilde{f}_\HH = \sum_{l=1}^L \widetilde{q}_l \cdot \hf{l}$

    \label{alg: plug-in}
\end{algorithm}

\section{Additional Numerical Results}
\label{sec: appendix-exp}

\subsection{Minimax Fairness}
\label{subsec: appendix-simu-fairness}
As discussed in Section \ref{subsec: appendix-fairness}, when there is no covariate shift between the source and target domains, the proposed robust prediction model in the MSDA setting aligns with the minimax fairness formulation. In this simulation experiment, we consider the no covariate shift setting to study the performance of robust prediction models on each source group.

In this setting, the robust prediction model can be expressed as:
\[
\argmin_{f\in \FF}\max_{\TT\in \C(\QQ_X)}\E_{\TT}[\ell(X,Y;f)] = \argmin_{f\in \FF}\max_{1\leq l\leq L} \E_{\PP{l}}[\ell(X,Y;f)],
\]
where $\ell(X;Y;f)$ is the chosen loss function. For the squared error loss, we use $\ell(x,y;f) = (y-f(x))^2$; for our reward-based approach, we consider $\ell(x,y;f) = (y-f(x))^2 - y^2$. 

\begin{figure}[!ht]
    \centering
    \includegraphics[width=0.5\linewidth]{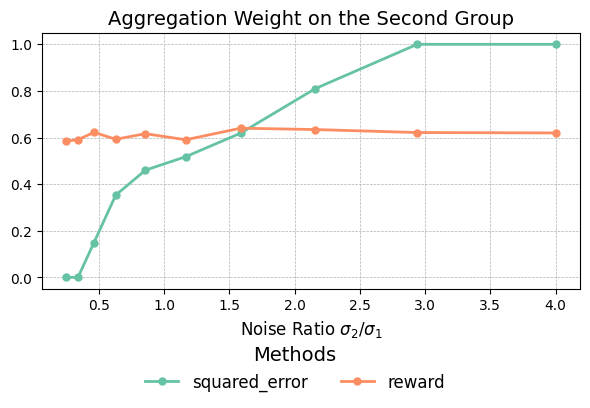}
    \caption{Comparison of robust prediction models under heterogeneous noise conditions. Two groups are considered with varying noise levels, with the noise ratio $\sigma_2/\sigma_1$ ranging from 0.25 to 4. The y-axis represents the distance between robust prediction model $\hat{f}$ and each group's true model $f^{(l)}$ (i.e., $\|\widehat{f} - f^{(l)}\|_{\ell_2}$) for $l=1,2$. The results are averaged over 200 simulations rounds, and the sample size for each source group is fixed at $n_l=2000$.}
    \label{fig:fairness}
\end{figure}

Next, we present a simulation study displaying the computed aggregation weights of source groups for the distributionally robust models using the squared error and our reward function, in scenarios with heterogeneous noise levels across groups. Labeled data are generated from $L$ source domains following the mechanism in \eqref{eq: fl simus}. We consider two groups ($L=2$) with heterogeneous noise levels, where the noise ratio $\sigma_2/\sigma_1$ varies over the interval $[0.25, 4]$. The results, averaged over 200 simulation rounds, are presented in Figure \ref{fig:fairness}.

The results in Figure \ref{fig:fairness} demonstrate that the squared error loss over-prioritizes the noisier group, whereas our reward function mitigates the impact of heterogeneous noise levels.

\subsection{Capturing shared associations in high-dimensional setting} 
\label{subsec: appendix-exp-highd}
Section \ref{subsec: identification} emphasizes that {DRoL} can capture the common factor across different sites. 
We now demonstrate this phenomenon using high-dimensional regressions. We set the group number $L=5$, the dimension $p=200$, and set each group to have the same number of observations, denoted as $n_l = n$ for all $l\in [L]$. We define the base model $f_{\rm base}: \R^p \mapsto \R$ as $f_{\rm base}(x) = \sum_{j=1}^{10} 0.5 x_j$, where only the first ten of $p=200$ variables have non-zero coefficients here.
For $l \in [L]$, we generate the $l$-th group source data following \eqref{eq: source-models} with $\PP{l}_X = \N(0_p, I_p)$, $\f{l}(x) = f_{\rm base}(x) + \sum_{j=11}^{13} \gamma^{(l)}_{j} x_{j}$, and $\eps{l}_i \stackrel{i.i.d}{\sim} \N(0,1)$.
The coefficients $\gamma^{(l)}_j$ are independently drawn from $\N(0,1)$ for $11\leq j\leq 13$. Consequently, the first $10$ coefficients remain the same across each source group, while the $11$-th to $13$-th coefficients are randomly generated to be scattered around $0$.

We consider the no covariate shift setting with $\QQ_X = \N(0_p, I_p)$. For each round of simulation, the target conditional outcome model is generated as: $f^\QQ(x) = f_{\rm base}(x) + \sum_{j=11}^{13} \gamma^\QQ_{j} x_{j}$, where $\gamma^\QQ_j \sim \N(0,1)$ for $11\leq j\leq 13$. As a result, the target data shares the same base model with source groups, but the $11$-th to $13$-th coefficients are randomly generated, not necessarily seen in the source groups. The target data is generated with $N = 10000$ observations. We utilize the LASSO algorithm with the hyperparameters tuned by cross-validation for fitting models $\f{l}$ for $l\in [L]$.

\begin{figure}[!h]
    \centering
    \includegraphics[width=0.7\textwidth]{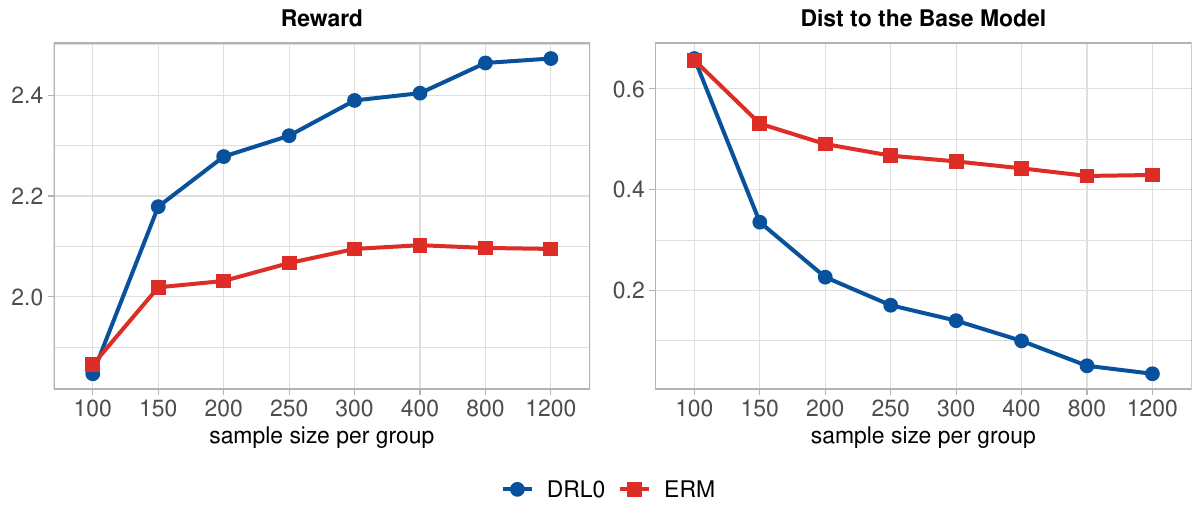}
    \caption{Comparison of \texttt{DRL0} and {\ERM} in high-dimensional setting with sample size per group varied across $\{100,150,200,250,300,400,800,1200\}$. \texttt{DRL0} is implemented by Algorithm \ref{alg: cs mm} with default $\HH=\Delta^L$, no sample splitting, and assuming no covariate shift. {\ERM} stands for the empirical risk minimizer fitted over the pooled source data sets. {Left panel:} the reward evaluated over the target data; { Right panel:} the squared $\ell_2$ distance between the computed coefficients to the base model's coefficients. \texttt{DRL0} is implemented by Algorithm \ref{alg: cs mm} without sample splitting.}
    \label{fig: mix-highd-fig1}
\end{figure}

In the left sub-figure of Figure \ref{fig: mix-highd-fig1}, we can observe that {\DRoLn} consistently outperforms {\ERM} with a higher reward, indicating that {\DRoLn} consistently produces a better prediction model compared to {\ERM}. In the right sub-figure, we compare the squared $\ell_2$ distance of the computed coefficients of {\DRoLn} and {\ERM} to the coefficients of the base model. With an increasing sample size per group, the distance for {\DRoLn} steadily diminishes, approaching zero. This trend indicates that the initially randomly generated coefficients (11th to 13th) are gradually nullified, leading to a successful recovery of the base model. In contrast, the distance associated with {\ERM} remains non-negligible and tends to stabilize at approximately 0.4 as the sample size grows. This analysis confirms that {\DRoLn} successfully captures the shared factors across the groups while effectively discarding the heterogeneous elements. On the contrary, {\ERM} retains those randomly generated coefficients, failing to recover the underlying base model.

\begin{figure}[!h]
    \centering
    \includegraphics[width=0.7\textwidth]{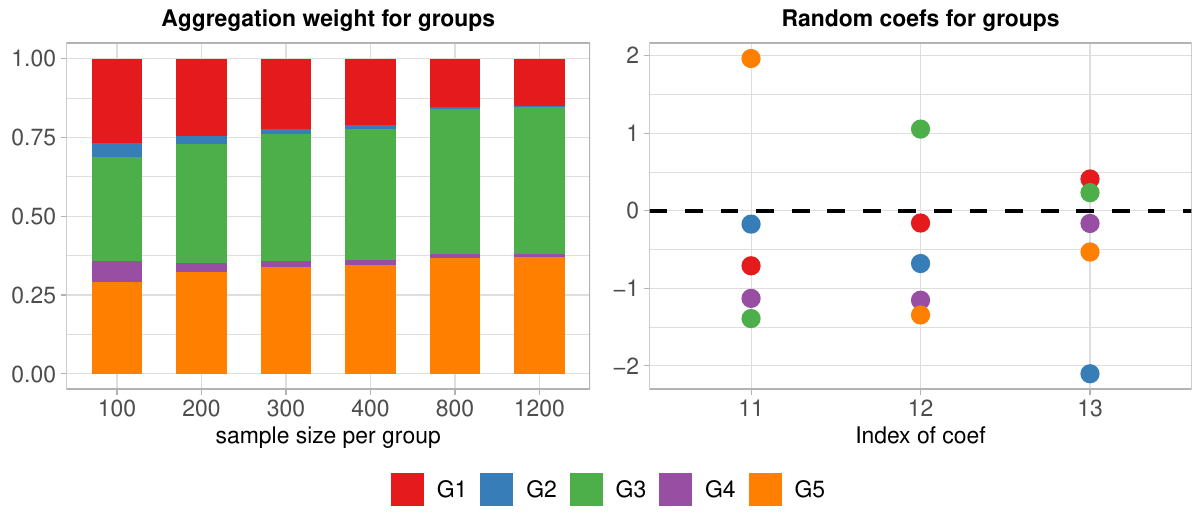}
    \caption{{Left panel:} the computed aggregation weights for the five source groups with sample size per group varied across $\{100,200,400,800,1200\}$; {Right panel:} the actual randomly generated coefficients (11th to 13th) for five groups.}
    \label{fig: mix-highd-fig2}
\end{figure}

The proposed {\DRoLn} provides aggregation weights for each source group. In the left sub-figure of Figure \ref{fig: mix-highd-fig2}, we display the computed weights for the five groups with varying sample sizes per group, ranging from $\{100,200,300,400,800,1200\}$. In the right sub-figure of Figure \ref{fig: mix-highd-fig2}, we present the 11th to 13th coefficients (which are randomly generated) of the five groups. The groups G1, G3, and G5 have leading weights in determining the final \texttt{DRoL} model. Note that the 11th to 13th coefficients for group G5 are of the opposite sign to those for group G3 and are mostly of the same sign as those for group G1. The weights determined by \texttt{DRoL} lead to the cancellation of the heterogeneous effects with opposite signs.

\section{Proofs}
\label{sec: appendix-proof-main}
We begin with the proofs for our main results. In subsections \ref{sec: proof-thm-identification}, \ref{sec: proof-thm-magging-plug-in}, and \ref{sec: proof-thm-magging-correct}, we will subsequently present the identification theorem in Theorem \ref{thm: identification}, the approximation errors for plug-in DRoL and bias-corrected DRoL estimators in Theorem \ref{thm: Magging plug-in} and Theorem \ref{thm: Magging correct}, respectively. 
In subsections \ref{appendix: proof of squared} and \ref{appendix: proof of squared}, we present the proofs of the identification results for the squared error (Proposition \ref{prop: squared error}, Corollary \ref{coro: squared error}) and regret loss (Proposition \ref{prop: identification MRO}, Corollary \ref{coro: MRO issue}).
Then in the remaining subsections, we will present the proofs for Corollary \ref{cor: high-dim}, Theorem \ref{prop: reward-diff}, Propositions \ref{prop: shrink H}. The proofs of technical lemmas will be deferred to Section \ref{sec: proof-lemmas}.

\subsection{Proof of Theorem \ref{thm: identification}}
\label{sec: proof-thm-identification}
We fix $q \in \mathcal{H} \subseteq \Delta^L$ and consider the joint distribution $\TT = \left(\QQ_X, \TT_{Y|X} \right)$ with $\TT_{Y|X} = \sum_{l=1}^L q_l \cdot \PP{l}_{Y|X}$. 
Then the reward for the function $f(\cdot)$ with respect to $\TT$ is defined as:
\begin{align*}
    \RR_\TT(f) &:= \E_{(X_i, Y_i) \sim \TT} \left[Y_i^2 - (Y_i - f(X_i))^2 \right]
    = \E_{(X_i, Y_i) \sim \TT} \left[ 2Y_i f(X_i) - f(X_i)^2 \right].
\end{align*}
By definition of $\TT$, $\TT_{Y|X}$ is a weighted average of $\PP{l}_{Y|X}$,
\begin{align*}
    \RR_\TT(f) &= \E_{X_i \sim \QQ_X}\left[\sum_{l=1}^L q_l \E_{Y_i|X_i \sim \PP{l}_{Y|X}}\left[ 2Y_i f(X_i) - f(X_i)^2 \right] \right] \\
    &= \sum_{l=1}^L q_l \E_{X_i \sim \QQ_X} \E_{Y_i|X_i \sim \PP{l}_{Y|X}}\left[ 2Y_i f(X_i) - f(X_i)^2 \right].
\end{align*}
We can further derive the following equation of $\RR_\TT(f)$
\[
\RR_\TT(f) = \sum_{l=1}^L q_l \E_{X_i \sim \QQ_X} \left[ 2\f{l}(X_i)f(X_i) - f(X_i)^2\right],
\]
where the equality follows from the model for each source group in \eqref{eq: source-models}.
Then the objective in \eqref{eq: f_star H} is simplified as:
\begin{align}
    &\max_{f \in \FF} \min_{\TT \in \mathcal{C}(\QQ_X, \mathcal{H})} 
    \R_{\TT}(f) 
    = \max_{f \in \FF} \min_{q \in \HH}
    \left\{2\sum_{l=1}^L q_l \cdot \E_{\QQ_X} \left[\f{l}(X_i) f(X_i) \right] - \E_{\QQ_X}\left[ f(X_i)^2 \right]\right\}.
    \label{eq: thm1-obj1}
\end{align}
Suppose we can swap the order of max and min, which would be justified later, the new objective after swapping is,
\begin{equation}
    \min_{q\in \HH} \max_{f \in \FF} 
        \left\{2\sum_{l=1}^L q_l \cdot \E_{\QQ_X} \left[\f{l}(X_i) f(X_i) \right] - \E_{\QQ_X}\left[f(X_i)^2 \right]\right\}.
    \label{eq: thm1-obj2}
\end{equation}
Fix any $q\in \HH$, denote $f^q$ as the solution to the inner maximizer, i.e., 
\begin{equation}
    f^q(\cdot) = \argmax_{f \in \FF} \left\{2\sum_{l=1}^L q_l \cdot \E_{\QQ_X} \left[\f{l}(X_i) f(X_i) \right] - \E_{\QQ_X}\left[f(X_i)^2\right]\right\}.
    \label{eq: thm1-fq-objective}
\end{equation}
For any $h(\cdot) \in \FF$ and $t\in (0,1)$, since $\FF$ is a convex set, it holds that $f^q(\cdot) + t\left( h(\cdot) - f^q(\cdot) \right) \in \FF$. And by definition of $f^q(\cdot)$, we have
\begin{multline*}
    2\sum_{l=1}^L q_l \E_{\QQ_X}\left[ \f{l}(X_i)\left(f^q(X_i) + t (h(X_i) - f^q(X_i))\right) \right] - \E_{\QQ_X}\left[f^q(X_i) + t(h(X_i) - f^q(X_i))\right]^2 \\
    \leq 2\sum_{l=1}^L q_l \E_{\QQ_X}\left[\f{l}(X_i)f^q(X_i)\right] - \E_{\QQ_X}\left[f^q(X_i)\right]^2.
\end{multline*}
Hence for any $t \in (0,1)$, we have 
$$
2t\E_{\QQ_X}\left[\left(\sum_{l=1}^L q_l \f{l}(X_i) - f^q(X_i)\right)\left(h(X_i) - f^q(X_i)\right)\right] - t^2 \E_{\QQ_X}\left[h(X_i) - f^q(X_i)\right]^2 \leq 0
$$
By taking $t \rightarrow 0_+$, for any $h(\cdot) \in \FF$, we have:
$$
\E_{\QQ_X}\left[\left(\sum_{l=1}^L q_l \f{l}(X_i) - f^q(X_i)\right)\left(h(X_i) - f^q(X_i)\right)\right] \leq 0.
$$
We take $h(\cdot) = \sum_{l=1}^L q_l \f{l}(\cdot)$. Since $q \in \HH$ and $\f{l}(\cdot)\in \FF$ for each $l$, we have $h(\cdot) = \sum_{l=1}^L q_l \f{l}(\cdot) \in \FF$. Then we obtain
$$
\E_{\QQ_X}\left[\sum_{l=1}^L q_l \f{l}(X_i) - f^q(X_i)\right]^2 = 0,
$$ 
which implies
\begin{equation}
    f^q(\cdot)= \sum_{l=1}^L q_l \f{l}(\cdot) \quad \textrm{almost surely}.
    \label{eq: thm1-fq}
\end{equation}
Thus the optimizers for the objective \eqref{eq: thm1-obj2} is:
\begin{equation}
    q^* = \argmin_{q\in \mathcal{H}} \E_{\QQ_X} \left[\sum_{l=1}^L q_l \cdot \f{l}(X_i) \right]^2 \quad \textrm{and} \quad f^*(\cdot) = \sum_{l=1}^L q_l^* \cdot \f{l}(\cdot).
    \label{eq: thm1-fstar}
\end{equation}
Next we justify that the interchange of max and min by showing that $f^*(\cdot) = \sum_{l=1}^L q_l^* \cdot \f{l}(\cdot)$ is also the solution to the original objective \eqref{eq: thm1-obj1}, then the proof is completed. For any $t\in [0,1]$ and $\lambda \in \mathcal{H}$, we use the fact that $q^* + t(\lambda - q^*) \in \mathcal{H}$ and by the definition of $q^*$, we obtain that:
$$
\E_{\QQ_X} \left[ \sum_{l=1}^L (q^*_l + t (\lambda_l - q^*_l) ) \cdot \f{l}(X_i) \right]^2  \geq \E_{\QQ_X} \left[ \sum_{l=1}^L q^*_l \cdot \f{l} (X_i) \right]^2.
$$
The above inequality further implies
$$
2 t \E_{\QQ_X} \left[\left( \sum_{l=1}^L q_l^* \f{l}(X_i)\right) \left(\sum_{l=1}^L (\lambda_l - q^*_l) \f{l}(X_i) \right)\right] + t^2 \E_{\QQ_X} \left[\sum_{l=1}^L (\lambda_l - q^*_l) \f{l}(X_i)\right]^2 \geq 0.
$$
By taking $t\rightarrow 0_+$, we obtain
$$
\E_{\QQ_X} \left[\left( \sum_{l=1}^L q_l^* \f{l}(X_i)\right) \left(\sum_{l=1}^L (\lambda_l - q^*_l) \f{l}(X_i) \right)\right] \geq 0,
$$
which is equivalent as, for any $\lambda\in \HH$,
\begin{equation}
    2\E_{\QQ_X} \left[ \left(\sum_{l=1}^L q_l^* \f{l}(X_i)\right) \left(\sum_{l=1}^L \lambda_l \f{l}(X_i) \right) \right] - \E_{\QQ_X} \left[\sum_{l=1}^L q_l^* \cdot \f{l}(X_i) \right]^2 \geq \E_{\QQ_X} \left[\sum_{l=1}^L q_l^* \cdot \f{l}(X_i) \right]^2.
    \label{eq: thm1-interm1}
\end{equation}
For the original objective \eqref{eq: thm1-obj1}, the following inequality holds:
{\small
\begin{equation}
\begin{aligned}
    &\max_{f \in \FF} \min_{q \in \HH}
    \left\{2\sum_{l=1}^L q_l \cdot \E_{\QQ_X} \left[ \f{l}(X_i) f(X_i) \right] - \E_{\QQ_X}\left[ f(X_i)^2 \right]\right\} \\
    &\quad \geq \min_{q \in \mathcal{H}}
    \left\{2\sum_{l=1}^L q_l \cdot \E_{\QQ_X} \left[ \f{l}(X_i) f^*(X_i) \right] - \E_{\QQ_X}\left[ f^*(X_i)^2 \right]\right\}.
\end{aligned}
    \label{eq: thm1-interm2}
\end{equation}
}
Since $f^*(X_i)=\sum_{l=1}^L q_l^* \f{l}(X_i)$, we can rewrite the right hand side of \eqref{eq: thm1-interm2} as follows:
{\small
\begin{equation}
\begin{aligned}
    &2\sum_{l=1}^L q_l \cdot \E_{\QQ_X} \left[ \f{l}(X_i) f^*(X_i) \right] - \E_{\QQ_X}\left[ f^*(X_i)^2 \right]\\
    &\quad =2\E_{\QQ_X} \left[\left(\sum_{l=1}^L q_l \f{l}(X_i)\right) \left(\sum_{l=1}^L q_l^* \f{l}(X_i) \right) \right] - \E_{\QQ_X} \left[ \sum_{l=1}^L q_l^* \f{l}(X_i) \right]^2.
\end{aligned}
    \label{eq: thm1-interm3}
\end{equation}}
Combining \eqref{eq: thm1-interm2} and \eqref{eq: thm1-interm3}, we obtain:
{\small
\begin{equation*}
\begin{aligned}
    & \max_{f \in \FF} \min_{q \in \HH}
    \left\{2\sum_{l=1}^L q_l \cdot \E_{\QQ_X} \left[ \f{l}(X_i) f(X_i) \right] - \E_{\QQ_X}\left[ f(X_i)^2 \right]\right\} \\
    &\quad \geq \min_{q\in \mathcal{H}} \left\{2\E_{\QQ_X} \left[\left(\sum_{l=1}^L q_l \f{l}(X_i)\right) \left(\sum_{l=1}^L q_l^* \f{l}(X_i) \right) \right] - \E_{\QQ_X} \left[ \sum_{l=1}^L q_l^* \f{l}(X_i) \right]^2\right\}.
\end{aligned}
\end{equation*}}
As \eqref{eq: thm1-interm1} holds for any $\lambda \in \HH$, we apply it to the above inequality and establish that:
\begin{equation}
    \max_{f \in \FF} \min_{q \in \HH}
    \left\{2\sum_{l=1}^L q_l \cdot \E_{\QQ_X} \left[ \f{l}(X_i) f(X_i) \right] - \E_{\QQ_X}\left[ f(X_i)^2 \right]\right\} \geq \E_{\QQ_X} \left[\sum_{l=1}^L q_l^* \f{l}(X_i) \right]^2.
    \label{eq: thm1-direction-1}
\end{equation}
For the other direction, since $q^* \in \HH$, it holds that
{\small
\[
\begin{aligned}
    &\min_{q \in \mathcal{H}}
    \left\{2\sum_{l=1}^L q_l \cdot \E_{\QQ_X} \left[ \f{l}(X_i) f(X_i) \right] - \E_{\QQ_X}\left[ f(X_i)^2 \right]\right\}\\
    &\quad \leq 2\sum_{l=1}^L q_l^* \E_{\QQ_X}\left[\f{l}(X_i) f(X_i) \right] - \E_{\QQ_X} \left[f(X_i)^2\right] \leq \max_{f\in \mathcal{F}} \left\{2\sum_{l=1}^L q_l^* \E_{\QQ_X}\left[\f{l}(X_i) f(X_i) \right] - \E_{\QQ_X} \left[f(X_i)^2\right]\right\}.
\end{aligned}
\]}
Since the above inequality holds for any $f\in \FF$, we then have
{\small
\begin{equation}
    \begin{aligned}
    &\max_{f \in \mathcal{F}} \min_{q \in \mathcal{H}}
    \left\{2\sum_{l=1}^L q_l \cdot \E_{\QQ_X} \left[ \f{l}(X_i) f(X_i) \right] - \E_{\QQ_X}\left[ f(X_i)^2 \right]\right\} \\
    &\quad \leq \max_{f\in \mathcal{F}} \left\{2\sum_{l=1}^L q_l^* \E_{\QQ_X}\left[\f{l}(X_i) f(X_i) \right] - \E_{\QQ_X} \left[f(X_i)^2\right]\right\}.
\end{aligned}
\label{eq: thm1-interm4}
\end{equation}}
Recall the definition of $f^q(\cdot)$ in \eqref{eq: thm1-fq-objective}, the maximizer for the right hand side of the above inequality will be exactly $f^{*}$, that is defined in \eqref{eq: thm1-fq}. And by the result in \eqref{eq: thm1-fstar}, we obtain that
\begin{equation}
    \max_{f\in \mathcal{F}} \left\{2\sum_{l=1}^L q_l^* \E_{\QQ_X}\left[\f{l}(X_i) f(X_i) \right] - \E_{\QQ_X} \left[f(X_i)^2\right]\right\} = \E_{\QQ_X} \left[\sum_{l=1}^L q_l^* \f{l}(X_i) \right]^2.
    \label{eq: thm1-interm5}
\end{equation}
We combine \eqref{eq: thm1-interm4} and \eqref{eq: thm1-interm5} to establish
\begin{equation}
    \max_{f \in \mathcal{F}} \min_{q \in \mathcal{H}}
    \left\{2\sum_{l=1}^L q_l \cdot \E_{\QQ_X} \left[ \f{l}(X_i) f(X_i) \right] - \E_{\QQ_X}\left[ f(X_i)^2 \right]\right\} \leq \E_{\QQ_X} \left[\sum_{l=1}^L q_l^* \f{l}(X_i) \right]^2.
    \label{eq: thm1-direction-2}
\end{equation}
By matching the above two directions \eqref{eq: thm1-direction-1} and \eqref{eq: thm1-direction-2}, we have
\[
\max_{f \in \mathcal{F}} \min_{q \in \mathcal{H}}
    \left\{2\sum_{l=1}^L q_l \cdot \E_{\QQ_X} \left[ \f{l}(X_i) f(X_i) \right] - \E_{\QQ_X}\left[ f(X_i)^2 \right]\right\} = \E_{\QQ_X} \left[\sum_{l=1}^L q_l^* \f{l}(X_i) \right]^2.
\]
Since $f^*(\cdot) = \sum_{l=1}^L q_l^* \f{l}(\cdot) \in \FF$ and $q^*\in \HH$, we can draw the conclusion that $f^*(\cdot)$ is the optimizer to the original objective.

\subsection{Proof of Proposition \ref{prop: squared error} and Corollary \ref{coro: squared error} }
\label{appendix: proof of squared}
The squared loss for the function $f(\cdot)$ with respect to the distribution $\TT$ is defined as:
\begin{align*}
    \E_\TT[(Y_i - f(X_i))^2] &= \E_{X_i\sim \QQ_X}\E_{Y_i\sim \TT_{Y|X}}[(Y_i - f(X_i))^2] .
\end{align*}
As the distribution $\TT$ falls into the uncertainty class $\C(\QQ_X, \HH)$, the preceding can be expressed as
\[
\E_{X_i \sim \QQ_X}\left[\sum_{l=1}^L q_l \E_{Y_i\sim \PP{l}_{Y|X}}[(Y_i - f(X_i))^2]\right].
\]
Considering the data generation mechanism and the noise level assumption in Proposition \ref{prop: squared loss model}, it can be further expressed as follows:
\begin{align*}
    \E_\TT[(Y_i - f(X_i))^2]
    &= \E_{X_i \sim \QQ_X}\left[\sum_{l=1}^L q_l \cdot \E_{\eps{l}_i|X_i}\left[(\f{l}(X_i) - f(X_i))^2 + (\eps{l}_i)^2\right]\right] \\
    &= \E_{X_i \sim \QQ_X}\left[\sum_{l=1}^L q_l \cdot (\f{l}(X_i) - f(X_i))^2 + \sum_{l=1}^L q_l \cdot \sigma_l^2\right] \\
    &= \sum_{l=1}^L q_l \cdot \E_{\QQ_X} \left[\f{l}(X_i) - f(X_i)\right]^2 + \sum_{l=1}^L q_l \cdot \sigma_l^2.
\end{align*}
The above inequality leads to the following equivalence in terms of the optimization problem,
\[
\min_{f\in \FF}\max_{\TT \in \mathcal{C}(\QQ_X, \HH)} \E_\TT[(Y_i - f(X_i))^2] = 
\min_{f\in \FF}\max_{q\in \HH} \left\{\sum_{l=1}^L q_l \cdot \E_{\QQ_X} \left[\f{l}(X_i) - f(X_i)\right]^2 + \sum_{l=1}^L q_l \cdot \sigma_l^2\right\}.
\]
By minimax theorem \cite{sion1958general,komiya1988elementary} with the convex concave objective, we can swap the order of min and max, the objective after swapping is:
\begin{equation}
    \max_{q\in \HH} \min_{f\in \FF} \left\{\sum_{l=1}^L q_l \cdot \E_{\QQ_X} \left[\f{l}(X_i) - f(X_i)\right]^2 + \sum_{l=1}^L q_l \cdot \sigma_l^2\right\}.
    \label{proof-prop-sqloss-obj-swap}
\end{equation}
For any fixed $q\in \HH$, we define $f^q$ as the solution for the inner minimization problem in \eqref{proof-prop-sqloss-obj-swap}.
\begin{align*}
    f^q &:= \argmin_{f\in \FF}\left\{\sum_{l=1}^L q_l \E_{\QQ_X} \left[\f{l}(X_i) - f(X_i)\right]^2 + \sum_{l=1}^L q_l \sigma_l^2\right\} \\
    &= \argmin_{f\in \FF}\left\{\sum_{l=1}^L q_l \E_{\QQ_X} \left[\f{l}(X_i) - f(X_i)\right]^2\right\},
\end{align*}
where we remove the term $\sum_{l=1}^L q_l \sigma_l^2$ as it is irrelevant to the $f$. Then we can further express $f^q$ as follows:
\begin{equation}
    \begin{aligned}
        f^q &= \argmin_{f\in \FF}\left\{\sum_{l=1}^L q_l \E_{\QQ_X}\left[(\f{l}(X_i))^2 - 2\f{l}(X_i)f(X_i) + (f(X_i))^2\right]\right\} \\
    &= \argmin_{f\in \FF}\left\{-2\sum_{l=1}^L q_l \cdot \E_{\QQ_X}\left[\f{l}(X_i)f(X_i)\right] + \E_{\QQ_X}(f(X_i))^2\right\},
    \end{aligned}
\end{equation}
where we remove term $\sum_{l=1}^L q_l \E_{\QQ_X}[\f{l}(X_i)^2]$ as it is irrelevant to the $f$ again. With this new expression for $f^q$, we note that $f^q$ is defined exactly the same way as \eqref{eq: thm1-fq-objective} in Theorem \ref{thm: identification}. Therefore by \eqref{eq: thm1-fq}, we can obtain that
\[
f^q(\cdot) = \sum_{l=1}^L q_l \f{l}(\cdot) \quad \textrm{almost surely}.
\]
Hence the objective after swapping in \eqref{proof-prop-sqloss-obj-swap} can be expressed as
\[
\max_{q\in \HH} \left\{\sum_{l=1}^L q_l \E_{\QQ_X}\left[\f{l}(X_i) - f^q(X_i)\right]^2 + \sum_{l=1}^L q_l \sigma_l^2\right\}
\]
Next we define $q^*$ as the solution for the above objective as follows,
\begin{gather}
    \begin{aligned}
    q^* &= \argmax_{q\in \HH} \left\{\sum_{l=1}^L q_l \E_{\QQ_X}\left[\f{l}(X_i) - f^q(X_i)\right]^2 + \sum_{l=1}^L q_l \sigma_l^2\right\} \\
    &= \argmax_{q\in \HH} \left\{\sum_{l=1}^L q_l \E_{\QQ_X}\left[\f{l}(X_i)^2\right] - \E_{\QQ_X}\left(\sum_{l=1}^L q_l \f{l}(X_i)\right)^2 + \sum_{l=1}^L q_l \sigma_l^2\right\} \\
    &= \argmin_{q\in \HH} \left\{q^\intercal \Gamma q - q^\intercal (\gamma + \bm{\sigma}^2)\right\},
\end{aligned}
\label{eq: prop-squaredloss-1}
\end{gather}
where $\Gamma_{k,l} = \E_{\QQ_X}\E[\f{k}(X)\f{l}(X)]$ for $k,l\in [L]$, $\gamma_{l}=\Gamma_{l,l}$ for $l\in [L]$, and $\bm{\sigma}^2 = (\sigma_1^2,...,\sigma_L^2).$

Now, we start the proof of Corollary \ref{coro: squared error}.
Considering $\HH = \Delta^L$ and $L=2$,
with Lagrangian multiplier $\lambda$,  we define
\[
\mathcal{L}(q_1, q_2, \lambda) = q_1\E(\f{1})^2 + q_2\E(\f{2})^2 - \E(q_1 \f{1} + q_2 \f{2})^2 + q_1 \sigma_1^2 + q_2 \sigma_2^2 + \lambda (1 - q_1 - q_2),
\]
where we use the simplified notations for convenience. Then the following three equalities hold simultaneously,
\[
\frac{\partial \mathcal{L}}{\partial q_1} = \E(\f{1})^2 - 2q_1\E(\f{1})^2  - 2q_2 \E(\f{1}\f{2}) + \sigma_1^2 - \lambda = 0,
\]
\[
\frac{\partial \mathcal{L}}{\partial q_2} = \E(\f{2})^2 - 2q_2\E(\f{2})^2  - 2q_1 \E(\f{1}\f{2}) + \sigma_2^2 - \lambda = 0,
\]
\[
\frac{\partial \mathcal{L}}{\partial \lambda} = 1 - q_1 - q_2 = 0.
\]
After simple organization of the preceding equalities, we establish that
{\small
\[
\E(\f{1})^2 - 2q_1\E(\f{1})^2  - 2(1-q_1) \E(\f{1}\f{2}) + \sigma_1^2 = \E(\f{2})^2 - 2(1-q_1)\E(\f{2})^2  - 2q_1 \E(\f{1}\f{2}) + \sigma_2^2,
\]}
which implies that
\[
q_1\E \left[2(\f{1})^2 + 2(\f{2})^2 - 4\f{1}\f{2}\right] = \E \left[(\f{1})^2 + (\f{2})^2 - 2(\f{1}\f{2}) \right] + \sigma_1^2 - \sigma_2^2.
\]
Therefore, we obtain that
\[
2q_1 \E(\f{1}- \f{2})^2 = \E(\f{1}- \f{2})^2 + \sigma_1^2 - \sigma_2^2
\]
By computation, with boundary constraints that $0\leq q_1\leq 1$, we can obtain that
\[
q_1^* = 0 \vee \left(\frac{1}{2}+\frac{\sigma_1^2 - \sigma_2^2}{\left\|\f{1} - \f{2})^2\right\|_{\ell_2(\QQ)}^2}\right)\wedge 1; \quad 
q_2^* = 1-q_1^*
\]

\subsection{Proof of Proposition \ref{prop: identification MRO} and Corollary \ref{coro: MRO issue}} 
\label{appendix: proof of mro}
Given a weight $q\in \Delta^L$, we consider the joint distribution $\TT = (\QQ_X, \TT_{Y|X})$ with $\TT_{Y|X} = \sum_{l=1}^L q_l \cdot \PP{l}_{Y|X}$. We denote $f^\TT(x) = \E_{\TT}[Y|X = x]$, and $\f{l}(x) = \E_{\PP{l}}[Y|X=x]$, then we obtain that
\[
f^\TT(x) = \sum_{l=1}^L q_l \cdot \f{l}(x).
\]
Since the function class $\FF$ is convex set and contains all $\{\f{l}\}_{l\in [L]}$, it is established that $f^\TT \in \FF.$ Then
\[
\inf_{f'\in \FF}\E_\TT[(Y - f'(X))^2] = \E_\TT[(Y - f^\TT(X))^2].
\]
Then we study the regret:
\begin{equation}
    \begin{aligned}
        {\rm Regret}_\TT(f) &= \E_\TT[(Y - f(X))^2] - \inf_{f'\in \FF}\E_\TT[(Y - f'(X))^2]\\
        &= \E_\TT[(Y - f(X))^2] - \E_\TT[(Y - f^\TT(X))^2] \\
        &= \E_\TT[(Y-f^\TT(X) + f^\TT(X) - f(X))^2] - \E_\TT[(Y - f^\TT(X))^2]\\
        &= \E_{\QQ_X}[(f^\TT(X) - f(X))^2].
    \end{aligned}
    \label{proof MRO regret}
\end{equation}
Then the optimization can be expressed equivalently as follows:
\begin{equation}
\begin{aligned}
    \min_{f\in \FF}\max_{\TT\in \C(\QQ_X, \HH)} {\rm Regret}_\TT(f) &= \min_{f\in \FF}\max_{\TT\in \C(\QQ_X, \HH)} \E_{\QQ_X}[(f^\TT(X) - f(X))^2] \\
    &= \min_{f\in \FF}\max_{q\in \HH} \E_{\QQ_X}\left(\sum_{l=1}^L q_l \f{l}(X) - f(X)\right)^2.
\end{aligned}
\label{proof MRO equiv}
\end{equation}
Equivalently, we can write the right-hand side of the above equation as follows:
\[
\min_{f\in \FF}\left\{r\geq 0:~~ \max_{q\in \HH} \E_{\QQ_X}\left(\sum_{l=1}^L q_l \f{l}(X) - f(X)\right)^2 \leq r\right\}.
\]

In the special case when $\HH = \Delta^L$, we have
\[
\min_{f\in \FF}\max_{\TT\in \C(\QQ_X)} {\rm Regret}_\TT(f) = \min_{f\in \FF}\max_{q\in \Delta^L} \E_{\QQ_X}\left(\sum_{l=1}^L q_l \f{l}(X) - f(X)\right)^2.
\]
We denote the objective as follows:
\[
h(q) = \E_{\QQ_X}\left(\sum_{l=1}^L q_l \f{l}(X) - f(X)\right)^2.
\]
It is evident that $h(q)$ is convex w.r.t $q$, since its Hessian matrix is positive semi-definite as follows:
\[
\mathbf{H}_q (h) = 2 \Gamma \succeq 0, 
\]
where $\Gamma_{k,l} = \E_{\QQ_X}[\f{k}(X) \f{l}(X)]$, for $k,l\in [L].$ 
Notice that $q = \sum_{l=1}^L q_l \mathbf{e}_l$, where $\mathbf{e}_l\in \R^L$ is the $l$-th basis vector, with $l$-th entry being one and also other entries are zeros. 
Therefore by the convexity of $h$, it holds that
\[
h(q) = h(\sum_{l=1}^L q_l \mathbf{e}_l) \leq \sum_{l=1}^L q_l h(\mathbf{e}_l) \leq \max_{1\leq l\leq L} h(\mathbf{e}_l).
\]
The above inequality takes equality when $q = \mathbf{e}_{l^*}$ with $l^* \in \argmax_{1\leq l\leq L} h(\mathbf{e}_l).$. Therefore, it follows that
\[
\max_{q\in \Delta^L} h(q) = \max_{1\leq l\leq L} h(\mathbf{e}_l) = \max_{1\leq l\leq L} \E_{\QQ_X}\left(\f{l}(X)-f(X)\right)^2.
\]
Now we put the above result back to \eqref{proof MRO equiv}, then the MRO optimization becomes
\begin{equation}
    \begin{aligned}
    \min_{f\in \FF}\max_{\TT\in \C(\QQ_X)} {\rm Regret}_\TT(f) &= \min_{f\in \FF}\max_{1\leq l\leq L}\E_{\QQ_X}\left(\f{l}(X)-f(X)\right)^2\\
    &= \min_{f\in \FF}\max_{w\in \Delta^L}\sum_{l=1}^L w_l \cdot \E_{\QQ_X}\left(\f{l}(X)-f(X)\right)^2,
\end{aligned}
\label{proof MRO equiv 2}
\end{equation}
where the last equality holds because the maximum is always taken at the vertices of the simplex.

We define
\begin{equation}
    w^{\rm reg} = \argmin_{w\in  \Delta^L}\left\{w^\intercal \Gamma w - \gamma^\intercal w\right\},\quad f^{\rm reg} = \sum_{l=1}^L w^{\rm reg}_l \cdot \f{l},
    \label{w MRO def}
\end{equation}
where $\Gamma_{k,l} = \E_{\QQ_X}[\f{k}(X) \f{l}(X)]$, for $k,l\in [L]$, and $\gamma_{l} = \Gamma_{l,l}$ for $l\in [L].$ In the following discussion, we will show that $f^{\rm reg}$ is the optimizer of \eqref{proof MRO equiv 2}.

Firstly,
\[
\begin{aligned}
    &\min_{f\in \FF}\max_{w\in \Delta^L}\sum_{l=1}^L w_l \cdot \E_{\QQ_X}\left(\f{l}(X)-f(X)\right)^2 \\
    &\geq \min_{f\in \FF} \sum_{l=1}^L w^{\rm reg}_l \cdot \E_{\QQ_X}\left(\f{l}(X)-f(X)\right)^2 \\
    &= \sum_{l=1}^L w^{\rm reg}_l \E_{\QQ_X}[\f{l}(X)]^2 + \min_{f\in \FF}\left\{\E_{\QQ_X}[f(X)]^2 - 2\sum_{l=1}^L w^{\rm reg}_l \E_{\QQ_X}[\f{l}(X)f(X)]\right\}.
\end{aligned}
\]
Now we apply the result in \eqref{eq: thm1-fq-objective} and \eqref{eq: thm1-fq} and it is established that
\[
f^{\rm reg} = \argmin_{f\in \FF}\left\{\E_{\QQ_X}[f(X)]^2 - 2\sum_{l=1}^L w^{\rm reg}_l \E_{\QQ_X}[\f{l}(X)f(X)]\right\},
\]
and
\[
\min_{f\in \FF}\left\{\E_{\QQ_X}[f(X)]^2 - 2\sum_{l=1}^L w^{\rm reg}_l \E_{\QQ_X}[\f{l}(X)f(X)]\right\} = - \E_{\QQ_X}[f^{\rm reg}(X)]^2.
\]
therefore,
\begin{equation}
    \min_{f\in \FF}\max_{w\in \Delta^L}\sum_{l=1}^L w_l \cdot \E_{\QQ_X}\left(\f{l}(X)-f(X)\right)^2 \geq \sum_{l=1}^L w_l^{\rm reg}\E_{\QQ_X}[\f{l}(X)]^2 - \E_{\QQ_X}[f^{\rm reg}(X)]^2.
    \label{MRO ineq - 1}
\end{equation}

On the other hand, we have
\[
\begin{aligned}
    &\min_{f\in \FF}\max_{w\in \Delta^L}\sum_{l=1}^L w_l \cdot \E_{\QQ_X}\left(\f{l}(X)-f(X)\right)^2 \\
    &\leq \max_{w\in \Delta^L}\sum_{l=1}^L w_l \cdot \E_{\QQ_X}\left(\f{l}(X)- \sum_{k=1}^L w_k \f{k}(X)\right)^2 \\
    &= \max_{w\in \Delta^L}\sum_{l=1}^L w_l \cdot\left\{\E_{\QQ_X}[\f{l}(X)]^2 - 2\E_{\QQ_X}\left[\f{l}(X) \left(\sum_{k=1}^L w_k \f{k}(X)\right)\right] + \E_{\QQ_X}\left[\sum_{k=1}^L w_k \f{k}(X)\right]^2\right\} \\
    &= \max_{w\in \Delta^L} \sum_{l=1}^L w_l \E_{\QQ_X}\left[\f{l}(X)\right]^2 - \E_{\QQ_X}\left[\sum_{k=1}^L w_k \f{l}(X)\right]^2 \\
    &= \max_{w\in \Delta^L} \left\{\gamma^\intercal w - w^\intercal \Gamma w\right\},
\end{aligned}
\]
where $\gamma, \Gamma$ are defined in \eqref{w MRO def}.
Recall the definition of $w^{\rm reg}$ in \eqref{w MRO def}, it follows that
\[
\begin{aligned}
    \max_{w\in \Delta^L} \left\{\gamma^\intercal w - w^\intercal \Gamma w\right\} &= \gamma^\intercal w^{\rm reg} - (w^{\rm reg})^\intercal \Gamma w^{\rm reg} \\
&= \sum_{l=1}^L w_l^{\rm reg}\E_{\QQ_X}[\f{l}(X)]^2 - \E_{\QQ_X}[f^{\rm reg}(X)]^2.
\end{aligned}
\]
Combining the above two results, we obtain that
\begin{equation}
    \min_{f\in \FF}\max_{w\in \Delta^L}\sum_{l=1}^L w_l \cdot \E_{\QQ_X}\left(\f{l}(X)-f(X)\right)^2 \leq \sum_{l=1}^L w_l^{\rm reg}\E_{\QQ_X}[\f{l}(X)]^2 - \E_{\QQ_X}[f^{\rm reg}(X)]^2.
    \label{MRO ineq - 2}
\end{equation}

Therefore, it follows from \eqref{MRO ineq - 1} and \eqref{MRO ineq - 2} that
\[
\begin{aligned}
    \min_{f\in \FF}\max_{w\in \Delta^L}\sum_{l=1}^L w_l \cdot \E_{\QQ_X}\left(\f{l}(X)-f(X)\right)^2 
    &= \sum_{l=1}^L w_l^{\rm reg}\E_{\QQ_X}[\f{l}(X)]^2 - \E_{\QQ_X}[f^{\rm reg}(X)]^2\\
    &= \sum_{l=1}^L w^{\rm reg}_l \cdot \E_{\QQ_X}\left(\f{l}(X)-f^{\rm reg}(X)\right)^2.
\end{aligned}
\]
Therefore, together with \eqref{proof MRO equiv 2} that
\begin{equation}
\begin{aligned}
    f^{\rm reg} &= \argmin_{f\in \FF}\max_{w\in \Delta^L}\sum_{l=1}^L w_l \cdot \E_{\QQ_X}\left(\f{l}(X)-f(X)\right)^2 \\
    &= \argmin_{f\in \FF}\max_{\TT\in \C(\QQ_X)} {\rm Regret}_\TT(f).
\end{aligned}
\label{eq: MRO identification}
\end{equation}

\subsection{Proof of Theorem \ref{thm: Magging plug-in}}
\label{sec: proof-thm-magging-plug-in}
With $\tfH$ and $\fH$ defined in \eqref{eq: Magging plug-in} and \eqref{eq: identification}, respectively, it holds that
\[
\left(\tfH(x) - \fH(x)\right)^2 = \left(\sum_{l=1}^L \widetilde{q}_l \hf{l}(x) - \sum_{l=1}^L q_l^* \f{l}(x)\right)^2 = \left(\sum_{l=1}^L \widetilde{q}_l \Del{l}(X_i) + \sum_{l=1}^L (\widetilde{q}_l - q_l^*) \f{l}(x)\right)^2.
\]
where $\Del{l}(\cdot) := \hf{l}(\cdot) - \f{l}(\cdot)$. It follows that
\[
\left(\sum_{l=1}^L \widetilde{q}_l \Del{l}(X_i) + \sum_{l=1}^L (\widetilde{q}_l - q_l^*) \f{l}(x)\right)^2 \leq 2\left(\left(\sum_{l=1}^L \widetilde{q}_l \Del{l}(x)\right)^2 + \left(\sum_{l=1}^L (\widetilde{q}_l - q_l^*) \f{l}(x)\right)^2 \right).
\]
We then apply the Cauchy-Schwarz inequality to both terms on the right-hand side of the above inequality and establish that
\[
\left(\sum_{l=1}^L \widetilde{q}_l \Del{l}(x)\right)^2\leq \sum_{l=1}^L \widetilde{q}_l^2 \sum_{l=1}^L \Del{l}(x)^2 \leq \sum_{l=1}^L \Del{l}(x)^2,
\]
where the last inequality is due to the fact that $\widetilde{q}\in \Delta^L$, and
\[
\left(\sum_{l=1}^L (\widetilde{q}_l - q_l^*) \f{l}(x)\right)^2 
\leq \|\widetilde{q} - q^*\|_2^2 \sum_{l=1}^L \f{l}(x)^2.
\]
Therefore, it follows that
\[
\left(\tfH(x) - \fH(x)\right)^2 \leq 2\left(\sum_{l=1}^L \Del{l}(x)^2 + \|\widetilde{q} - q^*\|_2^2 \sum_{l=1}^L \f{l}(x)^2 \right).
\]
Then we can quantify the error of $\tfH$:
\begin{equation}
\begin{aligned}
\left\|\tfH - \fH\right\|_{\ell_2(\QQ)} 
&= \sqrt{\E_\QQ \left(\tfH(X) - \fH(X)\right)^2} \leq \sqrt{2\left(\sum_{l=1}^L \E_\QQ \left[\Del{l}(X)^2\right] + \|\widetilde{q} - q^*\|_2^2 \sum_{l=1}^L \E_\QQ \left[\f{l}(x)^2\right]\right)}\\
&\leq \sqrt{2L \left(\delta_n^2 + \|\widetilde{q} - q^*\|_2^2M^2\right)} \leq \sqrt{2L}\left(\delta_n + \|\widetilde{q} - q^*\|_2 M\right)
\end{aligned}
\label{eq: approx1-f_diff}
\end{equation}
To upper bound the term $\|\widetilde{q} - q^*\|_2$, we will use the following Lemma, which essentially follows the Lemma 6 in \cite{guo2020inference}. We delay its proof to Appendix \ref{sec: proof-lemma-approx}.
\begin{Lemma}
For the convex subset $\HH$ of $\Delta^L$, its diameter $\rho_{\HH}$ is defined as $\rho_\HH = \max_{q,q'\in \HH}\|q - q'\|_2$. For the semi-positive definite matrix $\Gamma$, $\widehat{\Gamma}$, we define
$q^* = \argmin_{q\in \HH} q^\intercal\Gamma q$ and $\widehat{q} = \argmin_{q\in \HH }q^\intercal \hGamma q $.
If $\lambda_{\rm min}(\Gamma) > 0$, then
\[
\|\widehat{q} - q^*\|_2 \leq \frac{L\|\widehat{\Gamma} - \Gamma\|_\infty}{\lambda_{\rm min}(\Gamma)} \wedge \rho_\HH.
\]
\label{lemma: approx}
\end{Lemma}
\noindent We apply Lemma \ref{lemma: approx} with $\hGamma=\tGamma$, and obtain that
\begin{equation}
    \|\widetilde{q} - q^*\|_2 \leq \frac{L\|\tGamma - \Gamma\|_\infty}{\lambda_{\rm min}(\Gamma)} \wedge \rho_\HH.
    \label{eq: approx1-qdiff}
\end{equation}
Combining \eqref{eq: approx1-f_diff} and \eqref{eq: approx1-qdiff} and \eqref{eq: approx1-Gamma_diff}, we establish that,
\begin{equation}
    \left\|\tfH - \fH\right\|_{\ell_2(\QQ)} \leq \sqrt{2L}\left[\delta_n + \left(\frac{L\|\tGamma - \Gamma\|_\infty}{\lambda_{\rm min}(\Gamma)} \wedge \rho_\HH\right)M\right].
    \label{eq: approx1-basic}
\end{equation}
In the following discussions, we will continue the analysis of the term $\|\tGamma - \Gamma\|_\infty$.

For fixed $k,l\in [L]$, we decompose the error of $\widetilde{\Gamma}$ as follows:
{\small
\begin{equation}
    \begin{aligned}
    &\left({\tGamma}- \Gamma\right)_{k,l} = \frac{1}{N}\sum_{i=1}^{N}\hf{k}(X_i)\hf{l}(X_i) - \E_\QQ \f{k}(X)\f{l}(X) \\
    &= \underbrace{\frac{1}{N}\sum_{i=1}^{N} \hf{k}(X_i)\hf{l}(X_i) - \frac{1}{N}\sum_{i=1}^{N} \f{k}(X_i)\f{l}(X_i)}_{(\RN{1})_{k,l}} \\
    &\quad \quad+ \underbrace{\frac{1}{N}\sum_{i=1}^{N} \f{k}(X_i)\f{l}(X_i) - \E_\QQ \f{k}(X)\f{l}(X)}_{(\RN{2})_{k,l}}.
\end{aligned}
\label{eq: proof-thm-plug-error-decompose}
\end{equation}}
We will upper bound the error parts $(\RN{1})_{k,l}$ and $(\RN{2})_{k,l}$ in the followings. 

\subsubsection{\underline{\textbf{Part $(\RN{1})_{k,l}$}}}
We define $\Del{l}(\cdot) = \hf{l}(\cdot) - \f{l}(\cdot)$, then $(\RN{1})_{k,l}$ can be re-expressed as follows:
{\small
\begin{align*}
    \left|(\RN{1})_{k,l}\right| &= \left|\frac{1}{N}\sum_{i=1}^{N} \Del{k}(X_i)\f{l}(X_i) + \frac{1}{N}\sum_{i=1}^{N} \Del{l}(X_i)\f{k}(X_i) + \frac{1}{N}\sum_{i=1}^{N} \Del{k}(X_i)\Del{l}(X_i)\right| \\
    &\leq \left|\frac{1}{N}\sum_{i=1}^{N} \Del{k}(X_i)\f{l}(X_i)\right| + \left|\frac{1}{N}\sum_{i=1}^{N} \Del{l}(X_i)\f{k}(X_i)\right| +\left|\frac{1}{N}\sum_{i=1}^{N} \Del{k}(X_i)\Del{l}(X_i)\right| \\ 
    &\leq \sqrt{\frac{1}{N}\sum_{i=1}^{N}(\Del{k}(X_i))^2}\sqrt{\frac{1}{N}\sum_{i=1}^{N}(\f{l}(X_i))^2} + \sqrt{\frac{1}{N}\sum_{i=1}^{N}(\Del{l}(X_i))^2}\sqrt{\frac{1}{N}\sum_{i=1}^{N}(\f{k}(X_i))^2} \\
    &\quad \quad +\sqrt{\frac{1}{N}\sum_{i=1}^{N}(\Del{k}(X_i))^2}\sqrt{\frac{1}{N}\sum_{i=1}^{N}(\Del{l}(X_i))^2}.
\end{align*}}
where the first inequality is due to the triangle inequality, and the second inequality is derived from the Cauchy-Schwarz inequality. Next, we apply Markov's inequality to further upper bound the terms in the above inequality. For any $t>1$, the following inequalities hold separately:
\begin{equation}
    \prob\left\{\frac{1}{N}\sum_{i=1}^{N} \Del{l}(X_i)^2 \geq 2 t L \delta_n^2\right\} \leq \frac{\E_\QQ \left[\Del{l}(X)^2\right]}{2 t L \delta_n^2} \leq \frac{1}{2tL}.
    \label{eq: approx1-markov1}
\end{equation}
\begin{equation}
    \prob\left\{\frac{1}{N}\sum_{i=1}^{N} \f{l}(X_i)^2 \geq  2 t L M^2 \right\} \leq \frac{\E_\QQ \left[\f{l}(X)^2\right]}{2tLM^2}\leq \frac{1}{2tL}.
    \label{eq: approx1-markov2}
\end{equation}
Define the event $\Omega_1$ as
\[
\Omega_1 = \left\{\frac{1}{N}\sum_{i=1}^{N} \Del{l}(X_i)^2 \geq 2 t L \delta_n^2,\; \frac{1}{N}\sum_{i=1}^{N} \f{l}(X_i)^2 \geq 2 t L M^2 \quad \textrm{for some $l=1,..., L$} \right\}.
\]
Then by union bound, we obtain that $\prob(\Omega_1) \leq 1/t$. It follows that, with probability at least $1 - \frac{1}{t}$,
\begin{equation}
    \max_{k,l}\left|(\RN{1})_{k,l}\right| \leq 2tL( 2 M \delta_n + \delta_n^2).
    \label{eq: approx1-term1}
\end{equation}
\subsubsection{\underline{\textbf{Part $(\RN{2})_{k,l}$}}}
We apply Chebyshev's inequality to upper bound the term.
{\small
\[
\begin{aligned}
    \prob\left\{\left|\frac{1}{N}\sum_{i=1}^{N} \f{k}(X_i)\f{l}(X_i) - \E_\QQ\f{k}(X)\f{l}(X)\right| \geq a \right\} &\leq \frac{\textrm{Var}_\QQ(\f{k}(X)\f{l}(X))}{N a^2} \\
    &\leq \frac{\E_\QQ \left[\left(\f{k}(X)\f{l}(X)\right)^2\right]}{N a^2},
\end{aligned}
\]}
where $\textrm{Var}_\QQ(\f{k}(X)\f{l}(X))$ is the variance for $\f{k}(X)\f{l}(X)$ evaluated on the target covariates population. With the Cauchy-Schwarz inequality, we have
\[
\E_\QQ \left[\left(\f{k}(X)\f{l}(X)\right)^2\right] \leq \sqrt{\E_\QQ\left[\f{k}(X)^4\right]} \sqrt{\E_\QQ\left[\f{l}(X)^4\right]} \leq M^4.
\]
Therefore, it leads to
\[
\prob\left\{\left|\frac{1}{N}\sum_{i=1}^{N} \f{k}(X_i)\f{l}(X_i) - \E_\QQ\f{k}(X)\f{l}(X)\right| \geq a \right\} \leq \frac{M^4}{N a^2}.
\]
For $t>1$, we take $a = t L M^2/ \sqrt{N}$, then the above inequality becomes:
\begin{equation}
    \prob\left\{\left|\frac{1}{N}\sum_{i=1}^{N} \f{k}(X_i)\f{l}(X_i) - \E_\QQ\f{k}(X)\f{l}(X)\right| \geq t L \frac{M^2}{\sqrt{N}}\right\} \leq \frac{1}{t^2 L^2}.
    \label{eq: approx1-chebyshev1}
\end{equation}
Define the event $\Omega_2$ as
\[
\Omega_2 = \left\{ \left|\frac{1}{N}\sum_{i=1}^{N} \f{k}(X_i)\f{l}(X_i) - \E_\QQ\f{k}(X)\f{l}(X)\right| \geq t L \frac{M^2}{\sqrt{N}} \quad \textrm{for some $k,l\in [L]$}\right\}.
\]
it holds that $\prob(\Omega_2)\leq \frac{1}{t^2}$ by union bound. It follows that with probability at least $1 - \frac{1}{t^2}$,
\begin{equation}
    \max_{k,l}\left|(\RN{2})_{k,l}\right| \leq t L \frac{M^2}{\sqrt{N}}.
    \label{eq: approx1-term2}
\end{equation}

\subsubsection{\underline{\textbf{Proof of \eqref{eq: error plug-in}}}}
Combining the inequalities in \eqref{eq: approx1-term1} and \eqref{eq: approx1-term2}, we establish that, with probability at least $1 - \frac{1}{t} - \frac{1}{t^2}$
\begin{equation}
    \left\|\widetilde{\Gamma} - \Gamma\right\|_\infty \leq \max_{k,l}\left|(\RN{1})_{k,l}\right| + \max_{k,l}\left|(\RN{2})_{k,l}\right| \leq t L \cdot \left(2 \delta_n^2  + \frac{M^2}{\sqrt{N}} + 4M \delta_n\right).
    \label{eq: approx1-Gamma_diff}
\end{equation}

Combining \eqref{eq: approx1-basic} and \eqref{eq: approx1-Gamma_diff}, we establish that with probability at least $1-\frac{1}{t}-\frac{1}{t^2}$:
{\small
\begin{multline}
    \left\|\tfH - \fH\right\|_{\ell_2(\QQ)} \leq \sqrt{2L}\left[\delta_n + \left(\frac{L\|\tGamma - \Gamma\|_\infty}{\lambda_{\rm min}(\Gamma)} \wedge \rho_\HH\right)M\right] \\
    \lesssim \delta_n + t\cdot \max\left\{\frac{M}{\lambda_{\rm min}(\Gamma)}\left(\delta_n^2+ \frac{M^2}{\sqrt{N}} + M\delta_n\right), M\rho_\HH\right\}.
    \label{proof-drl-plug-in-f-diff}
\end{multline}}


\subsection{Proof of Theorem \ref{thm: Magging correct}}
\label{sec: proof-thm-magging-correct}
We present the relaxed version of Assumption \ref{ass: correct} without requiring the boundeness of $\w{l}(x)$ for any $x$.
\begin{Assumption}
For any $l\in [L]$ and $x\in \R^p$, $\w{l}(x)>0$. With probability larger than $1-\tau_n$ with $\tau_n\rightarrow 0,$ there exists a positive sequence  $\eta_{\omega}$ such that for any $l\in [L]$, 
{\small
\[
\max\left\{\left\|\frac{\hw{l}}{\w{l}} - 1\right\|_{\ell_2(\QQ)}, \left\|\frac{\hw{l}}{\w{l}} - 1\right\|_{\ell_4(\QQ)},\left\|\hw{l} - \w{l}\right\|_{\ell_4(\QQ)}\right\}\leq \eta_{\omega}
\]
}
\label{ass: correct-relax}
\end{Assumption}
We follow the same argument as \eqref{eq: approx1-basic}, which is derived for the plug-in DRoL estimator, but replace $\tGamma$ with $\hGamma$. Here, we present the analysis for the term $\left\|\hGamma - \Gamma\right\|_\infty$.
In this proof, we use subscript $i$ as the index for the observations $\{\XQ_i\}_{i\in [N]}$ in the target data, and use subscript $j$ for the observations $\{(\X{l}_j, \Y{l}_j)\}_{j\in [n_l], l\in [L]}$ in the source data.

For any fixed $k,l\in [L]$, we combine \eqref{eq: error decomposition} and \eqref{eq: Gamma-debias-CS} and decompose $(\hGamma-\Gamma)_{k,l}$ as follows:
{\small
\begin{align*}
    (\widehat{\Gamma}_\mathcal{A} - &\Gamma)_{k,l}
    = \underbrace{\frac{1}{N}\sum_{i=1}^{N}\f{k} (X_i)\f{l}(X_i)-\E_\QQ \f{k}(X)\f{l}(X)}_{(\RN{1})_{k,l}} \\
    &+ \underbrace{\frac{1}{N}\sum_{i=1}^{N}\hf{k}(X_i)\left(\hf{l}(X_i) - \f{l}(X_i)\right) - \frac{1}{n_l}\sum_{j=1}^{n_l} \hw{l}(X_j)\hf{k}(X_j)\left(\hf{l}(X_j) - Y_j\right)}_{(\RN{2})_{k,l}}  \\
    & + \underbrace{\frac{1}{N}\sum_{i=1}^{N}\hf{l}(X_i)\left(\hf{k}(X_i) - \f{k}(X_i)\right) - \frac{1}{n_k}\sum_{j=1}^{n_k} \hw{k}(X_j)\hf{l}(X_j)\left(\hf{k}(X_j) - Y_j\right)}_{(\RN{3})_{k,l}} \\
    &- \underbrace{\frac{1}{N}\sum_{i=1}^{N} \left(\hf{k}(X_i) - \f{k}(X_i)\right)\left(\hf{l}(X_i) - \f{l}(X_i)\right)}_{(\RN{4})_{k,l}}.
\end{align*}}
Subsequently, we will study the terms $(\RN{1})_{k,l}$, $(\RN{4})_{k,l}$, $(\RN{2})_{k,l}$, and $(\RN{3})_{k,l}$ one by one.

\subsubsection{\underline{\textbf{Term $(\RN{1})_{k,l}$}}}

Following from \eqref{eq: approx1-chebyshev1} and \eqref{eq: approx1-term2}, we show that with probability at least $1-\frac{1}{t^2}$, for $t>1$,
\begin{equation}
    \max_{k,l}|(\RN{1})_{k,l}|\leq tL \frac{M^2}{\sqrt{N}}.
    \label{proof-drl-correct-part1-final}
\end{equation}

\subsubsection{\underline{\textbf{Term $(\RN{4})_{k,l}$}}}
It follows from the Cauchy-schwarz inequality that
\[
|(\RN{4})_{k,l}| = \left|\frac{1}{N}\sum_{i=1}^{N} \Del{k}(X_i)\Del{l}(X_i)\right| \leq \sqrt{\frac{1}{N}\sum_{i=1}^{N} (\Del{k}(X_i))^2} \sqrt{\frac{1}{N}\sum_{i=1}^{N} (\Del{l}(X_i))^2}.
\]
where we denote $\Del{l}(X_i) = \hf{l}(X_i) - \f{l}(X_i)$ for each $l\in [L]$.
We leverage Markov's inequality as \eqref{eq: approx1-markov1}, then by union bound, it holds that: with probability at least $1-\frac{1}{t}$,
\begin{equation}
    \max_{k,l}|(\RN{4})_{k,l}|\leq tL\delta_n^2.
    \label{proof-drl-correct-part4-final}
\end{equation}

\subsubsection{\underline{\textbf{Term $(\RN{2})_{k,l}$}}}
For the observation $j\in \B_l$, we have $\Y{l}_j = \f{l}(X_j) + \eps{l}_j$, and express $(\RN{2})_{k,l}$ as follows:
{\small
\begin{equation}
    (\RN{2})_{k,l} = \frac{1}{N}\sum_{i=1}^{N} \hf{k}(X_i)\Del{l}(X_i) - \frac{1}{n_l}\sum_{j=1}^{n_l}\hw{l}(X_j)\hf{k}(X_j)\Del{l}(X_j)  +  \frac{1}{n_l}\sum_{j=1}^{n_l}\hw{l}(X_j)\hf{k}(X_j)\varepsilon_j^{(l)}.
    \label{proof-drl-correct-part2-decompose-1}
\end{equation}}
After investigating the above decomposed parts on the right-hand side, the first and the second components have the same population mean  $\E_{\QQ}\hf{k}(X)\Del{l}(X)$, if we replace $\hw{l}$ with $\w{l}$. Therefore, we can decompose the first two components in \eqref{proof-drl-correct-part2-decompose-1} as follows:
{\small
\begin{equation}
    \begin{aligned}
        &\underbrace{\left(\frac{1}{N}\sum_{i=1}^{N} \hf{k}(X_i)\Del{l}(X_i) - \E_{\QQ}\hf{k}(X)\Del{l}(X)\right)}_{(\RN{2}-1)_{k,l}} \\
        &\quad - \underbrace{\left(\frac{1}{n_l}\sum_{j=1}^{n_l}\w{l}(X_j)\hf{k}(X_j)\Del{l}(X_j) - \E_{\QQ}\hf{k}(X)\Del{l}(X)\right)}_{(\RN{2}-2)_{k,l}}\\
        &\quad - \underbrace{\left(\frac{1}{n_l}\sum_{j=1}^{n_l}\left(\hw{l}(X_j)-\w{l}(X_j)\right)\hf{k}(X_j)\Del{l}(X_j)\right)}_{(\RN{2}-3)_{k,l}}.
    \end{aligned}
    \label{proof-drl-correct-part2-decompose-2}
\end{equation}
}
The third component in \eqref{proof-drl-correct-part2-decompose-1} can be further decomposed as
{\small
\begin{equation}
    \underbrace{\left(\frac{1}{n_l}\sum_{j=1}^{n_l}\w{l}(X_j)\hf{k}(X_j)\varepsilon_j^{(l)}\right)}_{(\RN{2}-4)_{k,l}} +
        \underbrace{\left(\frac{1}{n_l}\sum_{j=1}^{n_l}\left(\hw{l}(X_j)-\w{l}(X_j)\right)\hf{k}(X_j)\varepsilon_j^{(l)}\right)}_{(\RN{2}-5)_{k,l}}.
    \label{proof-drl-correct-part2-decompose-3}
\end{equation}}
It is noteworthy that for terms $(\RN{2}-2)_{k,l}$ and $(\RN{2}-4)_{k,l}$, each observation $j$ is originally drawn from the $l$-th source population $\PP{l}$, however with the true density ratio $\w{l}$, it can be viewed as being drawn from target population $\QQ$ instead.
Due to sample splitting, both estimators $\hf{l}_\A$ and $\hw{l}_\A$ are independent of $\{(X_j, \eps{l}_j)\}_{j\in \B_l}$, thus the term $(\RN{2}-2)_{k,l}$, $(\RN{2}-4)_{k,l}$, and $(\RN{2}-5)_{k,l}$ is mean 0 by construction. In addition, the term  $(\RN{2}-1)_{k,l}$ is mean 0 as well.
In the following, we will construct the upper bounds for the $(\RN{2}-1)_{k,l}$, $(\RN{2}-2)_{k,l}$, and $(\RN{2}-4)_{k,l}$, subsequently. After that, we will continue studying the components $(\RN{2}-3)_{k,l}$, and $(\RN{2}-5)_{k,l}$.
\begin{itemize}
    \item \underline{{\textbf{Component $(\RN{2}-1)_{k,l}$}:}}
\end{itemize}
The following inequality holds due to Chebyshev's inequality that for any $a_1>0$,
\[
\prob\left\{ \left|(\RN{2}-1)_{k,l}\right| \geq a_1 \right\} \leq \frac{\Var_\QQ\left(\hf{k}(X)\Del{l}(X)\right)}{N a_1^2}.
\]
To upper bound $\Var_\QQ\left(\hf{k}(X) \Del{l}(X)\right)$, the followings hold that
{\small
\begin{equation}
\begin{aligned}
    \Var_\QQ\left(\hf{k}(X)\Del{l}(X)\right) \leq \E_\QQ \left[\left(\hf{k}(X)\Del{l}(X)\right)^2\right] = \E_\QQ\left[\left(\f{k}(X)+\Del{l}(X)\right)^2 \left(\Del{l}(X)\right)^2\right].
\end{aligned}
\label{proof-drl-correct-part2-VAR1-interm1}
\end{equation}}
It follows from the Cauchy-Schwarz inequality that:
{\small\begin{equation}
    \E_\QQ\left[\left(\f{k}(X)+\Del{l}(X)\right)^2 \left(\Del{l}(X)\right)^2\right] \leq \sqrt{\E_\QQ\left[\left(\f{k}(X)+\Del{l}(X)\right)^4 \right]} \sqrt{\E_\QQ\left[ \left(\Del{l}(X)\right)^4 \right]}.
    \label{proof-drl-correct-part2-VAR1-interm2}
\end{equation}}
With the fourth-moment Minkowski's inequality, we have
\begin{equation}
    \left(\E_\QQ\left[\left(\f{k}(X)+\Del{l}(X)\right)^4\right]\right)^{1/4} \leq \left(\E_\QQ\left[\f{k}(X)^4\right]\right)^{1/4} + \left(\E_\QQ\left[\Del{l}(X)^4\right]\right)^{1/4}.
    \label{proof-drl-correct-part2-VAR1-interm3}
\end{equation}
Combining \eqref{proof-drl-correct-part2-VAR1-interm1},\eqref{proof-drl-correct-part2-VAR1-interm2}, and \eqref{proof-drl-correct-part2-VAR1-interm3}, we establish that
{\small
\begin{equation}
\begin{aligned}
    \Var_\QQ\left(\hf{k}(X)\Del{l}(X)\right) 
    &\leq \E_\QQ \left[\left(\hf{k}(X)\Del{l}(X)\right)^2\right]\\
    &\leq \left[\left(\E_\QQ\left[\f{k}(X)^4\right]\right)^{1/4} + \left(\E_\QQ\left[\Del{l}(X)^4\right]\right)^{1/4} \right]^2 \left[\left(\E_\QQ\left[\Del{l}(X)^4\right]\right)^{1/4} \right]^2 \\
    &\leq (M + \delta_n)^2 \delta_n^2.
\end{aligned}
\label{proof-drl-correct-part2-VAR1}
\end{equation}}
With Chebyshev's inequality, we apply the above inequality in \eqref{proof-drl-correct-part2-VAR1} and establish that for any $a_1>0$,
\[
\prob\left\{ \left|(\RN{2}-1)_{k,l}\right| \geq a_1 \right\} \leq \frac{\Var_\QQ\left(\hf{k}(X)\Del{l}(X)\right)}{N a_1^2} \leq \frac{(M + \delta_n)^2 \delta_n^2}{N a_1^2}. 
\]
We set $a_1 = \frac{t L}{\sqrt{N}}(M + \delta_n) \delta_n$, then by union bound with probability $1-\frac{1}{t^2}$,
\begin{equation}
    \max_{k,l}\left|(\RN{2}-1)_{k,l}\right|\leq \frac{t L}{\sqrt{N}}(M + \delta_n) \delta_n.
\label{proof-drl-correct-part2-1}
\end{equation}
\begin{itemize}
    \item \underline{{\textbf{Component $(\RN{2}-2)_{k,l}$}:}}
\end{itemize}
Following the same argument in the analysis of $(\RN{2}-1)_{k,l}$, we can construct the similar arguments for the term $(\RN{2}-2)_{k,l}$. With Chebyshev's inequality, for any $a_2>0$,
\[
\prob\left\{ \left|(\RN{2}-2)_{k,l}\right| \geq a_2 \right\} \leq \frac{\Var_\QQ\left(\hf{k}(X)\Del{l}(X)\right)}{n_l a_2^2} \leq \frac{(M + \delta_n)^2 \delta_n^2}{n_l a_2^2}.
\]
We set $a_2 = \frac{t L}{\sqrt{n/2}}(M + \delta_n)\delta_n$, then by union bound with probability $1 - \frac{1}{t^2}$,
\begin{equation}
    \max_{k,l}\left|(\RN{2}-2)_{k,l}\right| \leq \frac{tL}{\sqrt{n/2}}(M + \delta_n)\delta_n .
\label{proof-drl-correct-part2-2}
\end{equation}

\begin{itemize}
    \item \underline{\textbf{Component $(\RN{2}-4)_{k,l}$}:}
\end{itemize}
We apply Chebyshev's inequality, the following inequality holds for any $a_3>0$
\[
\prob\left\{ \left|(\RN{2}-4)_{k,l}\right| \geq a_3\right\} \leq \frac{\Var_\QQ\left(\hf{k}(X)\varepsilon^{(l)}\right)}{n_l a_3^2}.
\]
Now we construct the upper bound for $\Var_\QQ\left(\hf{k}(X)\eps{l}\right)$:
\begin{multline*}
    \Var_\QQ\left(\hf{k}(X)\eps{l}\right) \leq \E_\QQ\left[(\hf{k}(X)\eps{l})^2\right] \\= \E_\QQ\left[(\hf{k}(X))^2 \E[(\eps{l})^2|X]\right] =\E_\QQ\left[(\hf{k}(X))^2 \E[(\eps{l})^2]\right].
\end{multline*}
where the last inequality is due to the independence of $\eps{l}$ and target covariates.
With the assumption \ref{ass: eigen} requiring the bounded moments for $\eps{l}$, it follows that
\begin{equation}
    \Var_\QQ\left(\hf{k}(X)\eps{l}\right) 
    \leq \sigma_\varepsilon^2 \E_\QQ\left[(\hf{k}(X))^2\right] 
    = \sigma_\varepsilon^2 \E_\QQ\left[\left(\f{k}(X) + \Del{k}(X)\right)^2\right].
    \label{proof-drl-correct-part2-VAR2-interm1}
\end{equation}
With the second-moment Minkowski's inequality, it holds that
\begin{equation}
    \left(\E_\QQ\left[\left(\f{k}(X) + \Del{k}(X)\right)^2\right]\right)^{1/2} \leq 
    \left(\E_\QQ\left[\f{k}(X)^2\right]\right)^{1/2} + \left(\E_\QQ\left[\Del{l}(X)^2\right]\right)^{1/2}.
    \label{proof-drl-correct-part2-VAR2-interm2}
\end{equation}
Combining \eqref{proof-drl-correct-part2-VAR2-interm1} and \eqref{proof-drl-correct-part2-VAR2-interm2}, we construct the following upper bound,
\begin{equation}
    \Var_\QQ\left(\hf{k}(X)\eps{l}\right) \leq 
    \sigma_\varepsilon^2\left[\left(\E_\QQ\left[\f{k}(X)^2\right]\right)^{1/2} + \left(\E_\QQ\left[\Del{l}(X)^2\right]\right)^{1/2}\right]^{2} \leq \sigma_\varepsilon^2 (M + \delta_n)^2.
    \label{proof-drl-correct-part2-VAR2}
\end{equation}
Now with the upper bound result in \eqref{proof-drl-correct-part2-VAR2}, we construct for any $a_3>0$,
\[
\prob\left\{ \left|(\RN{2}-4)_{k,l}\right| \geq a_3\right\} \leq \frac{\Var_\QQ\left(\hf{k}(X)\varepsilon^{(l)}\right)}{n_l a_3^2} \leq \frac{\sigma_{\varepsilon}^2(M + \delta_n)^2}{n_l a_3^2}.
\]
Set $a_3 = \frac{tL }{\sqrt{n/2}} \sigma_\varepsilon(M + \delta_n)$. Then by union bound with probability $1 - \frac{1}{t^2}$,
\begin{equation}
\max_{k,l}\left|(\RN{2}-4)_{k,l}\right|\leq \frac{t L}{\sqrt{n/2}} \sigma_\varepsilon (M + \delta_n).
\label{proof-drl-correct-part2-4}
\end{equation}
Now in the decomposition of $(\RN{2})_{k,l}$ in \eqref{proof-drl-correct-part2-decompose-2} and \eqref{proof-drl-correct-part2-decompose-3}, the only terms left are terms $(\RN{2}-3)_{k,l}$ and $(\RN{2}-5)_{k,l}$. In the following, we will leverage the approximation error of the density ratio estimator $\hw{l}$ to quantify these two terms further.

\begin{itemize}
    \item \underline{\textbf{Component $(\RN{2}-3)_{k,l}$:}}
\end{itemize}
We re-express $(\RN{2}-3)_{k,l}$ as follows:
\[
(\RN{2}-3)_{k,l} = \frac{1}{n_l}\sum_{j=1}^{n_l}\left(\frac{\hw{l}(X_j)}{\w{l}(X_j)}-1\right)\w{l}(X_j)\hf{k}(X_j)\Del{l}(X_j).
\]
It follows from the Cauchy-Schwarz inequality that
{\small\begin{equation}
    |(\RN{2}-3)_{k,l}| \leq \sqrt{\frac{1}{n_l} \sum_{j=1}^{n_l} \w{l}(X_j) \left(\frac{\hw{l}(X_j)}{\w{l}(X_j)}-1\right)^2} \sqrt{\frac{1}{n_l} \sum_{j=1}^{n_l} \w{l}(X_j)\left(\hf{k}(X_j)\Del{l}(X_j)\right)^2}.
    \label{proof-drl-correct-part2-3-interm1}
\end{equation}}
Since the density ratio estimator $\hw{l}(\cdot)$ is independent of the data in $\B_l$, we have
\[
\E_{\PP{l}}\left[\frac{1}{n_l} \sum_{j=1}^{n_l} \w{l}(X_j) \left(\frac{\hw{l}(X_j)}{\w{l}(X_j)}-1\right)^2\right] = \E_{\QQ} \left(\frac{\hw{l}(X_j)}{\w{l}(X_j)}-1\right)^2.
\]
Therefore, the upper bound for the first component on the right-hand side of \eqref{proof-drl-correct-part2-3-interm1} can be established by Markov's inequality. For any $a_1>0$,
\[
\prob\left\{ 
\frac{1}{n_l} \sum_{j=1}^{n_l} \w{l}(X_j)\left(\frac{\hw{l}(X_j)}{\w{l}(X_j)}-1\right)^2 
\geq a_1
\right\} \leq \frac{\E_{\QQ} \left(\frac{\hw{l}(X_j)}{\w{l}(X_j)}-1\right)^2}{a_1}
\leq \frac{\eta_\omega^2}{a_1}.
\]
The similar strategy using Markov's inequality can be applied to the second component on the right-hand side of \eqref{proof-drl-correct-part2-3-interm1}, such that for any $a_2>0$:
\[
\prob\left\{
\frac{1}{n_l} \sum_{j\in \B_l} \w{l}(X_j) \left(\hf{k}(X_j)\Del{l}(X_j)\right)^2
\geq a_2
\right\} \leq \frac{\E_{\QQ} \left(\hf{k}(X)\Del{l}(X)\right)^2}{a_2}\leq \frac{(M + \delta_n)^2 \delta_n^2}{a_2}.
\]
where the last inequality follows the result in \eqref{proof-drl-correct-part2-VAR1}. We then take $a_1 = tL \eta_\omega^2$ and $a_2 = t L(M + \delta_n)^2 \delta_n^2$. By the union bound, with probability at least $1-\frac{2}{t}$, the following inequalities hold
\[
\max_{l\in [L]}\left|\frac{1}{n_l} \sum_{j=1}^{n_l} \w{l}(X_j)\left(\frac{\hw{l}(X_j)}{\w{l}(X_j)}-1\right)^2 \right| \leq t L \eta_\omega^2,
\]
and
\[
\max_{l\in [L]}\left|\frac{1}{n_l} \sum_{j=1}^{n_l} \w{l}(X_j) \left(\hf{k}(X_j)\Del{l}(X_j)\right)^2\right| \leq t L (M + \delta_n)^2 \delta_n^2.
\]
Combining the above two inequalities together with \eqref{proof-drl-correct-part2-3-interm1}, we establish that with probability greater than $1-\frac{2}{t}$,
\begin{equation}
    \max_{k,l} |(\RN{2}-3)_{k,l}| \leq t L (M + \delta_n) \delta_n \eta_\omega.
    \label{proof-drl-correct-part2-3}
\end{equation}
\begin{itemize}
    \item \underline{\textbf{Component $(\RN{2}-5)_{k,l}$:}}
\end{itemize}
For the term $\frac{1}{n_l}\sum_{j=1}^{n_l}\left(\hw{l}(X_j)-\w{l}(X_j)\right)\hf{k}(X_j)\varepsilon_j^{(l)}$, its expectation satisfies mean 0 as follows:
{\small
\begin{multline*}
    \E_{\PP{l}}\left[\frac{1}{n_l}\sum_{j=1}^{n_l}\left(\hw{l}(X_j)-\w{l}(X_j)\right)\hf{k}(X_j)\varepsilon_j^{(l)}\right] \\= \E\left[\frac{1}{n_l}\sum_{j=1}^{n_l}\left(\hw{l}(X_j)-\w{l}(X_j)\right)\hf{k}(X_j)\left[\E[\varepsilon_j^{(l)}|X_j\right]\right] = 0.
\end{multline*}
}
We apply Chebyshev's inequality to upper bound the term, such that $\forall a>0$:
\begin{equation*}
    \prob\left\{\left|(\RN{2}-5)_{k,l}) - 0\right|\geq a\right\} \leq \frac{\Var_{\PP{l}}\left((\hw{l}(X)-\w{l}(X))\hf{k}(X)\varepsilon^{(l)}\right)}{n_la^2}.
\end{equation*}
In the following, we will establish the upper bound for the variance term. With the assumption \ref{ass: eigen}, it holds that
\begin{equation}
    \Var_{\PP{l}}\left((\hw{l}(X)-\w{l}(X))\hf{k}(X)\varepsilon^{(l)}\right) \leq  \sigma_\varepsilon^2\E_{\PP{l}}\left[\left((\hw{l}(X)-\w{l}(X))\hf{k}(X)\right)^2\right].
    \label{proof-drl-correct-part2-5-1}
\end{equation}
The right-hand side of the above inequality can be rewritten as follows:
\[
\E_{\PP{l}}\left[\left((\hw{l}(X)-\w{l}(X))\hf{k}(X)\right)^2\right] = \E_{\PP{l}}\left[\w{l}(X)^2\left(\frac{\hw{l}(X)}{\w{l}(X)}-1\right)^2\hf{k}(X)^2\right].
\]
It follows from the Cauchy-Schwarz inequality that
{\small
\begin{multline}
    \E_{\PP{l}}\left[\w{l}(X)^2\left(\frac{\hw{l}(X)}{\w{l}(X)}-1\right)^2\hf{k}(X)^2\right] \\\leq
\sqrt{\E_{\PP{l}}\left[\w{l}(X)^3\left(\frac{\hw{l}(X)}{\w{l}(X)}-1\right)^4\right] \E_{\PP{l}}\left[\w{l}(X)\left(\hf{k}(X)\right)^4\right]}.
\label{proof-drl-correct-part2-5-2}
\end{multline}}
We further upper bound both terms on the right hand side in \eqref{proof-drl-correct-part2-5-2}. For the first term, we have
\begin{equation}
    \E_{\PP{l}}\left[\w{l}(X)^3\left(\frac{\hw{l}(X)}{\w{l}(X)}-1\right)^4\right] = \E_\QQ\left[\w{l}(X)^2\left(\frac{\hw{l}(X)}{\w{l}(X)}-1\right)^4\right].
    \label{proof-drl-correct-part2-5-3}
\end{equation}
Since $1 + \w{l}(X)^4\geq 2\w{l}(X)^2$, we have
\begin{equation}
    \E_{\QQ}\left[\w{l}(X)^2\left(\frac{\hw{l}(X)}{\w{l}(X)}-1\right)^4\right]\leq \frac{1}{2}\E_\QQ\left[\left(\frac{\hw{l}(X)}{\w{l}(X)}-1\right)^4 + \left(\hw{l}(X)-\w{l}(X)\right)^4\right] \leq \eta_\omega^4.
    \label{proof-drl-correct-part2-5-3}
\end{equation}
For the other term, it holds that
\begin{equation}
    \E_{\PP{l}}\left[\w{l}(X)\left(\hf{k}(X)\right)^4\right] = \E_{\QQ}\left[\left(\hf{k}(X)\right)^4\right] = \E_{\QQ}\left[\left(\f{k}(X) + \Del{k}(X)\right)^4\right].
    \label{proof-drl-correct-part2-5-4}
\end{equation}
It follows from the fourth moment Minowski's inequality that
\begin{equation}
    \left(\E_{\QQ}\left[\left(\f{k}(X) + \Del{k}(X)\right)^4\right]\right)^{1/4} \leq \left(\E_\QQ[\f{k}(X)^4]\right)^{1/4} + \left(\E_\QQ[\Del{k}(X)^4]\right)^{1/4}\leq M + \delta_n.
    \label{proof-drl-correct-part2-5-5}
\end{equation}
Combining \eqref{proof-drl-correct-part2-5-1},\eqref{proof-drl-correct-part2-5-2},\eqref{proof-drl-correct-part2-5-3},\eqref{proof-drl-correct-part2-5-4}, and \eqref{proof-drl-correct-part2-5-5}, we establish the upper bound for the variance:
\begin{equation*}
    \Var_{\PP{l}}\left((\hw{l}(X)-\w{l}(X))\hf{k}(X)\varepsilon^{(l)}\right) \leq \sigma_\varepsilon^2 (M + \delta_n)^2\eta_\omega^2.
\end{equation*}
Therefore, with Chebyshev's inequality, $\forall a>0$:
\begin{equation*}
    \prob\left\{\left|(\RN{2}-5)_{k,l})\right|\geq a\right\} \leq \frac{\Var_{\PP{l}}\left((\hw{l}(X)-\w{l}(X))\hf{k}(X)\varepsilon^{(l)}\right)}{n_la^2} 
    \leq \frac{\sigma_\varepsilon^2 (M + \delta_n)^2\eta_\omega^2}{n_la^2} .
\end{equation*}
Let $a = t L \frac{\sigma_\varepsilon (M + \delta_n)\eta_\omega}{\sqrt{n/2}} $, and by union bound, with probability at least $1-\frac{1}{t^2}$, the following holds:
\begin{equation}
    \max_{k,l}\left|(\RN{2}-5)_{k,l}\right| \leq t L \frac{\sigma_\varepsilon (M + \delta_n)\eta_\omega}{\sqrt{n/2}}.
    \label{proof-drl-correct-part2-5}
\end{equation}
Combining all components for $(\RN{2})_{k,l}$ in \eqref{proof-drl-correct-part2-1},\eqref{proof-drl-correct-part2-2},\eqref{proof-drl-correct-part2-4},\eqref{proof-drl-correct-part2-3}, and \eqref{proof-drl-correct-part2-5}, we have with probability at least $1-\frac{1}{t}-\frac{1}{t^2}$:
{\small
\begin{equation}
\begin{aligned}
    \max_{k,l}|(\RN{2})_{k,l}|&\leq \max_{k,l}|(\RN{2}-1)_{k,l}|+\max_{k,l}|(\RN{2}-2)_{k,l}|+\max_{k,l}|(\RN{2}-3)_{k,l}|+\max_{k,l}|(\RN{2}-4)_{k,l}|+\max_{k,l}|(\RN{2}-5)_{k,l}|\\
    &\lesssim \frac{M+\delta_n}{\sqrt{n}} + \frac{(M+\widetilde{\delta_n})\delta_n}{\sqrt{n\wedge N}} + (M+\widetilde{\delta_n})\delta_n\eta_\omega + \frac{(M+\widetilde{\delta_n})\eta_\omega}{\sqrt{n}}.
\end{aligned}
\label{proof-drl-correct-part2-final}
\end{equation}}

\subsubsection{\underline{\textbf{Term $(\RN{3})_{k,l}$}}}

For part $(\RN{3})_{k,l}$, we just switch the roles indices $k$ and $l$, the same arguments holds just as ones for $(\RN{2})_{k,l}$, that is, with probability at least $1-\frac{1}{t}-\frac{1}{t^2}$:
\begin{equation}
    \max_{k,l}|(\RN{3})_{k,l}|\lesssim \frac{M+\delta_n}{\sqrt{n}} + \frac{(M+\widetilde{\delta_n})\delta_n}{\sqrt{n\wedge N}} + (M+\widetilde{\delta_n})\delta_n\eta_\omega + \frac{(M+\widetilde{\delta_n})\eta_\omega}{\sqrt{n}}.
    \label{proof-drl-correct-part3-final}
\end{equation}

\subsubsection{\underline{\textbf{Proof of \eqref{eq: error correct}}}}
Till now, we've finished analyzing all parts in the error decomposition of $\hGamma^\A_{k,l}$. We combine the results for terms $(\RN{1})_{k,l}$, $(\RN{2})_{k,l}$, $(\RN{3})_{k,l}$, and $(\RN{4})_{k,l}$ in \eqref{proof-drl-correct-part1-final}, \eqref{proof-drl-correct-part2-final}, \eqref{proof-drl-correct-part3-final}, and \eqref{proof-drl-correct-part4-final}, respectively. We construct the following: with the probability at least $1-\frac{1}{t}-\frac{1}{t^2}$:
\begin{equation*}
\begin{aligned}
    \left\|\hGamma^\A - \Gamma\right\|_\infty &\leq \max_{k,l}|(\RN{1})_{k,l}|+\max_{k,l}|(\RN{2})_{k,l}|+\max_{k,l}|(\RN{3})_{k,l}|+\max_{k,l}|(\RN{4})_{k,l}| \\
    &\lesssim t \cdot \left(\delta_n^2 + \frac{M^2}{\sqrt{N}} + \frac{M+\delta_n}{\sqrt{n}} + \frac{(M+\widetilde{\delta_n})\delta_n}{\sqrt{n\wedge N}} + (M+\widetilde{\delta_n})\delta_n\eta_\omega + \frac{(M+\widetilde{\delta_n})\eta_\omega}{\sqrt{n}}\right).
\end{aligned}
\end{equation*}
Next, we switch the roles of sample split dataset $\{\A_l\}_{l\in [L]}$ and $\{\B_l\}_{l\in [L]}$, and construct the same upper bound for $\left\|\hGamma^\B - \Gamma\right\|_\infty$. By construction, $\hGamma = \frac{1}{2}(\hGamma^\A + \hGamma^\B)$. We apply the triangle inequality and construct the upper bound for the error of $\hGamma$, that is,  with the probability at least $1-\frac{1}{t}-\frac{1}{t^2}$:
\begin{equation}
    \left\|\hGamma - \Gamma\right\|_\infty \lesssim t \cdot \left(\delta_n^2 + \frac{M^2}{\sqrt{N}} + \frac{M+\delta_n}{\sqrt{n}} + \frac{(M+\widetilde{\delta_n})\delta_n}{\sqrt{n\wedge N}} + (M+\widetilde{\delta_n})\delta_n\eta_\omega + \frac{(M+\widetilde{\delta_n})\eta_\omega}{\sqrt{n}}\right)
\end{equation}
With the error of $\hGamma$ bounded, we move forward bounding the error of the bias-corrected DRoL estimator $\hfH$. We follow the same argument as \eqref{eq: approx1-basic} but replace $\tGamma$ with $\hGamma$. Then with probability at least $1-\frac{1}{t}-\frac{1}{t^2}$, we have
{\small
\begin{equation}
\begin{aligned}
    &\left\|\hfH - \fH\right\|_{\ell_2(\QQ)} \lesssim \left[\delta_n + \left(\frac{L\|\hGamma - \Gamma\|_\infty}{\lambda_{\rm min}(\Gamma)} \wedge \rho_\HH\right)M\right] \\
    &\lesssim \delta_n + t\cdot \max\left\{\frac{M}{\lambda_{\rm min}(\Gamma)}\left(\delta_n^2 + \frac{M^2}{\sqrt{N}} + \frac{M+\delta_n}{\sqrt{n}} + \frac{(M+\widetilde{\delta_n})\delta_n}{\sqrt{n\wedge N}} + (M+\widetilde{\delta_n})\delta_n\eta_\omega + \frac{(M+\widetilde{\delta_n})\eta_\omega}{\sqrt{n}}\right), M\rho_\HH\right\}.
\end{aligned}
    \label{proof-drl-correct-f-diff}
\end{equation}
}

\subsection{Proof of Corollary \ref{cor: high-dim}}
\label{sec: appendix-density}
We start with the analysis of $\delta_n$ and $\delta_n$. To simplify the discussion, we only present the analysis of $\widehat{\beta}^{(l)}$ fitted over the full data, and the estimators $\widehat{\beta}^{(l)}$ and $\widehat{\beta}^{(l)}_\B$ fitted on the subset data $\A_l$ and $\B_l$ shall follow the same argument. and Known that $\XQ\sim \QQ_X$ is sub-Gaussian random vector, as in the assumption \ref{ass: linear}, and $\XQ$ is independent of $\widehat{\beta}^{(l)} - \beta^{(l)}$, we establish that
    $(\XQ)^\intercal \frac{\widehat{\beta}^{(l)} - \beta^{(l)}}{\|\widehat{\beta}^{(l)} - \beta^{(l)}\|_2} $
    is a sub-Gaussian random variable, following the sub-Gaussian random vector definition (Def 5.22) in the book \cite{vershynin2010introduction}. Then for all $q\geq 1$:
    {\small
    \[
    \left(\E_{\QQ_X}\left[\left((\XQ)^\intercal \frac{\widehat{\beta}^{(l)} - \beta^{(l)}}{\|\widehat{\beta}^{(l)} - \beta^{(l)}\|_2}\right)^q\right]\right)^{1/q}\leq C\sqrt{q}.
    \]}
    Since $\widehat{\beta}^{(l)}$ satisfies \eqref{eq: beta-rate},  the above inequality implies that
    \[
    \|(\XQ)^\intercal(\widehat{\beta}^{(l)} - \beta^{(l)})\|_{\ell_2(\QQ)}\leq \sqrt{2}C \|\widehat{\beta}^{(l)} - \beta^{(l)}\|_2 \lesssim \sqrt{\frac{s_\beta \log p}{n_l}},
    \]
    and
    \[
    \|(\XQ)^\intercal(\widehat{\beta}^{(l)} - \beta^{(l)})\|_{\ell_4(\QQ)}\leq 2C \|\widehat{\beta}^{(l)} - \beta^{(l)}\|_2 \lesssim \sqrt{\frac{s_\beta \log p}{n_l}}.
    \]
    Therefore,
    \[
    \delta_n \lesssim \sqrt{s_\beta \log p/n}.
    \]
    
    \noindent Next, we move forward to the analysis of $\eta_\omega$ and $\eta_\omega$. For simplicity, we only analyze the density ratio estimator $\widehat{\omega}^{(l)}_\A$ fitted on the split sample subset $\A_l$. We replace the notations $\widehat{\gamma}^{(l)}_\A, \widehat{\omega}^{(l)}_\A$ with $\widehat{\gamma}^{(l)},\widehat{\omega}^{(l)}$ for the use of convenience.
    Assuming that the density ratio models in \eqref{eq: density ratio} are correctly specified, we follow the regime in the \cite{qin1998inferences}, and denote $\pi(x)$ as the marginal density for the random vector $\widetilde{X}^{(l)}_i$. We define $p_l$ to be the marginal probability of $\{G^{(l)}_i = 0\}$ in the population, i.e., $p_l = p(G^{(l)}_i = 0) = \int p(G^{(l)}_i = 0|\widetilde{X}^{(l)}_i = x) \pi(x) dx$. Due to the Assumption \ref{ass: linear}, $p_l \in [c_{\rm min}, \frac{1}{2})$ by definition. In the following discussion, we treat $p_l$ as an unknown constant satisfying $p_l \in (0,\frac{1}{2})$ without reaching the boundaries.
    
    \noindent We can see that the density ratio is:
    \[
    \frac{p(x| G^{(l)}_i = 1)}{p(x| G^{(l)}_i = 0)} = \frac{p_l}{1-p_l} \exp(x^\intercal \gamma^{(l)}).
    \]
    The random variable $G_i^{(l)}$ for $i\in [n_G]$ with $n_G = |\A_l| + N$ is bernoulli distributed with $p(G_i^{(l)}=0) = p_l$ and $p(G_i^{(l)}=1) = 1-p_l$.  We consider the independent $U_i \sim {\rm Unif}(0,1)$ for $i\in [n_G]$, then $n_l = \sum_{i=1}^{n_G}\ind_{[U_i\leq p_l]}$.

    \noindent Again, known that $\XQ\sim \QQ_X$ is sub-Gaussian random vector,
    it holds that
    $(\XQ)^\intercal \frac{\widehat{\gamma}^{(l)} - \gamma^{(l)}}{\|\widehat{\gamma}^{(l)} - \gamma^{(l)}\|_2} $
    is a sub-Gaussian random variable. Thus for all $q\geq1$:
    {\small
    \[
    \left(\E_{\QQ_X}\left[\left((\XQ)^\intercal \frac{\widehat{\gamma}^{(l)} - \gamma^{(l)}}{\|\widehat{\gamma}^{(l)} - \gamma^{(l)}\|_2}\right)^q\right]\right)^{1/q}\leq C\sqrt{q}.
    \]}
    The above inequality implies that
    \begin{equation*}
        \|(\XQ)^\intercal(\widehat{\gamma}^{(l)} - \gamma^{(l)})\|_{\QQ,q}\leq \sqrt{q}C \|\widehat{\gamma}^{(l)} - \gamma^{(l)}\|_2. 
    \end{equation*}
    Together with $\widehat{\gamma}^{(l)}$ satisfying \eqref{eq: gamma-rate}, we know that for all $q\geq 1$,
    \begin{equation}
        \|(\XQ)^\intercal(\widehat{\gamma}^{(l)} - \gamma^{(l)})\|_{\QQ,q}\leq \sqrt{q}C \|\widehat{\gamma}^{(l)} - \gamma^{(l)}\|_2 \lesssim \sqrt{\frac{s_\gamma \log p}{n_G}} \longrightarrow 0,
        \label{proof-coro-linear-5}
    \end{equation}
    as $n_G\rightarrow \infty$.
    
    \noindent For the term $\frac{\hw{l}(x)}{\w{l}(x)} - 1$, we analyze it as follows:
    \begin{align*}
        \frac{\hw{l}(x)}{\w{l}(x)} - 1 &= \frac{\frac{|\A_l|}{N}\exp(x^\intercal \widehat{\gamma}^{(l)})}{\frac{p_l}{1-p_l}\exp(x^\intercal \gamma^{(l)})} - 1
        = \frac{\frac{|\A_l|}{N}\exp(x^\intercal \widehat{\gamma}^{(l)}) - \frac{p_l}{1-p_l}\exp(x^\intercal \gamma^{(l)})}{\frac{p_l}{1-p_l}\exp(x^\intercal \gamma^{(l)})} \\
        &= \frac{\frac{|\A_l|}{N}\left(\exp(x^\intercal \widehat{\gamma}^{(l)}) - \exp(x^\intercal\gamma^{(l)})\right) + \left(\frac{|\A_l|}{N} - \frac{p_l}{1-p_l}\right)\exp(x^\intercal \gamma^{(l)})}{\frac{p_l}{1-p_l}\exp(x^\intercal \gamma^{(l)})} \\
        &= \frac{|\A_l|/N}{p_l/(1-p_l)}\left[\exp\left(x^\intercal(\widehat{\gamma}^{(l)} - \gamma^{(l)})\right) - 1\right] + \frac{|\A_l|/N - p_l/(1-p_l)}{p_l/(1-p_l)} .
    \end{align*}
    It follows from the triangle inequality that
    \begin{equation}
        \left|\frac{\hw{l}(x)}{\w{l}(x)} - 1\right|\leq  \left|\frac{|\A_l|/N}{p_l/(1-p_l)}\right|\left|\exp\left(x^\intercal(\widehat{\gamma}^{(l)} - \gamma^{(l)})\right) - 1\right| + \left|\frac{|\A_l|/N - p_l/(1-p_l)}{p_l/(1-p_l)}\right|.
        \label{proof-coro-linear-6}
    \end{equation}
    \begin{Lemma}
        With probability greater than $1-2c_1$, for $c_1\in (0,\frac{1}{2})$, if $n_G > \frac{\log(1/c_1)}{\min\{(1-p_l)^2, p_l^2\}}$, it holds that
        \[
        \left|\frac{|\A_l|}{N} - \frac{p_l}{1-p_l}\right| \leq C_1 \sqrt{\frac{1}{n_G}},
        \]
        where $C_1 = \frac{\sqrt{\log(1/c_1)}}{(\sqrt{2}-1)(1-p_l)^2}$.
        \label{lemma: proof-coro-linear}
    \end{Lemma}
    \noindent Applying the above Lemma \ref{lemma: proof-coro-linear} together with the inequality in \eqref{proof-coro-linear-6}, we establish that
    {\small
    \begin{equation}
        \left|\frac{\hw{l}(x)}{\w{l}(x)} - 1\right|\leq \left(1+C_1\sqrt{\frac{1}{n_G}}\right)\left|\exp\left(x^\intercal(\widehat{\gamma}^{(l)} - \gamma^{(l)})\right) - 1\right| + \frac{C_1}{p_l/(1-p_l)}\sqrt{\frac{1}{n_G}}.
         \label{proof-coro-linear-3}
    \end{equation}}
    We continue upper-bounding the term $\left|\exp\left(x^\intercal(\widehat{\gamma}^{(l)} - \gamma^{(l)})\right) - 1\right|$ in the above inequality \eqref{proof-coro-linear-3}.
    Due to the Mean Value Theorem,
    \[
    \exp\left(x^\intercal(\widehat{\gamma}^{(l)} - \gamma^{(l)})\right) - 1 = \theta \cdot x^\intercal(\widehat{\gamma}^{(l)} - \gamma^{(l)}),
    \]
    where $\theta$ takes some value between $0$ and $\exp(x^\intercal(\widehat{\gamma}^{(l)} - \gamma^{(l)}))$, and then
    \begin{equation}
        \left|\exp\left(x^\intercal(\widehat{\gamma}^{(l)} - \gamma^{(l)})\right) - 1\right| = \left|\theta\right| \cdot \left|x^\intercal(\widehat{\gamma}^{(l)} - \gamma^{(l)})\right|.
    \label{proof-coro-linear-7}
    \end{equation}
    {
    Since $(\XQ)^\intercal(\widehat{\gamma}^{(l)} - \gamma^{(l)})$ is subGaussian, we obtain that for all $t>0$,
    \[
    \prob\left\{\left|(\XQ)^\intercal(\widehat{\gamma}^{(l)} - \gamma^{(l)})\right| > t\right\} \leq \exp(1- t^2/K_1^2) .
    \]
    and $K_1\lesssim \sqrt{\frac{s_\gamma \log p}{n_G}}$ by the equivalence of sub-Gaussian property and the result in \eqref{proof-coro-linear-5}.
    Taking $t = \log 2$, we establish that with probability greater than $1-\exp(1-(\log 2)^2/K_1^2)$, the following event holds.
    \[
    \Omega_1 = \left\{\left|(\XQ)^\intercal(\widehat{\gamma}^{(l)} - \gamma^{(l)})\right| \leq \log 2\right\}.
    \]
    i.e., $\prob(\Omega_1)\geq 1-\exp(1-(\log 2)^2/K_1^2) = 1-p^{-c} - \exp(-cn)$. 

    \underline{Conditional on event $\Omega_1$}, we have
    \[
    |\theta| \leq \exp(\left|(\XQ)^\intercal(\widehat{\gamma}^{(l)} - \gamma^{(l)})\right|)\leq 2.
    \]
    Then, it follows from \eqref{proof-coro-linear-3} and \eqref{proof-coro-linear-7} that
    {\small
    \begin{equation}
        \left|\frac{\hw{l}(\XQ)}{\w{l}(\XQ)} - 1\right|\leq 2\left(1+C_1\sqrt{\frac{1}{n_G}}\right)\left|(\XQ)^\intercal(\widehat{\gamma}^{(l)} - \gamma^{(l)})\right| + \frac{C_1}{p_l/(1-p_l)}\sqrt{\frac{1}{n_G}}.
        \label{eq: proof-coro-linear-8}
    \end{equation}}
    Therefore, its $\ell_2$ and $\ell_4$ norms satisfy:
    \begin{equation}
    \begin{aligned}
        &\left\|\frac{\hw{l}}{\w{l}} - 1\right\|_{\ell_2(\QQ)} \leq 2\left(1+C_1\sqrt{\frac{1}{n_G}}\right)
        \left(\|(\XQ)^\intercal(\widehat{\gamma}^{(l)} - \gamma^{(l)})\|_{\ell_2(\QQ)}\right) + \frac{C_1}{p_l/(1-p_l)}\sqrt{\frac{1}{n_G}}\\
        &\left\|\frac{\hw{l}}{\w{l}} - 1\right\|_{\ell_4(\QQ)} \leq 2\left(1+C_1\sqrt{\frac{1}{n_G}}\right)
        \left(\|(\XQ)^\intercal(\widehat{\gamma}^{(l)} - \gamma^{(l)})\|_{\ell_4(\QQ)}\right) + \frac{C_1}{p_l/(1-p_l)}\sqrt{\frac{1}{n_G}}
    \end{aligned}
    \end{equation}
        Combining \eqref{proof-coro-linear-5}, we establish that
    \begin{equation}
        \max\left\{\left\|\frac{\hw{l}}{\w{l}} - 1\right\|_{\ell_2(\QQ)}\;,\;\left\|\frac{\hw{l}}{\w{l}} - 1\right\|_{\ell_4(\QQ)}\right\}\lesssim \sqrt{\frac{s_\gamma \log p}{n_G}} \lesssim \sqrt{\frac{s_\gamma \log p}{n + N}}.
        \end{equation}
    Next, we upper bound the term $\left\|\hw{l}-\w{l}\right\|_{\ell_4(\QQ)}$. By assumption \ref{ass: linear}, with probability larger than $1-p^{-c_0}$, the following holds:
    \[
    \w{l}(x) = \frac{p_l}{1-p_l} \exp(x^\top \gamma^{(l)}) \leq \frac{p_l}{1-p_l} \frac{1-c_{\rm min}}{c_{\rm min}}.
    \]
    It implies that
    {\small
    \begin{equation}
        \left\|\hw{l}-\w{l}\right\|_{\ell_4(\QQ)}=  \left\|\w{l}\left(\frac{\hw{l}}{\w{l}}-1\right)\right\|_{\ell_4(\QQ)} \leq
        \frac{p_l}{1-p_l} \frac{1-c_{\rm min}}{c_{\rm min}} \left\|\frac{\hw{l}}{\w{l}} - 1\right\|_{\ell_4(\QQ)} \lesssim \sqrt{\frac{s_\gamma \log p}{n_G}}\lesssim \sqrt{\frac{s_\gamma \log p}{n + N}}.
        \label{proof-coro-linear-density-2}
    \end{equation}}
    Therefore, we show that
    \[
    \eta_\omega\lesssim \sqrt{s_\gamma \log p/(n+N)}.
    \]
    
    }

\subsection{Proof of Theorem \ref{prop: reward-diff}}
We care about the reward difference of the population version DRoL $\fH$ and empirical version DRoL $\hfH$. For any fixed target distribution $\QQ$:
\[
\begin{aligned}
    \RR_\QQ(\fH) - \RR_\QQ(\hfH) &= \E_\QQ(Y - \fH(X))^2 - \E_\QQ(Y - \hfH(X))^2\\
    &= \E_\QQ(Y - \fH(X) + \fH(X) - \hfH(X))^2 - \E_\QQ(Y-\fH(X))^2 \\
    &= 2\E_\QQ\left[(Y-\fH(X))(\fH(X) - \hfH(X))\right] + \E_\QQ(\fH(X) - \hfH(X))^2. 
\end{aligned}
\]
We rewrite the target outcome as follows: $Y = f^\QQ(X) + \epsilon^\QQ$ with $f_\QQ(X) = \E_{\QQ}[Y|X]$ and $\E_\QQ[\epsilon^\QQ|X] = 0$. 
Thus
\[
\E_\QQ\left[(Y-\fH(X))(\fH(X) - \hfH(X))\right] = 
\E_\QQ\left[(f^\QQ(X)-\fH(X))(\fH(X) - \hfH(X))\right].
\]
By the Cauchy-Schwarz inequality, it holds that
\[
\E_\QQ\left[(f^\QQ(X)-\fH(X))(\fH(X) - \hfH(X))\right] \leq \sqrt{\E_\QQ[(f^\QQ(X)-\fH(X))^2]}\sqrt{\E_\QQ[(\fH(X) - \hfH(X))^2]}.
\]
It implies that
\[
\begin{aligned}
    \left|\RR_\QQ(\fH) - \RR_\QQ(\hfH)\right|&\leq 2\sqrt{\E_\QQ[(f^\QQ(X)-\fH(X))^2]}\|\hfH - \fH\|_{\ell_2(\QQ)} + \|\hfH - \fH\|_{\ell_2(\QQ)}^2\\
    &= 2\|f^\QQ - \fH\|_{\ell_2(\QQ)}\cdot \|\hfH - \fH\|_{\ell_2(\QQ)} + \|\hfH - \fH\|_{\ell_2(\QQ)}^2.
\end{aligned}
\]

\subsection{Proof of Proposition \ref{prop: shrink H}}
Since $\QQ_{Y|X} = \sum_{l=1}^L q_l^0 \cdot \PP{l}_{Y|X}$, the conditional outcome model for target distribution satisfies:
\begin{equation}
    f^\QQ(x) = \E_{\QQ}[Y|X=x] = \sum_{l=1}^L q_l^0 \cdot \E_{\PP{l}}[Y|X=x] = \sum_{l=1}^L q_l \cdot \f{l}(x)
    \label{eq: prop-shinkH-1}
\end{equation}
It follows from the identification theorem \ref{thm: identification} that
\begin{align*}
    \|\fH - f^\QQ\|_{\ell_2(\QQ)}^2 &= \E_{\QQ_X}\left[\sum_{l=1}^L (q_l^* - q_l^0)\cdot \f{l}(X)\right]^2 \leq 
    \E_{\QQ_X}\left[L \sum_{l=1}^L (q_l^* - q_l^0)^2 (\f{l}(X))^2 \right] \\
    &= L\sum_{l=1}^L(q_l^* - q_l^0)^2 \|\f{l}\|_{\ell_2(\QQ)}^2.
\end{align*}
If $\max_{q\in \HH}\|q - q^0\|_2 \leq \rho$, since in the above inequality $q^*\in \HH$, it follows that
\[
\|\fH - f^\QQ\|_{\ell_2(\QQ)}^2 \leq L\rho^2\max_{l\in [L]}\|\f{l}\|_{\ell_2(\QQ)}^2.
\]
which implies that $\|\fH - f^\QQ\|_{\ell_2(\QQ)} \leq  \rho\sqrt{L}\max_{l\in [L]}\|\f{l}\|_{\ell_2(\QQ)}$.

\section{Proofs of Lemmas}
\label{sec: proof-lemmas}
\subsection{Proof of Lemma \ref{lemma: approx}} 
\label{sec: proof-lemma-approx}
The proof essentially follows the proof of Lemma 6 in \cite{guo2020inference}. By definition of $q^*$, for any $t\in (0,1)$, we have
\[
(q^*)^\intercal \Gamma q^* \leq [q^* + t(\widehat{q} - q^*)]^\intercal \Gamma [q^* + t(\widehat{q} - q^*)].
\]
hence
\[
0\leq 2 t(q^*)^\intercal \Gamma (\widehat{q} - q^*) + t^2 (\widehat{q} - q^*)^\intercal \Gamma (\widehat{q} - q^*).
\]
By taking $t \rightarrow 0_+$, we have 
\begin{equation}
    (q^*)^\intercal \Gamma (\widehat{q} - q^*) \geq 0
    \label{eq: lemma-approx-1}.
\end{equation}
Similarly, by definition of $\widehat{q}$, for any $t\in (0,1)$, we have
\[
\widehat{q}\widehat{\Gamma}\widehat{q} \leq \left[\widehat{q} + t(q^* - \widehat{q})\right]^\intercal \widehat{\Gamma}\left[\widehat{q} + t(q^* - \widehat{q})\right].
\]
that results in 
\[
2(q^*)^\intercal \widehat{\Gamma}(q^* - \widehat{q}) + (t-2)(q^* - \widehat{q})\widehat{\Gamma}(q^* - \widehat{q}) \geq 0.
\]
Since $2-t>0$, it holds that
\begin{equation}
    \left(q^* - \widehat{q}\right)^\intercal \widehat{\Gamma}\left(q^* - \widehat{q}\right) \leq \frac{2}{2-t}(q^*)^\intercal \widehat{\Gamma}(q^* - \widehat{q}).
    \label{eq: lemma-approx-2}
\end{equation}
It follows from \eqref{eq: lemma-approx-1} that
\[
(q^*)^\intercal \widehat{\Gamma}(q^* - \widehat{q}) = (q^*)^\intercal {\Gamma}(q^* - \widehat{q}) + (q^*)^\intercal \left(\widehat{\Gamma}-\Gamma\right)(q^* - \widehat{q}) \leq (q^*)^\intercal \left(\widehat{\Gamma}-\Gamma\right)(q^* - \widehat{q}).
\]
Combined with \eqref{eq: lemma-approx-2}, we have
\begin{equation}
    \left(q^* - \widehat{q}\right)^\intercal \widehat{\Gamma}\left(q^* - \widehat{q}\right) \leq \frac{2}{2-t}(q^*)^\intercal \left(\widehat{\Gamma}-\Gamma\right)(q^* - \widehat{q}) \leq \frac{2\|q^*\|_2}{2-t} \|\widehat{\Gamma} - \Gamma\|_2 \left\|q^* - \widehat{q}\right\|_2.
    \label{eq: lemma-approx-3}
\end{equation}
Since the $\widehat{q}$ and $q$ are symmetric, we switch the roles of $\{\widehat{\Gamma},\widehat{q}\}$, $\{\Gamma, q\}$ and establish:
\begin{equation}
    \left(q^* - \widehat{q}\right)^\intercal \Gamma \left(q^* - \widehat{q}\right) \leq \frac{2\|\widehat{q}\|_2}{2-t} \|\widehat{\Gamma} - \Gamma\|_2 \left\|q^* - \widehat{q}\right\|_2.
\end{equation}
If $\lambda_{\rm min}(\Gamma) > 0$, it follows that 
\[
\lambda_{\rm min}(\Gamma)\left\|q^* - \widehat{q}\right\|_2^2 \leq \frac{2\|\widehat{q}\|_2}{2-t} \|\widehat{\Gamma} - \Gamma\|_2 \left\|q^* - \widehat{q}\right\|_2.
\]
Take $t\rightarrow 0_+$, we establish that
\[
\left\|q^* - \widehat{q}\right\|_2 \leq \frac{1}{\lambda_{\min}(\Gamma)}\left\|\widehat{\Gamma} - \Gamma\right\|_2 \|\widehat{q}\|_2 \leq \frac{\|\widehat{\Gamma} - \Gamma\|_F}{\lambda_{\min}(\Gamma)} \leq \frac{L\|\widehat{\Gamma} - \Gamma\|_\infty}{\lambda_{\min}(\Gamma)}.
\]
Lastly, considering that both $q^*\in \HH$ and $\widehat{q}\in \HH$, so 
\[
\left\|q^* - \widehat{q}\right\|_2\leq \frac{L\|\widehat{\Gamma} - \Gamma\|_\infty}{\lambda_{\min}(\Gamma)}\wedge \rho_\HH.
\]

\subsection{Proof of Lemma \ref{lemma: proof-coro-linear}}
    Let $|\A_l| = \sum_{i=1}^{n_G}\ind_{[U_i\leq p_l]}$, by Hoeffeding's inequality, it holds that
    \[
    \prob\left(\left|\frac{1}{n_G}\sum_{i=1}^{n_G}(\ind_{[U_i\leq p_l]} - p_l)\right|\geq t\right)\leq 2\exp(-2n_G t^2).
    \]
    Let $t = \sqrt{\frac{\log(1/c_1)}{2n_G}}$. Then with probability greater than $1-2c$, the following event holds
    \begin{equation}
        \left|\frac{|\A_l|}{n_G} - p_l\right|\leq \sqrt{\frac{\log(1/c_1)}{2n_G}}.
        \label{proof-coro-linear-2}
    \end{equation}
    In the following, we will discuss how to identify such constant $C_1$ to make $\left||\A_l|/N - p_l/(1-p_l)\right|\leq C_1 \sqrt{\frac{1}{n_G}}$. 

    \noindent We firstly discuss the case that $|\A_l| /N - p_l/(1-p_l) \leq C_1 \sqrt{\frac{1}{n_G}}$.
    Considering the case that $n_G$ is sufficiently large such that $n_G > \frac{\log(1/c_1)}{(1-p_l)^2}$,
    \begin{align*}
        \frac{|\A_l|}{N} = \frac{|\A_l|/n_G}{N/n_G} = \frac{|\A_l|/n_G - p_l + p_l}{1 - p_l + p_l - |\A_l|/n_G} \leq \frac{\sqrt{\frac{\log(1/c_1)}{2n_G}}+p_l}{1-p_l - \sqrt{\frac{\log(1/\epsilon)}{2n_G}}},
    \end{align*}
    where the last inequality is due to \eqref{proof-coro-linear-2}.
    To show that $|\A_l| /N - p_l/(1-p_l) \leq C_1\frac{1}{\sqrt{n_G}}$, it suffices to show that
    \[
    \frac{\sqrt{\frac{\log(1/c_1)}{2n_G}}+p_l}{1-p_l - \sqrt{\frac{\log(1/c_1)}{2n_G}}} - \frac{p_l}{1-p_l} \leq C_1\frac{1}{\sqrt{n_G}}.
    \]
    The preceding inequality holds as long as 
    \[
    C_1 \geq \frac{\frac{1}{1-p_l}\sqrt{\log(1/c_1)}}{\sqrt{2}(1-p_l) - \sqrt{\frac{\log(1/c_1)}{n_G}}} \quad \textrm{for any $n_G>\frac{\log(1/c_1)}{(1-p_l)^2}$}.
    \]
    Thus we make $C_1 \geq \frac{\sqrt{\log(1/c_1)}}{(\sqrt{2}-1)(1-p_l)^2}$. 

    \noindent For the other case that $p_l/(1-p_l) - |\A_l|/N \leq C_1\sqrt{\frac{1}{n_G}}$. Consider the case that $n_G$ is sufficiently large such that $n_G > \frac{\log(1/c_1)}{p_l^2}$,
    \[
    \frac{|\A_l|}{N} = \frac{|\A_l|/n_G}{N/n_G} = \frac{|\A_l|/n_G - p_l + p_l}{1 - p_l + p_l - |\A_l|/n_G} \geq \frac{p_l - \sqrt{\frac{\log(1/c_1)}{2n_G}}}{1-p_l + \sqrt{\frac{\log(1/c_1)}{2n_G}}},
    \]
    where the last inequality is due to \eqref{proof-coro-linear-2}. To show that $p_l/(1-p_l) - |\A_l|/N \leq C_1\sqrt{\frac{1}{n_G}}$, it suffices to show that
    \[
    \frac{p_l}{1-p_l} - \frac{p_l-\sqrt{\frac{\log(1/c_1)}{2n_G}}}{1-p_l+ \sqrt{\frac{\log(1/c_1)}{2n_G}}} \leq C_1\sqrt{\frac{1}{n_G}}.
    \]
    The preceding inequality holds as long as 
    \[
    C_1 \geq \frac{\frac{1}{1-p_l}\sqrt{\log(1/c_1)}}{\sqrt{2}(1-p_l) + \sqrt{\frac{\log(1/c_1)}{n_G}}} \quad \textrm{for any $n_G>\frac{\log(1/c_1)}{p_l^2}$}.
    \]
    Thus we make $C_1 = \frac{\sqrt{\log(1/c_1)}}{\sqrt{2}(1-p_l)^2}$. 

    \noindent To summarize, when $n_G$ is sufficiently large such that $n_G > \frac{\log(1/c_1)}{\min\{(1-p_l)^2, p_l^2\}}$, with probability at least $1-2c_1$, it holds that
    \[
    \left|\frac{|\A_l|}{N} - \frac{p_l}{1-p_l}\right|\leq C_1\sqrt{\frac{1}{n_G}},
    \]
    where $C_1 = \frac{\sqrt{\log(1/c_1)}}{(\sqrt{2}-1)(1-p_l)^2}$.


\end{document}